\documentclass{article}
\usepackage{graphicx} 
\usepackage[utf8]{inputenc}
\usepackage[]{amsthm} 
\usepackage{caption}
\usepackage[]{amssymb} 
\usepackage[]{amsmath,txfonts}
\usepackage{float}
\usepackage{longtable}
\usepackage{bm}
\usepackage{multirow}
\usepackage{subcaption}
\usepackage{graphicx}
\graphicspath{ {./images/} }
\usepackage[toc,page]{appendix}
\usepackage{lscape}
\usepackage{multirow}
\usepackage{booktabs}
\usepackage{array}    
\usepackage{hyperref}
\hypersetup{
    colorlinks=true, allcolors = blue
    }
\usepackage{authblk}
\usepackage{stackengine}
\usepackage[style=apa]{biblatex}
\title{Advancing Causal Inference: A Nonparametric Approach to ATE and CATE Estimation with Continuous Treatments}
\author[1]{Hugo Gobato Souto}
\affil[1]{\stackunder{{\stackunder{Institute of Mathematics and Computer Sciences at University of São Paulo, Brazil}{Av. Trab. São Carlense 400, 13566-590 São Carlos (SP), Brazil}}}{\stackunder{{hgsouto@usp.br}. {https://orcid.org/0000-0002-7039-0572}}}}
\author[2]{Francisco Louzada Neto}
\affil[2]{\stackunder{{\stackunder{Institute of Mathematics and Computer Sciences at University of São Paulo, Brazil}{Av. Trab. São Carlense, 400, São Carlos, 13566-590, Brazil}}}{\stackunder{{louzada@icmc.usp.br}. {https://orcid.org/0000-0001-7815-9554}}}}
\date{}

\begin{document}

\maketitle

\begin{abstract}
    This paper introduces a generalized ps-BART model for the estimation of Average Treatment Effect (ATE) and Conditional Average Treatment Effect (CATE) in continuous treatments, addressing limitations of the Bayesian Causal Forest (BCF) model. The ps-BART model's nonparametric nature allows for flexibility in capturing nonlinear relationships between treatment and outcome variables. Across three distinct sets of Data Generating Processes (DGPs), the ps-BART model consistently outperforms the BCF model, particularly in highly nonlinear settings. The ps-BART model's robustness in uncertainty estimation and accuracy in both point-wise and probabilistic estimation demonstrate its utility for real-world applications. This research fills a crucial gap in causal inference literature, providing a tool better suited for nonlinear treatment-outcome relationships and opening avenues for further exploration in the domain of continuous treatment effect estimation.
\end{abstract}

\paragraph{Key Words}: Conditional Average Treatment Effect, Average Treatment Effect, Bayesian Additive Regression Trees, Bayesian Causal Forest Model, Continuous Treatment Effect.

\section{Introduction} \label{Sec1}

Causal inference aims to estimate the causal effect of a treatment on an outcome, adjusting for confounding factors that may influence both the treatment assignment and the outcome \parencite{Yao2021,10.1145/3394486.3406460,Kuang2020, Crown2019, Grimmer2014, Ohlsson2020, Bembom2007,Sobel2000, murnane2010methods}. In the context of continuous treatments, the estimation of causal effects becomes more complex than in the binary treatment case due to the need to model the treatment effect as a continuous function rather than a fixed difference \parencite{Hirano2004,https://doi.org/10.48550/arxiv.1811.00157,Wu2022,https://doi.org/10.48550/arxiv.1907.13258}.

Consider the potential outcomes framework, which is fundamental in causal inference. Let $Y(d)$ denote the potential outcome corresponding to treatment level $d \in \mathbb{R}$, where $D$ is a continuous treatment variable, and $X$ represents a vector of covariates. The observed outcome for an individual is given by $Y = Y(D)$. The primary objective is to estimate the causal effect of varying the treatment $D$ while controlling for covariates $ \mathbf{X}$.

In the case of binary treatments ($D \in \{0, 1\}$), the Average Treatment Effect (ATE) is defined as the difference between the expected potential outcomes:
\begin{equation}
\text{ATE} = \mathbb{E}[Y(1) - Y(0)].
\end{equation}
Similarly, the Conditional Average Treatment Effect (CATE) is defined as:
\begin{equation}
\text{CATE}(\mathbf{X} = \mathbf{x}) = \mathbb{E}[Y(1) - Y(0) \mid \mathbf{X} = \mathbf{x}].
\end{equation}
For binary treatments, ATE and CATE are scalar quantities, representing the difference in expected outcomes between two discrete levels of treatment.

However, when the treatment variable $D$ is continuous, the causal effect of interest becomes a function of the treatment level \parencite{Kennedy2016,Moodie2010}. The ATE for continuous treatments is expressed as a function of the treatment dose:
\begin{equation}
\phi(D) = \mathbb{E}[Y(D)],
\end{equation}
where $\phi(D)$ represents the dose-response function. The continuous ATE is often referred to as the dose-response function, which captures the expected outcome as a function of the continuous treatment level. In this context, the estimation challenge shifts from a scalar difference to modeling a function that may exhibit complex nonlinearities depending on the treatment dose.

The CATE for continuous treatments further generalizes the concept of CATE in binary settings \parencite{Kennedy2016,Moodie2010}. It is defined as:
\begin{equation}
\phi(D \mid \mathbf{X} = \mathbf{x}) = \mathbb{E}[Y(D) \mid \mathbf{X} = \mathbf{x}].
\end{equation}
This represents the expected outcome as a function of the treatment level $D$, conditional on the covariates $\mathbf{X}$. Unlike the binary case, where CATE is a scalar for each covariate value, in the continuous case, CATE is a function of both the treatment level and the covariates \parencite{https://doi.org/10.48550/arxiv.2007.09845}. The complexity arises from the fact that this function must account for the interaction between the treatment dose and the covariates, which may vary across different regions of the covariate space.

The estimation of ATE and CATE for binary and continuous treatments involves fundamental differences in both the interpretation and the modeling approach \parencite{Kennedy2016,Moodie2010,https://doi.org/10.48550/arxiv.2007.09845}. In the binary treatment case, the estimation typically involves models that predict the difference in outcomes between the treated and untreated groups, often using methods like propensity score matching or inverse probability weighting \parencite{Hahn2020,pmlr-v130-curth21a, CausalBook}. In the continuous treatment case, on the other hand, the model must capture the entire dose-response relationship, which may involve nonparametric or semi-parametric methods that do not assume a specific functional form for $\phi(D)$ \parencite{Kennedy2016,Moodie2010,https://doi.org/10.48550/arxiv.2007.09845}. The complexity increases with the dimensionality of $\mathbf{X}$, as the conditional treatment effect function $\phi(D \mid \mathbf{X} = \mathbf{x})$ must be estimated across the covariate space, which can be computationally intensive and requires careful regularization to avoid overfitting \parencite{Hirano2004,https://doi.org/10.48550/arxiv.1811.00157,Wu2022,https://doi.org/10.48550/arxiv.1907.13258}.

Additionally, the interpretation of ATE and CATE varies for binary and continuous treatments. For binary treatments, ATE and CATE provide clear comparisons between two scenarios (treatment vs. no treatment), making the interpretation straightforward \parencite{Knzel2019,https://doi.org/10.48550/arxiv.2004.14497,https://doi.org/10.48550/arxiv.1712.04912}. On the other hand, for continuous treatments, the interpretation involves understanding how the expected outcome changes with varying doses of the treatment. The CATE, in particular, requires analyzing how the treatment effect varies not just with the treatment level but also with the covariates. If two individuals have covariates that are close in the covariate space, the difference in their treatment effect functions $\phi(D \mid \mathbf{X} = \mathbf{x_1})$ and $\phi(D \mid \mathbf{X} = \mathbf{x_2})$ should be small. Conversely, if the covariates are far apart, the treatment effect functions may differ significantly.

These differences necessitate distinct approaches in both the theoretical framework and the empirical modeling strategies for ATE and CATE in the context of continuous treatments. The challenges inherent in estimating these functions, especially the conditional effect $\phi(D \mid \mathbf{X} = \mathbf{x})$, underscore the need for robust nonparametric methods that can flexibly model the potentially complex relationships between the treatment, covariates, and outcomes.

Despite the growing interest in causal inference for continuous treatments, significant gaps remain in the estimation of the CATE for these treatments \parencite{https://doi.org/10.48550/arxiv.2007.09845}. While there exists a considerable body of literature focusing on the estimation of the ATE for continuous treatments \parencite{Hirano2004,https://doi.org/10.48550/arxiv.1811.00157,Wu2022,https://doi.org/10.48550/arxiv.1907.13258, Knzel2019,https://doi.org/10.48550/arxiv.2004.14497,https://doi.org/10.48550/arxiv.1712.04912}, the extension to CATE has been comparatively understudied, especially for probabilistic estimations of the CATE functions \parencite{https://doi.org/10.48550/arxiv.2007.09845}. This gap is particularly pronounced given the practical importance of CATE, which allows for the heterogeneity of treatment effects across different levels of covariates $\mathbf{X}$. 

Another critical gap in the current literature is the lack of models that provide probabilistic estimates of ATE and CATE for continuous treatments \parencite{https://doi.org/10.48550/arxiv.2007.09845}. Probabilistic estimation, which includes the construction of confidence intervals (CIs) around the estimates, is crucial for quantifying the uncertainty associated with the treatment effect. Without these intervals, practitioners cannot assess the reliability of the estimates, making it difficult to draw robust conclusions from the data \parencite{https://doi.org/10.48550/arxiv.2007.09845}.

A notable exception in this landscape is the work of \citeauthor{https://doi.org/10.48550/arxiv.2007.09845} (\citeyear{https://doi.org/10.48550/arxiv.2007.09845}), who proposed a generalization of the nonparametric Bayesian Causal Forest (BCF) model to estimate both ATE and CATE for continuous treatments. The BCF model originally introduced for binary treatments was adapted by \citeauthor{https://doi.org/10.48550/arxiv.2007.09845} (\citeyear{https://doi.org/10.48550/arxiv.2007.09845}) to handle continuous treatments, marking a significant advancement in the field. Nonetheless, to do so, \citeauthor{https://doi.org/10.48550/arxiv.2007.09845} (\citeyear{https://doi.org/10.48550/arxiv.2007.09845}) needed to make the assumption effect of the treatment exposure on the outcome is linear, transforming the generalization of the BCF model into a semi-nonparametric approach.

While their model represents a significant step forward in terms of incorporating uncertainty into the estimation of ATE and CATE for continuous treatments, it does so at the cost of imposing a parametric structure on the relationship between $D$ and $Y$. This trade-off limits the model's applicability in scenarios where the treatment effect is not linear, potentially leading to biased estimates when the true relationship deviates from this assumption (i.e., model misspecification) \parencite{Hayashi2011,Liu2016,Martin2022,Cai2003}.

To address this limitation of the BCF model for continuous treatments, this paper proposes a model that generalizes the propensity score Bayesian Additive Regression Trees (ps-BART) model for the estimation of ATE and CATE for continuous treatments. The ps-BART model was initially proposed as a benchmark in the original paper introducing the BCF model by \citeauthor{Hahn2020} (\citeyear{Hahn2020}). However, unlike the BCF model, which incorporates a linear term to model the effect of $D$ on $Y$, the ps-BART model in its generalized form does not assume that the relationship between $D$ and $Y$ is linear.

The key advantage of the ps-BART model lies in its nonparametric nature. By not imposing a linearity assumption, the model can capture complex interactions and nonlinearities that may be present in the data. This is particularly valuable in settings where the treatment effect is expected to vary in a non-monotonic or interaction-dependent manner across different levels of $D$ and different covariate configurations.

The generalized ps-BART model offers several advantages over the BCF model:
\begin{enumerate}
    \item \textbf{Flexibility}: The model's nonparametric nature allows it to adapt to a wide variety of functional forms, capturing both simple and complex relationships between $D$, $X$, and $Y$.
    \item \textbf{Accuracy}: By avoiding the imposition of a linear structure, the model reduces the risk of misspecification and bias, particularly in settings where the true treatment effect is nonlinear or involves intricate interactions between covariates.
    \item \textbf{Generalizability}: The model can be applied to a broad range of DGPs, making it suitable for diverse empirical applications where the nature of the treatment effect is unknown or suspected to be nonlinear.
\end{enumerate}

In summary, this research addresses the current limitations of the existing models for estimation of continuous treatments effect, namely by proposing a fully nonparametric model that can estimate both ATE and CATE functions without the treatment-outcome-linearity assumption while providing CIs for these function estimations. Thereby, this paper contributes to the development of more robust methods in causal inference for continuous treatments effect estimation.

The remainder of this paper is composed of the following sections: Section \ref{Sec2} presents the proposed model for the estimation of continuous treatments effect, while Section \ref{Sec3} thoroughly describes the experimental design of this research to put the proposed model to the test against the current benchmark model (i.e., the BCF model). Additionally, the results of this paper's experiments and their respective analysis can be found in Section \ref{Sec4}. Finally, Section \ref{Sec5} concludes this paper by presenting the novel insights drawn from the experiments and implications of the results of these experiments.

\section{ps-BART for Continuous Treatments} \label{Sec2}

The Bayesian Additive Regression Trees (BART) model, originally developed by \citeauthor{Chipman2010} (\citeyear{Chipman2010}), is a Bayesian ensemble learning technique that constructs a predictive model by summing the outputs of multiple regression trees. Each tree in the ensemble makes a small contribution to the overall prediction, enabling BART to model intricate and nonlinear patterns in the data.

In BART, the response variable \( Y \) is modeled as the sum of the outputs from several regression trees, along with an additive error term. This model can be expressed mathematically as:
\begin{equation}
    Y_i = \sum_{j=1}^m h_j(\mathbf{x_i}; T_j, M_j) + \epsilon_i, \quad \epsilon_i \sim \mathcal{N}(0, \sigma^2),
\end{equation}
where \( Y_i \) denotes the response for the \( i \)-th observation, \( h_j(\cdot) \) is the function representing the \( j \)-th tree with structure \( T_j \) and terminal node parameters \( M_j \), \( \mathbf{x_i} \) are the covariates associated with the \( i \)-th observation, \( m \) is the number of trees, and \( \epsilon_i \) is the normally distributed error term with zero mean and variance \( \sigma^2 \). In the Bayesian framework, it is necessary to define priors for the model parameters, including the tree structures \( T_j \), the terminal node parameters \( M_j \), and the variance \( \sigma^2 \) of the errors.

The prior distribution for the tree structure \( T_j \) typically involves specifying the probability of a split at each node. This probability is commonly defined as:
\begin{equation}
    p(\text{split at node } t) = \alpha (1 + d_t)^{-\beta},
\end{equation}
where \( d_t \) denotes the depth of node \( t \), and \( \alpha \) and \( \beta \) are hyperparameters that control the depth of the tree, often set to 0.95 and 2, respectively \parencite{Chipman2010}. For the terminal node parameters \( M_j \), a normal distribution is usually chosen as the prior:
\begin{equation}
    \mu_{j,k} \sim \mathcal{N}(\mu_{\mu}, \sigma^2_\mu),
\end{equation}
where \( \mu_{j,k} \) is the mean response at the \( k \)-th terminal node of the \( j \)-th tree, and \( \mu_{\mu} \) and \( \sigma^2_\mu \) are hyperparameters. These hyperparameters are typically set by solving the equations \( m\mu_\mu - k\sqrt{m}\sigma_\mu = y_{\text{min}} \) and \( m\mu_\mu + k\sqrt{m}\sigma_\mu = y_{\text{max}} \) for a chosen value of \( k \) \parencite{Chipman2010}. A common choice is \( k = 2 \), which ensures a 95\% prior probability that \( \mathbb{E}(Y \mid x) \) lies within the interval \( (y_{\text{min}}, y_{\text{max}}) \).

The prior for the error variance \( \sigma^2 \) is typically chosen to follow an inverse gamma distribution:
\begin{equation}
    \sigma^2 \sim \text{IG}(\nu/2, \nu \lambda/2),
\end{equation}
where \( \nu \) and \( \lambda \) are hyperparameters. To specify these hyperparameters, a data-driven approach is often employed. This involves calibrating the prior degrees of freedom \( \nu \) and the scale parameter \( \lambda \) using a rough estimate \( \hat{\sigma} \) of \( \sigma \). Common strategies for estimating \( \hat{\sigma} \) include: (1) using the sample standard deviation of \( Y \) (the “naive” approach), or (2) using the residual standard deviation from a least squares linear regression of \( Y \) on the covariates \( X \) (the “linear model” approach), with the latter often providing better calibration \parencite{Chipman2010}. A typical value for \( \nu \) ranges between 3 and 10, which determines the shape of the distribution, and \( \lambda \) is chosen so that the \( q \)-th quantile of the prior distribution of \( \sigma \) corresponds to \( \hat{\sigma} \), i.e., \( P(\sigma < \hat{\sigma}) = q \). Common choices for \( q \) are 0.75, 0.90, or 0.99, effectively centering the distribution below \( \hat{\sigma} \). The default values for \( (\nu,q) \) are (3, 0.90) \parencite{Chipman2010}.

Additionally, the number of trees \( m \) used in the BART model is another hyperparameter, typically set to 200 \parencite{Chipman2010}.

Given the prior specifications and the likelihood function, the posterior distribution of the BART model parameters can be obtained via Markov Chain Monte Carlo (MCMC) methods. The objective is to sample from the joint posterior distribution:
\begin{equation}
    p(T_1, M_1, \ldots, T_m, M_m, \sigma^2 \mid Y, X).
\end{equation}

The MCMC procedure for BART generally involves the following steps:
\begin{enumerate}
    \item \textbf{Initialization}: Begin by initializing the tree structures \( T_j \), the terminal node parameters \( M_j \), and the error variance \( \sigma^2 \).
    \item \textbf{Gibbs Sampling}: Each tree structure and terminal node parameter is updated iteratively using Gibbs sampling, while conditioning on the other trees and the observed data. Specifically, for each tree \( j \):
    \begin{itemize}
        \item The tree structure \( T_j \) is updated using a Metropolis-Hastings step.
        \item The terminal node parameters \( M_j \) are updated by sampling from their full conditional posterior distribution.
    \end{itemize}
    \item \textbf{Update Error Variance}: The error variance \( \sigma^2 \) is updated from its full conditional posterior distribution.
    \item \textbf{Repeat}: These steps are repeated over many iterations to ensure that the samples converge to the posterior distribution.
\end{enumerate}

Upon convergence of the MCMC algorithm, predictions can be obtained by averaging the posterior samples of the trees.

The ps-BART model, on the other hand, is an extension of the BART framework designed to address specific challenges in causal inference, particularly those arising from Regularization-Induced Confounding (RIC). The ps-BART model enhances the BART model by incorporating an estimated propensity score, which is itself obtained through a BART model, to adjust for potential confounding effects that may distort the estimation of treatment effects \parencite{Hahn2020}.

RIC is a phenomenon that occurs in Bayesian models where regularization, imposed through prior distributions or model structures, influences the posterior distribution of treatment effects in unintended ways \parencite{Hahn2018}. The core of the RIC issue lies in the fact that multiple functional forms may yield similar likelihoods but imply vastly different causal effects. This ambiguity becomes particularly problematic in contexts characterized by strong confounding and relatively modest treatment effects \parencite{Hahn2018}.

Consider a scenario where the conditional expectation of the outcome variable \( Y \), given the covariates \( \mathbf{X} \) and treatment \( D \), is predominantly determined by the covariates rather than the treatment, leading the posterior distribution of the treatment effect to be heavily influenced by the prior over the conditional expectation function \parencite{Hahn2018}. This issue is exacerbated in scenarios where treatment assignment is based on predictions of the outcome in the absence of treatment—a situation known as target selection. Here, the covariates \( \mathbf{X} \) are used to assign treatment \( D \) based on predicted outcomes \( \mathbb{E}[Y \mid \mathbf{X} = \mathbf{x}, D = 0] \), creating a feedback loop that strengthens the confounding effect. Consequently, the posterior estimate of the treatment effect can be significantly biased, as the regularization inadvertently confounds the estimated relationship between \( D \) and \( Y \) \parencite{Hahn2018}.

To address the challenges posed by RIC, the ps-BART model incorporates an estimated propensity score \( \hat{\pi}(\mathbf{x}) \) into the BART framework \parencite{Hahn2020}. The propensity score, defined as the conditional probability of receiving treatment given the covariates, \( \pi(\mathbf{x}) = P(D = d \mid \mathbf{X} = \mathbf{x}) \), is estimated using a separate BART model. Given the fact that we are dealing with a continuous treatment, it is better to use the estimated CDF of the conditional probability of receiving treatment given the covariates, namely \( \pi(\mathbf{x}) = P(D \leq d \mid \mathbf{X} = \mathbf{x}) \). This score captures the treatment assignment mechanism, allowing the model to account explicitly for confounding factors.

The inclusion of the propensity score \( \hat{\pi}(\mathbf{x}) \) in the outcome model modifies the BART framework as follows:
\begin{equation}
Y_i = \sum_{j=1}^m h_j(\mathbf{x_i}, d_i, \hat{\pi}(\mathbf{x_i}); T_j, M_j) + \epsilon_i,
\end{equation}
where \( h_j(x_i, \hat{\pi}(\mathbf{x_i}); T_j, M_j) \) represents the \( j \)-th regression tree, which now takes both the covariates \( \mathbf{x_i} \), the treatment dose \( d_i \), and the estimated propensity score \( \hat{\pi}(\mathbf{x_i}) \) as inputs. The inclusion of \( \hat{\pi}(\mathbf{x_i}) \) helps disentangle the effect of the treatment from the effect of the covariates, thereby reducing the influence of RIC.

By incorporating \( \hat{\pi}(\mathbf{x_i}) \), the ps-BART model effectively adjusts for the treatment assignment process, mitigating the risk that the prior over the conditional expectation function \( f(\mathbf{x_i}) \) overly influences the posterior distribution of the treatment effect. This adjustment is particularly crucial in cases where target selection is present, as it ensures that the model remains robust against the feedback loop created by the treatment assignment mechanism.


After the ps-BART model has been properly trained, the ATE and CATE functions as described in Section \ref{Sec1} can be approximated using the following formulations:
\begin{equation}
  \hat{\phi}(D) = \sum_{j=1}^m h_j(\mathbf{\bar{x}}, D, \hat{\pi}(\mathbf{\bar{x}}); T_j, M_j)
\end{equation}

\begin{equation}
    \hat{\phi}(D \mid \mathbf{X} = \mathbf{x_i}) = \sum_{j=1}^m h_j(\mathbf{x_i}, D, \hat{\pi}(\mathbf{x_i}); T_j, M_j)
\end{equation}

where, $\mathbf{\bar{x}}$ is the empirical mean of the covariates $\mathbf{x}$. By estimating $\hat{\phi}(D)$ and $\hat{\phi}(D \mid \mathbf{X} = \mathbf{x_i})$ for a considerable number of points and interpolating them, we can approximate the real ATE and CATE functions (where the quality of this approximation depends on the performance of the ps-BART model). Additionally, given the bayesian aspect of the ps-BART model, we can also estimate CIs for $\hat{\phi}(D)$ and $\hat{\phi}(D \mid \mathbf{X} = \mathbf{x_i})$.

\section{Experimental Design} \label{Sec3}

\subsection{Data Generating Processes (DGPs) for Model Evaluation}

To rigorously evaluate the performance of the proposed ps-BART model in comparison to the benchmark BCF model, a comprehensive experimental design was implemented. This design involved the use of nine distinct DGPs, each carefully constructed to test the models under various conditions that mimic real-world scenarios in causal inference. The DGPs were designed to span a range of complexities, including linear and nonlinear relationships between the treatment and outcome, as well as varying levels of confounding.

The nine DGPs employed in this study were structured to assess the models across different sample sizes and simulation conditions. Specifically, for each DGP, datasets were generated with 100, 250, and 500 observations. Additionally, each scenario was subjected to 100 independent simulations to ensure the robustness and generalizability of the results. This approach resulted in a total of 2,700 distinct datasets, providing a broad basis for comparison between the ps-BART and BCF models.

The first four DGPs are extensions of the model initially proposed by \citeauthor{Hirano2004} (\citeyear{Hirano2004}), which was later refined by \citeauthor{Moodie2010} (\citeyear{Moodie2010}). Though \citeauthor{Moodie2010} (\citeyear{Moodie2010}) also enhanced this DGP by introducing a temporal component, allowing for the consideration of time-dependent effects, this study does not make use of this temporal component for the DGP. Thus, the DGP used for this study is given as:
\begin{enumerate}
    \item Let \( X_1 \) and \( X_2 \) be independent random variables drawn from an exponential distribution with rate parameter \( \lambda = 1 \):
\[
X_1 \sim \text{Exp}(1), \quad X_2 \sim \text{Exp}(1).
\]
\item Define a constant \( \alpha \), where $ \alpha \in \{ 1, 2, 4, 8\}$.
\item The treatment variable \( D \) is generated as a random variable from an exponential distribution with a rate parameter that depends on \( X_1 \) and \( X_2 \):
\[
D \sim \text{Exp}\left(\frac{\alpha}{X_1 + X_2}\right).
\]
\item The outcome variable \( Y \) is generated from a normal distribution where the mean is a function of \( D \), \( X_1 \), and \( X_2 \), and the standard deviation is 1:
\[
Y \sim \mathcal{N}\left(D + (X_1 + X_2) \exp\left(-D \cdot (X_1 + X_2)\right), 1\right).
\]
\end{enumerate}

By changing the \( \alpha \) value, one can increase the nonlinearity and heterogeneity aspect of the continuous treatment effect of the DGP by increasing \( \alpha \). Thus, the choice of various \( \alpha \) values, which differs from the proposed DGP of \citeauthor{Moodie2010} (\citeyear{Moodie2010}), allows us to test the ps-BART and BCF models for different levels of nonlinearity and heterogeneity of the continuous treatment effect. Given the fact that (semi-)parametric models outperform nonparametric models when correctly specified for the underlying DGP \parencite{Jabot2015,Liu2011,Robinson2010}, it is already expected that the relative performance of the ps-BART model will improve as the \( \alpha \) values increase. Thus, the choice of multiple \( \alpha \) values is to determine to the extent of treatment-outcome relationship nonlinearity where the ps-BART model becomes superior over the BCF model.

Moving to the next three DGPs employed in this research, they are based on those proposed by \citeauthor{Wu2022} (\citeyear{Wu2022}). These DGPs are specifically designed to capture more complex, nonlinear relationships between the treatment and the outcome, incorporating interaction terms and higher-order polynomial effects. They are structured as:

\begin{enumerate}
    \item Let \( C_1, C_2, C_3, C_4 \) be independent random variables drawn from a standard normal distribution, and let \( C_5 \) and \( C_6 \) be independent random variables drawn from uniform distributions with specified ranges:
\[
C_1, C_2, C_3, C_4 \sim \mathcal{N}(0, 1), \quad C_5 \sim \text{Unif}(-2, 2), \quad C_6 \sim \text{Unif}(-3, 3).
\]
\item The treatment variable \( D \) is generated based on the following model specifications, depending on the scenario \( \text{spec} \in \{1, 2, 3\} \):
\begin{itemize}
    \item Specification 1:  \[
   D = 9 \left(0.8 + 0.1C_1 + 0.1C_2 - 0.1C_3 + 0.2C_4 + 0.1C_5\right) - 3 + \epsilon_D,
   \]
   where \( \epsilon_D \sim \mathcal{N}(0, 5) \).
   \item Specification 2: \[
   D = 15 \left(0.8 + 0.1C_1 + 0.1C_2 - 0.1C_3 + 0.2C_4 + 0.1C_5\right) + 2 + 2\epsilon_D,
   \]
   where \( \epsilon_D \sim t_4 \) (Student's t-distribution with 4 degrees of freedom).
   \item Specification 3: \[
   D = 15 \left(0.8 + 0.1C_1 + 0.1C_2 - 0.1C_3 + 0.2C_4 + 0.1C_5\right) + 3\sqrt{2}C_3^2 + \epsilon_D,
   \]
   where \( \epsilon_D \sim t_4 \) (Student's t-distribution with 4 degrees of freedom).
\end{itemize}
\item The outcome variable \( Y \) is generated as a function of \( D \) and the confounders \( C_1, C_2, C_3, C_4, C_5 \). The true mean function \( \mu_{DC} \) is defined as:
\[
\mu_{DC} = -10 - (2C_1 + 2C_2 + 3C_3 - C_4 + 2C_5) - D \left(0.1 - 0.1C_1 + 0.1C_4 + 0.1C_5 + 0.1C_3^2\right) + 0.13D^3.
\]
The outcome \( Y \) is then drawn from a normal distribution with mean \( \mu_{DC} \) and standard deviation \( \sqrt{10} \):
\[
Y \sim \mathcal{N}\left(\mu_{DC}, \sqrt{10}\right).
\]
\end{enumerate}

The final two DGPs used in the experimental design are derived from the heterogeneous nonlinear DGP presented by \citeauthor{Hahn2020} (\citeyear{Hahn2020}). The first DGP is given as:
\begin{enumerate}
    \item Let \( X_1, X_2, X_3 \) be independent random variables drawn from a standard normal distribution, \( X_4 \) be a categorical variable taking values from \(\{1, 2, 3\}\) with equal probability, and \( X_5 \) be a binary variable taking values from \(\{0, 1\}\) with equal probability:
\[
X_1, X_2, X_3 \sim \mathcal{N}(0, 1), \quad X_4 \sim \text{Categorical}\{1, 2, 3\}, \quad X_5 \sim \text{Bernoulli}(0.5).
\]
  \item The treatment variable \( D \) is generated as an exponential random variable, where the rate parameter is a function of the absolute values of \( X_1 \), \( X_2 \), and \( X_3 \), along with \( X_5 \):
\[
D \sim \text{Exp}\left(\frac{\alpha}{|X_1| + X_5 + |X_2 - X_3|}\right),
\]
where \( \alpha = 2 \).
  \item The treatment effect \( \tau(X_2, X_5) \) is defined as:
   \[
   \tau(X_2, X_5) = 1 + 2 X_2 X_5.
   \]
   \item The baseline outcome function \( \mu(X_4, X_1, X_3) \) is defined as:
   \[
   \mu(X_4, X_1, X_3) = -6 + g(X_4) + 6 |X_3 - 1|.
   \]
   where \[
g(X_4) = 
\begin{cases}
2, & \text{if } X_4 = 1, \\
-1, & \text{if } X_4 = 2, \\
-4, & \text{if } X_4 = 3.
\end{cases}
\]

   \item The outcome is computed as:
   \[
   Y = \mu(X_4, X_1, X_3) + D \cdot \tau(X_2, X_5).
   \]
\end{enumerate}

Concerning the second DGP, it follows the exact same structure of the first one with the difference that the outcome is computed as:
\[
   Y = \mu(X_4, X_1, X_3) + log(D) \cdot \tau(X_2, X_5).
\]

Hence, the first DGP inspired by the heterogeneous nonlinear DGP presented by \citeauthor{Hahn2020} (\citeyear{Hahn2020}) has a continuous treatment that has a linear relationship with the outcome variable while the second DGP has a continuous treatment that has a nonlinear relationship with the outcome variable. Given the fact that (semi-)parametric models outperform nonparametric models when correctly specified for the underlying DGP \parencite{Jabot2015,Liu2011,Robinson2010}, it is already expected that the BCF model outperforms the ps-BART model for the first DGP. Thus, the choice of this DGP is to determine to what extent the BCF model outperforms the ps-BART model when its key parametric assumption is correctly met.

\subsection{Evaluation Metrics and Approach}

To rigorously assess the performance of the proposed ps-BART model in comparison to the benchmark BCF model, a set of carefully selected evaluation metrics is employed based on the proposed methodological practices of \citeauthor{Souto2024} (\citeyear{Souto2024}).

The primary objective in evaluating the models is to compare the estimated treatment effect functions, denoted as $ \hat{\phi}(D) $ for ATE and $ \hat{\phi}(D,\mathbf{xi}) $ for CATE, with the true underlying treatment effect functions \( \phi(D) = \mathbb{E}[Y \mid \mathbf{X} = \mathbb{E}[\mathbf{X}], D] \) for ATE and \( \phi(D,\mathbf{xi}) = \mathbb{E}[Y \mid \mathbf{X} = \mathbf{xi}, D] \) for CATE. Needless to say, it would be computationally impossible to consider all possible values of $D$ when comparing the estimated treatment effect functions with the actual functions given the fact that $D$ is continues. Hence, both the values of $D$ considered for the model evaluation are 250 values ranging from 0 (no treatment) to the empirical maximum value of $D$ in the $j$-the simulation for each DGP.

To quantify the accuracy of the models' estimates, three standard error metrics are used: Root Mean Squared Error (RMSE), Mean Absolute Error (MAE), and Mean Absolute Percentage Error (MAPE). These metrics provide a comprehensive assessment of the distance between the true treatment effect \( \phi \) and the estimated effect \( \hat{\phi} \):
   \[
   \text{RMSE}_j = \sqrt{\frac{1}{n} \sum_{i=1}^n \left( \hat{\phi}(D_i,.) - \phi(D_i,.) \right)^2},
   \]
   where $j$ represents the $j$-th simulation of the considered DGP, \( n \) is the number of observations and $.$ in $\hat{\phi}(D_i,.)$ and $\phi(D_i,.)$ is replaced by $\mathbb{E}[\mathbf{X}]$ and $\mathbf{xi}$ when considering ATE and CATE respectively. RMSE gives greater weight to larger errors, making it sensitive to outliers and useful for evaluating the overall accuracy of the model.
   \[
   \text{MAE}_j = \frac{1}{n} \sum_{i=1}^n \left| \hat{\phi}(D_i,.) - \phi(D_i,.) \right|,
   \]
   which measures the average magnitude of the errors without considering their direction. MAE is a straightforward measure of prediction accuracy and is less sensitive to outliers than RMSE.
   \[
   \text{MAPE}_j = \frac{100\%}{n} \sum_{i=1}^n \left| \frac{\hat{\phi}(D_i,.) - \phi(D_i,.)}{\phi(D_i,.)} \right|,
   \]
   where MAPE expresses the error as a percentage, providing an intuitive measure of relative prediction accuracy, particularly useful when the scales of $\phi(D_i,.)$ vary, which is the case here given the different DGPs used in this study.

In addition to point estimates, it is essential to evaluate the models' ability to provide reliable probabilistic estimates of ATE and CATE, particularly in terms of confidence intervals. Two key metrics are used for this purpose:

1. Coverage is defined as the proportion of times the true treatment effect \( \phi(D) \) falls within the estimated confidence interval \( [\hat{\phi}_{\text{lower}}(D), \hat{\phi}_{\text{upper}}(D)] \), which in this study is the confidence interval of level $95\%$. Mathematically:
   \[
   \text{Cover}_j = \frac{1}{n} \sum_{i=1}^n \mathbb{I} \left( \phi(D_i) \in [\hat{\phi}_{\text{lower}}(D_i), \hat{\phi}_{\text{upper}}(D_i)] \right),
   \]
   where \( \mathbb{I}(\cdot) \) is the indicator function. Ideal coverage would match the nominal confidence level, in this study 95\%.

2. The length of the confidence interval provides information about the precision of the estimate:
   \[
   \text{Len}_j = \frac{1}{n} \sum_{i=1}^n \left( \hat{\phi}_{\text{upper}}(D_i) - \hat{\phi}_{\text{lower}}(D_i) \right).
   \]
   Shorter intervals indicate more precise estimates, although this must be balanced against the coverage to avoid overly narrow intervals that exclude the true effect.

Besides this two key (and commonly used) metrics for uncertainty estimation evaluation, we employ the Squared Error for Coverage (SEC) and Absolute Error for Coverage (AEC) as additional metrics given their complementary power for model uncertainty estimation evaluation \parencite{Souto2024}:
\[
\text{SEC}_j = \frac{1}{n} \sum_{i=1}^{n} ( \text{Cover}_j  - \alpha )^2
\]
\[
\text{AEC}_j = \frac{1}{n} \sum_{i=1}^{n} |\text{Cover}_j  - \alpha|
\]

Furthermore, as proposed by \citeauthor{Souto2024} (\citeyear{Souto2024}) not only the average values of the chosen evaluation metrics is presented for each DGP (as commonly done in the literature), but also their respective standard deviation values. Additionally, statistical tests are performed to estimate whether the differences in average and standard deviation of the selected metrics between the ps-BART model and the benchmark model are statistically significant or not. The choice for these tests follow the algorithm proposed by \citeauthor{Souto2024} (\citeyear{Souto2024}), which considers the statistical properties of every metric results for each DGP to choose the proper set of statistical tests. The considered statistical tests for the algorithm are: 1. T-test, 2. Mann-Whitney U test, 3. Brown-Forsythe test, and 4. Fligner-Policello test for the average values of each metric and 1. F-test, 2. Levene's test, and 3. Brown-Forsythe test for standard deviations of every metric.

\section{Results and Analysis} \label{Sec4}

The analysis of the results present in this section has the aim to be concise and to-the-point to ensure the sparsity of this paper. Hence, the results of individual metrics will not be discussed in the details but rather their joint results with the division of point-wise estimation performance measures and uncertainty estimation performance measures. It is worth noting that when we say below that a DGP is (non)linear, we mean that the DGP has a (non)linear relationship between the treatment and outcome 

\subsection{1st Set of DGPs}

\subsubsection{$\alpha=1$}

Table \ref{Table1} presents the results of all metrics for the 1st Set of DGPs with $\alpha=1$. Starting with $n=100$, considering the point-wise estimation performance metrics, both models have a similar performance, with the BCF model being more robust in ATE function estimation than the ps-BART model. Moving to the uncertainty estimation performance measures, it can be seen that the ps-BART model is considerably better in ATE function estimation and slightly better in CATE function estimation. Now moving to $n=250$ and $n=500$, the ps-BART model becomes clearly superior in both point-wise estimation and uncertainty estimation, while remaining less robust than the BCF model for ATE function point-wise estimation.

\begin{figure}[H]
\centering
\begin{subfigure}{0.45\textwidth}
    \centering
     \includegraphics[scale=0.2]{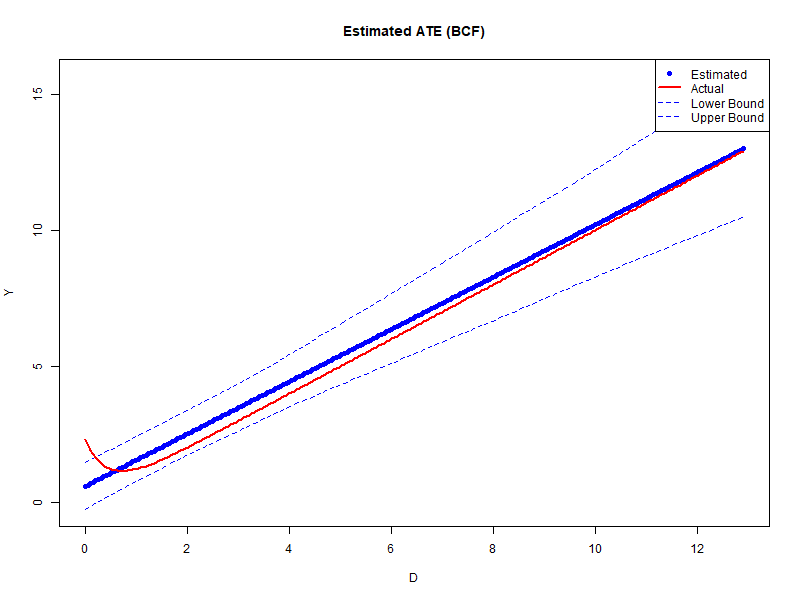}
\end{subfigure}
\hfill
\begin{subfigure}{0.45\textwidth}
    \centering
     \includegraphics[scale=0.2]{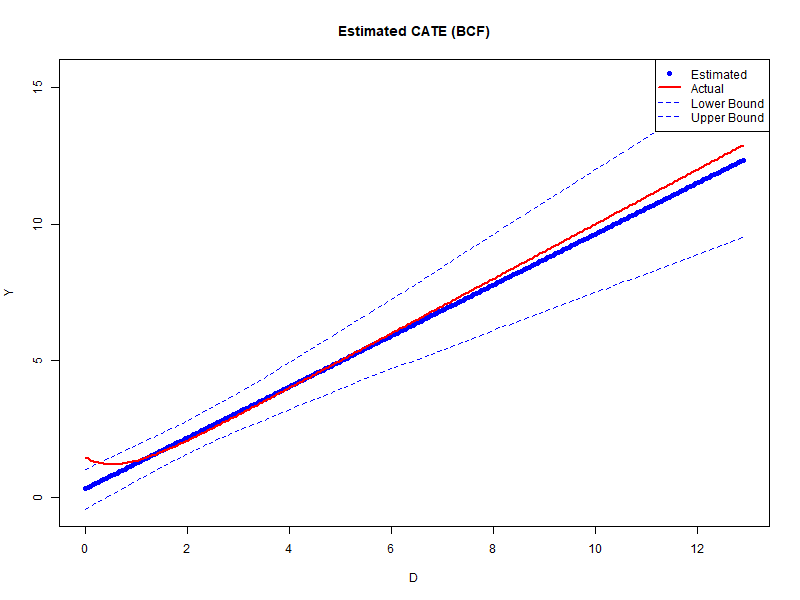}
\end{subfigure}
\caption{BCF ATE and CATE Functions Estimation for N=100 (for CATE, an Example of a Random $\mathbf{x_i}$ of a Random Simulation is used)}\label{Figure1}
\end{figure}

\begin{figure}[H]
\centering
\begin{subfigure}{0.45\textwidth}
    \centering
    \includegraphics[scale=0.2]{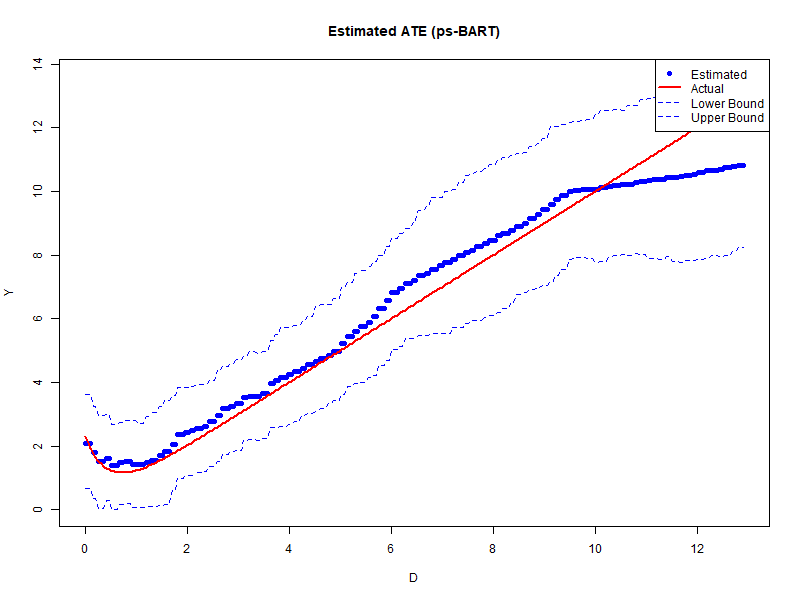}
\end{subfigure}
\hfill
\begin{subfigure}{0.45\textwidth}
    \centering
     \includegraphics[scale=0.2]{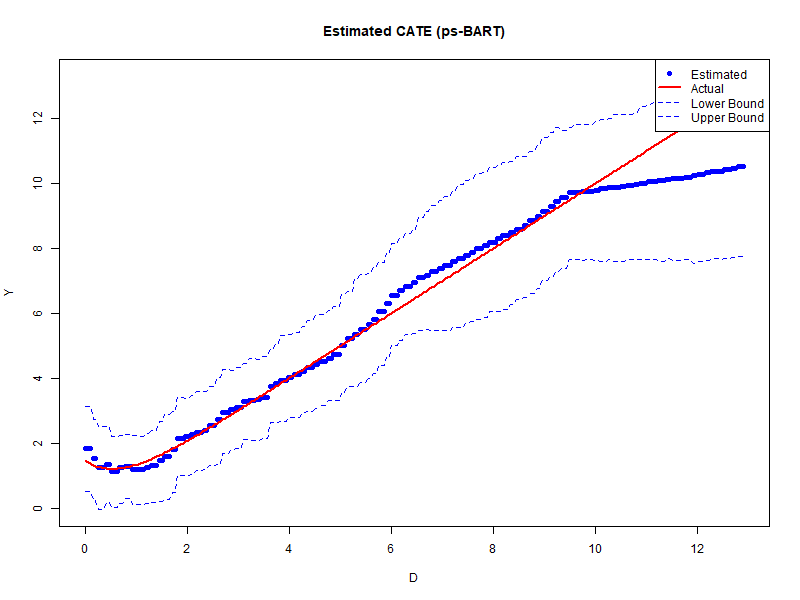}
\end{subfigure}
\caption{ps-BART ATE and CATE Functions Estimation for N=100 (for CATE, an Example of a Random $\mathbf{x_i}$ of a Random Simulation is used)}\label{Figure2}
\end{figure}

\begin{figure}[H]
\centering
\begin{subfigure}{0.45\textwidth}
    \centering
    \includegraphics[scale=0.2]{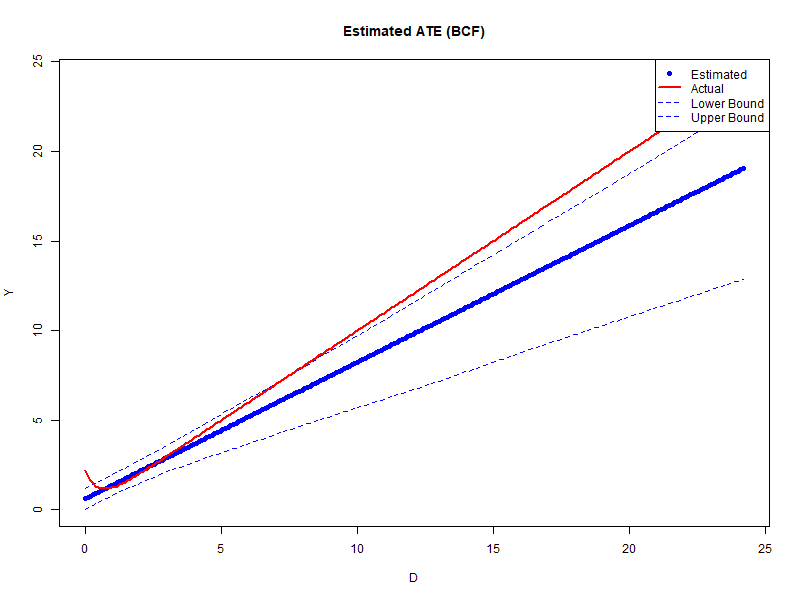}
\end{subfigure}
\hfill
\begin{subfigure}{0.45\textwidth}
    \centering
    \includegraphics[scale=0.2]{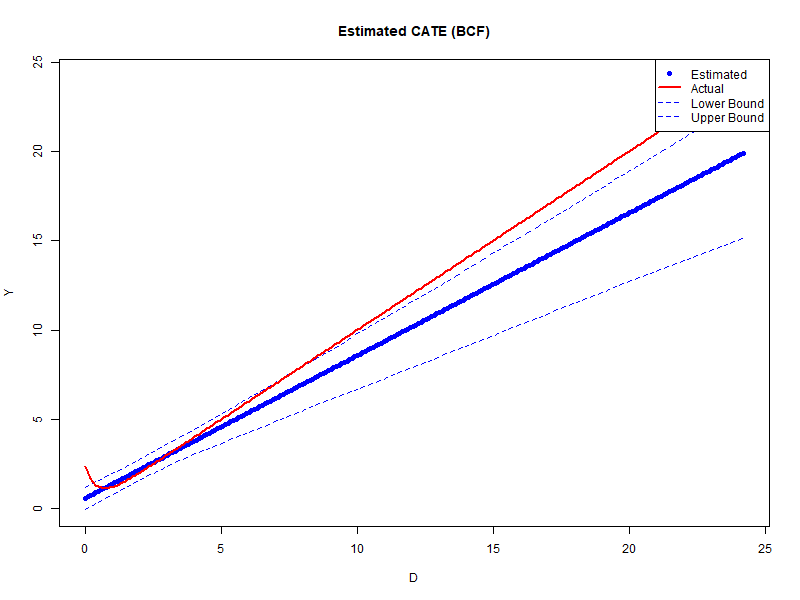}
\end{subfigure}
\caption{BCF ATE and CATE Functions Estimation for N=250 (for CATE, an Example of a Random $\mathbf{x_i}$ of a Random Simulation is used)}\label{Figure3}
\end{figure}

\begin{figure}[H]
\centering
\begin{subfigure}{0.45\textwidth}
    \centering
    \includegraphics[scale=0.2]{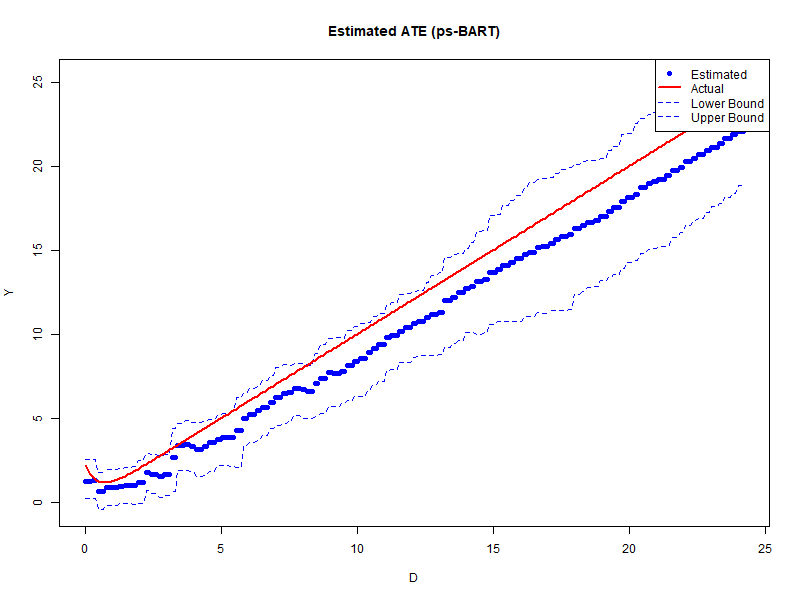}
\end{subfigure}
\hfill
\begin{subfigure}{0.45\textwidth}
    \centering
    \includegraphics[scale=0.2]{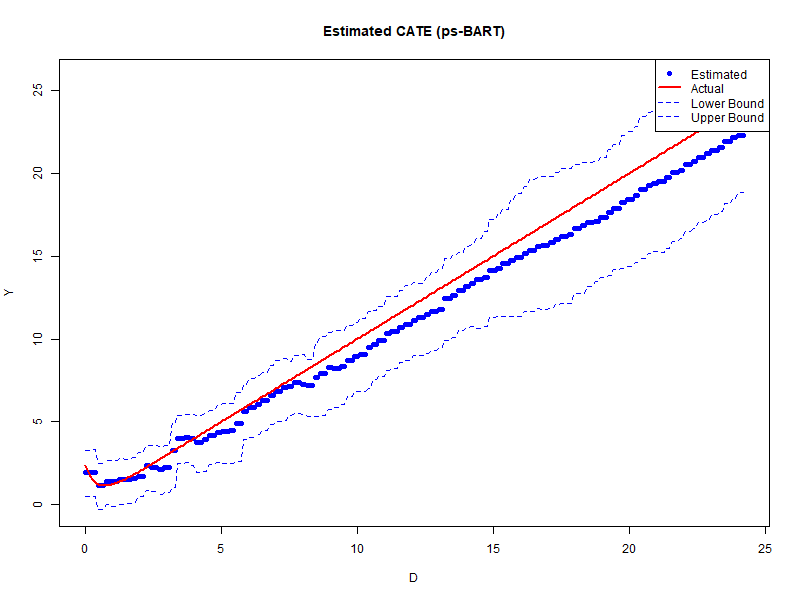}
\end{subfigure}
\caption{ps-BART ATE and CATE Functions Estimation for N=250 (for CATE, an Example of a Random $\mathbf{x_i}$ of a Random Simulation is used)}\label{Figure4}
\end{figure}

\begin{figure}[H]
\centering
\begin{subfigure}{0.45\textwidth}
    \centering
    \includegraphics[scale=0.2]{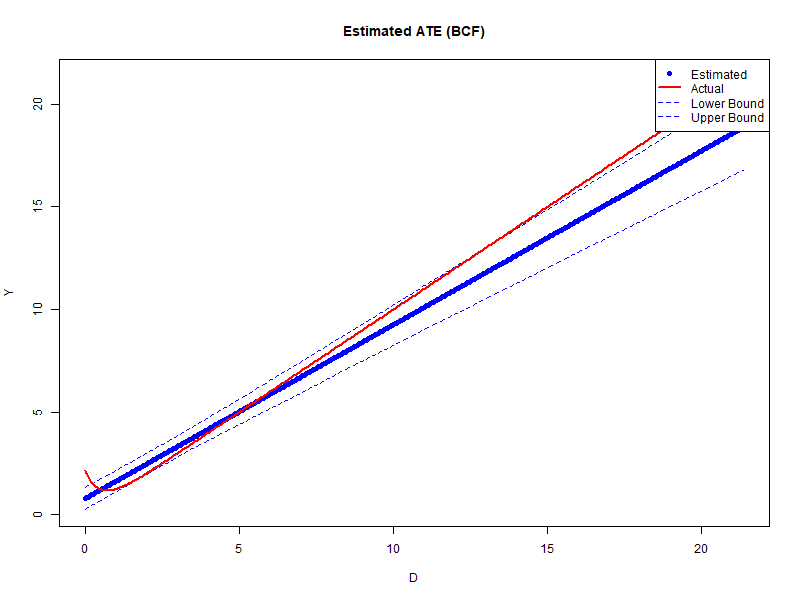}
\end{subfigure}
\hfill
\begin{subfigure}{0.45\textwidth}
    \centering
    \includegraphics[scale=0.2]{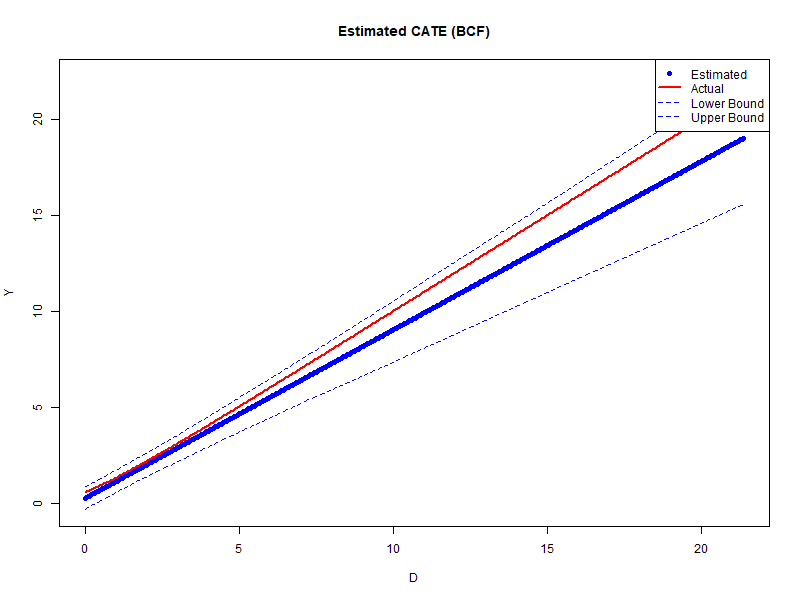}
\end{subfigure}
\caption{BCF ATE and CATE Functions Estimation for N=500 (for CATE, an Example of a Random $\mathbf{x_i}$ of a Random Simulation is used)}\label{Figure5}
\end{figure}

\begin{figure}[H]
\centering
\begin{subfigure}{0.45\textwidth}
    \centering
    \includegraphics[scale=0.2]{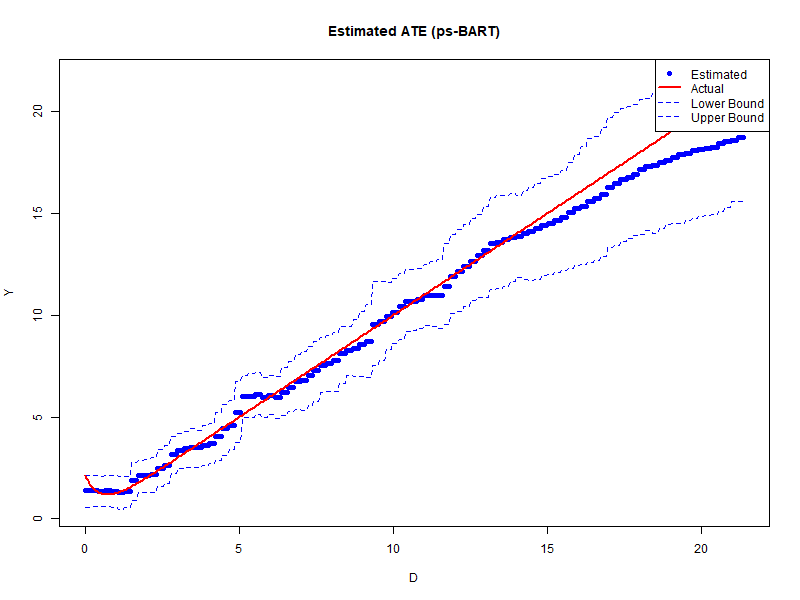}
\end{subfigure}
\hfill
\begin{subfigure}{0.45\textwidth}
    \centering
    \includegraphics[scale=0.2]{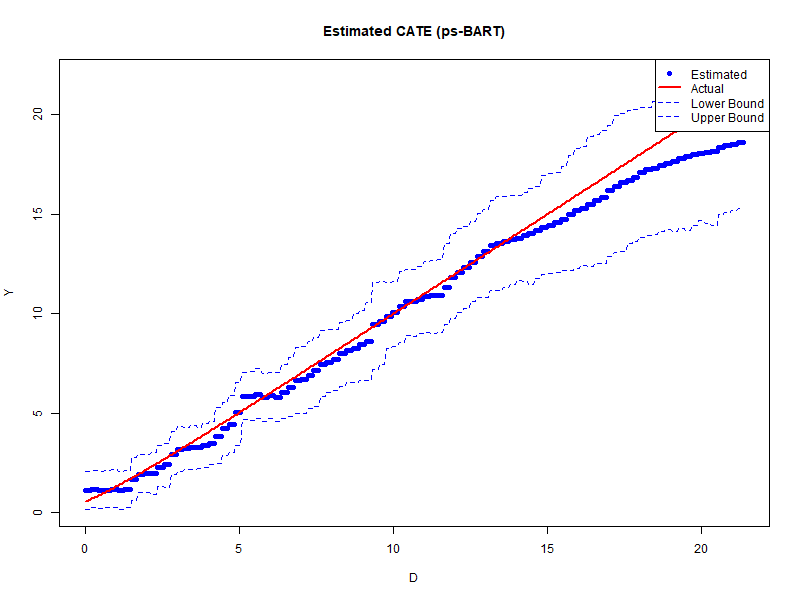}
\end{subfigure}
\caption{ps-BART ATE and CATE Functions Estimation for N=500 (for CATE, an Example of a Random $\mathbf{x_i}$ of a Random Simulation is used)}\label{Figure6}
\end{figure}

Figures \ref{Figure1}, \ref{Figure2}, \ref{Figure3}, \ref{Figure4}, \ref{Figure5}, and \ref{Figure6} graphically illustrate the results of Table \ref{Table1}. The misspecification of the BCF model for the underlying DGPs can be visually seen in these figures.

The statistical tests results for $n=100$, $n=250$, and $n=500$ can be seen in Tables \ref{Table2}, \ref{Table3}, and \ref{Table4} respectively. For $n=100$, we can infer that the BCF model is statistically significantly better and more robust than the ps-BART model for ATE function point-wise estimation, while being having a similar performance for CATE function point-wise estimation. Regarding uncertainty estimation, the ps-BART model is statistically significantly superior to the BCF model. For $n=250$ and $n=500$, it can be concluded that the superiority of the proposed model is statistically significant, while remaining less robust than the BCF model for ATE function point-wise estimation.

\begin{table}[H]
\centering
\caption{Metric Results}\label{Table1}
\begin{tabular}{llcc}
\hline
$n$ & \textbf{Metric} & \textbf{BCF (Mean $\pm$ SD)} & \textbf{ps-BART (Mean $\pm$ SD)} \\
\hline
100 & RMSE\(_{ATE}\) & $0.630 \pm 0.156$ & $0.630 \pm 0.266$ \\
& MAE\(_{ATE}\)  & $0.473 \pm 0.108$ & $0.477 \pm 0.258$ \\
& MAPE\(_{ATE}\) & $0.241 \pm 0.058$ & $0.242 \pm 0.145$ \\
& Len\(_{ATE}\) & $1.917 \pm 0.338$ & $3.007 \pm 0.584$ \\
& Cover\(_{ATE}\) & $0.869 \pm 0.092$ & $0.984 \pm 0.043$ \\
& RMSE\(_{CATE}\) & $1.374 \pm 0.783$ & $1.391 \pm 0.875$ \\
& MAE\(_{CATE}\)  & $1.132 \pm 0.651$ & $1.103 \pm 0.681$ \\
& MAPE\(_{CATE}\) & $0.149 \pm 0.043$ & $0.156 \pm 0.039$ \\
& Len\(_{CATE}\) & $5.392 \pm 2.187$ & $5.243 \pm 2.005$ \\
& Cover\(_{CATE}\) & $0.936 \pm 0.084$ & $0.954 \pm 0.069$ \\
& SEC\(_{ATE}\)  & $0.015 \pm 0.033$ & $0.003 \pm 0.006$ \\
& AEC\(_{ATE}\)  & $0.095 \pm 0.079$ & $0.048 \pm 0.026$ \\
& SEC\(_{CATE}\)  & $0.007 \pm 0.021$ & $0.005 \pm 0.010$ \\
& AEC\(_{CATE}\)  & $0.052 \pm 0.045$ & $0.052 \pm 0.045$ \\
\hline
250 & RMSE\(_{ATE}\) & $0.567 \pm 0.074$ & $0.484 \pm 0.169$ \\
& MAE\(_{ATE}\)  & $0.445 \pm 0.047$ & $0.382 \pm 0.172$ \\
& MAPE\(_{ATE}\) & $0.235 \pm 0.026$ & $0.201 \pm 0.101$ \\
& Len\(_{ATE}\) & $1.298 \pm 0.169$ & $2.412 \pm 0.279$ \\
& Cover\(_{ATE}\) & $0.753 \pm 0.102$ & $0.980 \pm 0.066$ \\
& RMSE\(_{CATE}\) & $1.453 \pm 0.713$ & $1.175 \pm 0.637$ \\
& MAE\(_{CATE}\)  & $1.188 \pm 0.595$ & $0.901 \pm 0.462$ \\
& MAPE\(_{CATE}\) & $0.119 \pm 0.032$ & $0.108 \pm 0.020$ \\
& Len\(_{CATE}\) & $3.905 \pm 1.153$ & $5.163 \pm 1.603$ \\
& Cover\(_{CATE}\) & $0.825 \pm 0.156$ & $0.980 \pm 0.034$ \\
& SEC\(_{ATE}\)  & $0.049 \pm 0.051$ & $0.005 \pm 0.020$ \\
& AEC\(_{ATE}\)  & $0.197 \pm 0.102$ & $0.055 \pm 0.048$ \\
& SEC\(_{CATE}\)  & $0.040 \pm 0.066$ & $0.002 \pm 0.003$ \\
& AEC\(_{CATE}\)  & $0.136 \pm 0.146$ & $0.041 \pm 0.019$ \\
\hline
500 & RMSE\(_{ATE}\) & $0.572 \pm 0.073$ & $0.397 \pm 0.157$ \\
& MAE\(_{ATE}\)  & $0.433 \pm 0.032$ & $0.309 \pm 0.134$ \\
& MAPE\(_{ATE}\) & $0.228 \pm 0.017$ & $0.160 \pm 0.079$ \\
& Len\(_{ATE}\) & $1.026 \pm 0.114$ & $1.810 \pm 0.255$ \\
& Cover\(_{ATE}\) & $0.682 \pm 0.118$ & $0.962 \pm 0.067$ \\
& RMSE\(_{CATE}\) & $1.684 \pm 0.922$ & $1.179 \pm 1.373$ \\
& MAE\(_{CATE}\)  & $1.381 \pm 0.781$ & $0.918 \pm 1.096$ \\
& MAPE\(_{CATE}\) & $0.105 \pm 0.021$ & $0.084 \pm 0.023$ \\
& Len\(_{CATE}\) & $3.343 \pm 1.396$ & $5.653 \pm 3.129$ \\
& Cover\(_{CATE}\) & $0.642 \pm 0.148$ & $0.987 \pm 0.025$ \\
& SEC\(_{ATE}\)  & $0.086 \pm 0.079$ & $0.005 \pm 0.011$ \\
& AEC\(_{ATE}\)  & $0.268 \pm 0.118$ & $0.054 \pm 0.040$ \\
& SEC\(_{CATE}\)  & $0.117 \pm 0.088$ & $0.002 \pm 0.003$ \\
& AEC\(_{CATE}\)  & $0.308 \pm 0.148$ & $0.042 \pm 0.015$ \\
\hline
\end{tabular}
\end{table}

\begin{landscape}
\begin{table}[h!]
\centering
\caption{Statistical Test Results: p-values for Different Metrics (n=100)}\label{Table2}
\begin{tabular}{lccccc}
\hline
\textbf{Metric} & \textbf{Fligner-Policello Test} & \textbf{Mann-Whitney U Test} & \textbf{Kruskal-Wallis H Test} & \textbf{Levene's Test} & \textbf{Brown-Forsythe Test} \\
\hline
RMSE\(_{ATE}\)  & $0.1338$ & \textit{N/A} & \textit{N/A} & $0.000109$ & $0.000596$ \\
MAE\(_{ATE}\)   & $0.0220$ & \textit{N/A} & \textit{N/A} & $2.65 \times 10^{-7}$ & $1.99 \times 10^{-5}$ \\
MAPE\(_{ATE}\)  & $0.0231$ & \textit{N/A} & \textit{N/A} & $2.77 \times 10^{-9}$ & $6.74 \times 10^{-7}$ \\
Len\(_{ATE}\)   & \textit{N/A} & $4.57 \times 10^{-30}$ & $4.50 \times 10^{-30}$ & $2.58 \times 10^{-6}$ & $4.14 \times 10^{-6}$ \\
RMSE\(_{CATE}\) & $0.6647$ & \textit{N/A} & \textit{N/A} & $0.2794$ & $0.4782$ \\
MAE\(_{CATE}\)  & \textit{N/A} & $0.4082$ & $0.4075$ & $0.5982$ & $0.7866$ \\
MAPE\(_{CATE}\) & \textit{N/A} & $0.1430$ & $0.1426$ & $0.4742$ & $0.3471$ \\
Len\(_{CATE}\)  & $0.7898$ & \textit{N/A} & \textit{N/A} & $0.2908$ & $0.4035$ \\
SEC\(_{ATE}\)   & \textit{N/A} & $3.31 \times 10^{-5}$ & $3.29 \times 10^{-5}$ & $1.18 \times 10^{-5}$ & $0.000282$ \\
AEC\(_{ATE}\)   & $8.01 \times 10^{-5}$ & \textit{N/A} & \textit{N/A} & $5.04 \times 10^{-13}$ & $5.08 \times 10^{-12}$ \\
SEC\(_{CATE}\)  & $0.00039$ & \textit{N/A} & \textit{N/A} & $0.0112$ & $0.2046$ \\
AEC\(_{CATE}\)  & $0.00039$ & \textit{N/A} & \textit{N/A} & $0.0434$ & $0.2495$ \\
\hline
\end{tabular}
\end{table}
\end{landscape}

\begin{landscape}
\begin{table}[h!]
\centering
\caption{Statistical Test Results: p-values for Different Metrics (n=250)}\label{Table3}
\begin{tabular}{lccccc}
\hline
\textbf{Metric} & \textbf{Fligner-Policello Test} & \textbf{Mann-Whitney U Test} & \textbf{Kruskal-Wallis H Test} & \textbf{Levene's Test} & \textbf{Brown-Forsythe Test} \\
\hline
RMSE\(_{ATE}\)  & $2.91 \times 10^{-9}$ & \textit{N/A} & \textit{N/A} & $3.91 \times 10^{-10}$ & $1.29 \times 10^{-7}$ \\
MAE\(_{ATE}\)   & $2.99 \times 10^{-8}$ & \textit{N/A} & \textit{N/A} & $5.50 \times 10^{-14}$ & $2.33 \times 10^{-9}$ \\
MAPE\(_{ATE}\)  & $3.16 \times 10^{-8}$ & \textit{N/A} & \textit{N/A} & $7.67 \times 10^{-15}$ & $6.15 \times 10^{-11}$ \\
Len\(_{ATE}\)   & $0$ & \textit{N/A} & \textit{N/A} & $1.72 \times 10^{-5}$ & $1.78 \times 10^{-5}$ \\
RMSE\(_{CATE}\) & \textit{N/A} & $0.0020$ & $0.0020$ & $0.1332$ & $0.1472$ \\
MAE\(_{CATE}\)  & $4.61 \times 10^{-5}$ & \textit{N/A} & \textit{N/A} & $0.0128$ & $0.0229$ \\
MAPE\(_{CATE}\) & $0.0201$ & \textit{N/A} & \textit{N/A} & $9.81 \times 10^{-5}$ & $0.00021$ \\
Len\(_{CATE}\)  & $5.20 \times 10^{-15}$ & \textit{N/A} & \textit{N/A} & $0.0136$ & $0.0380$ \\
SEC\(_{ATE}\)   & $8.13 \times 10^{-178}$ & \textit{N/A} & \textit{N/A} & $1.36 \times 10^{-16}$ & $3.38 \times 10^{-10}$ \\
AEC\(_{ATE}\)   & $8.13 \times 10^{-178}$ & \textit{N/A} & \textit{N/A} & $1.76 \times 10^{-17}$ & $1.48 \times 10^{-13}$ \\
SEC\(_{CATE}\)  & $0.04598$ & \textit{N/A} & \textit{N/A} & $8.68 \times 10^{-20}$ & $7.48 \times 10^{-9}$ \\
AEC\(_{CATE}\)  & $0.04598$ & \textit{N/A} & \textit{N/A} & $4.08 \times 10^{-30}$ & $2.18 \times 10^{-18}$ \\
\hline
\end{tabular}
\end{table}
\end{landscape}

\begin{landscape}
\begin{table}[h!]
\centering
\caption{Statistical Test Results: p-values for Different Metrics (n=500)}\label{Table4}
\begin{tabular}{lccccc}
\hline
\textbf{Metric} & \textbf{Fligner-Policello Test} & \textbf{Mann-Whitney U Test} & \textbf{Kruskal-Wallis H Test} & \textbf{Levene's Test} & \textbf{Brown-Forsythe Test} \\
\hline
RMSE\(_{ATE}\)  & $3.72 \times 10^{-30}$ & \textit{N/A} & \textit{N/A} & $8.56 \times 10^{-8}$ & $2.04 \times 10^{-5}$ \\
MAE\(_{ATE}\)   & $1.00 \times 10^{-18}$ & \textit{N/A} & \textit{N/A} & $6.27 \times 10^{-20}$ & $1.29 \times 10^{-11}$ \\
MAPE\(_{ATE}\)  & $5.35 \times 10^{-15}$ & \textit{N/A} & \textit{N/A} & $4.10 \times 10^{-22}$ & $1.56 \times 10^{-12}$ \\
Len\(_{ATE}\)   & $0$ & \textit{N/A} & \textit{N/A} & $2.95 \times 10^{-5}$ & $3.16 \times 10^{-5}$ \\
RMSE\(_{CATE}\) & \textit{N/A} & $2.07 \times 10^{-11}$ & $2.05 \times 10^{-11}$ & $0.8922$ & $0.7538$ \\
MAE\(_{CATE}\)  & \textit{N/A} & $3.99 \times 10^{-13}$ & $3.96 \times 10^{-13}$ & $0.6488$ & $0.5230$ \\
MAPE\(_{CATE}\) & \textit{N/A} & $4.14 \times 10^{-13}$ & $4.11 \times 10^{-13}$ & $0.3527$ & $0.2743$ \\
Len\(_{CATE}\)  & $3.99 \times 10^{-40}$ & \textit{N/A} & \textit{N/A} & $5.39 \times 10^{-4}$ & $0.00356$ \\
SEC\(_{ATE}\)   & $0$ & \textit{N/A} & \textit{N/A} & $9.15 \times 10^{-23}$ & $1.08 \times 10^{-11}$ \\
AEC\(_{ATE}\)   & $0$ & \textit{N/A} & \textit{N/A} & $8.30 \times 10^{-19}$ & $3.17 \times 10^{-12}$ \\
SEC\(_{CATE}\)  & $1.05 \times 10^{-57}$ & \textit{N/A} & \textit{N/A} & $1.05 \times 10^{-27}$ & $2.43 \times 10^{-25}$ \\
AEC\(_{CATE}\)  & $1.05 \times 10^{-57}$ & \textit{N/A} & \textit{N/A} & $4.25 \times 10^{-25}$ & $6.36 \times 10^{-24}$ \\
\hline
\end{tabular}
\end{table}
\end{landscape}

\subsubsection{$\alpha=2$}

The simulation results for $\alpha=2$ can be found in \ref{Table5}. For all considered $n$ values, regarding the point-wise estimation performance measures, the ps-BART model is now superior to the BCF model, though the latter remains more robust for ATE function estimation. Moving to the uncertainty estimation performance metrics, it can be seen that the ps-BART model is considerably better in both ATE and CATE function estimation. Such results were already expected as by increasing the $\alpha$ value, we increase the nonlinearity relationship between the treatment and the outcome, increasing the misspecification level of the BCF model and thus favoring the fully-nonparametric ps-BART model.

\begin{figure}[H]
\centering
\begin{subfigure}{0.45\textwidth}
    \centering
     \includegraphics[scale=0.2]{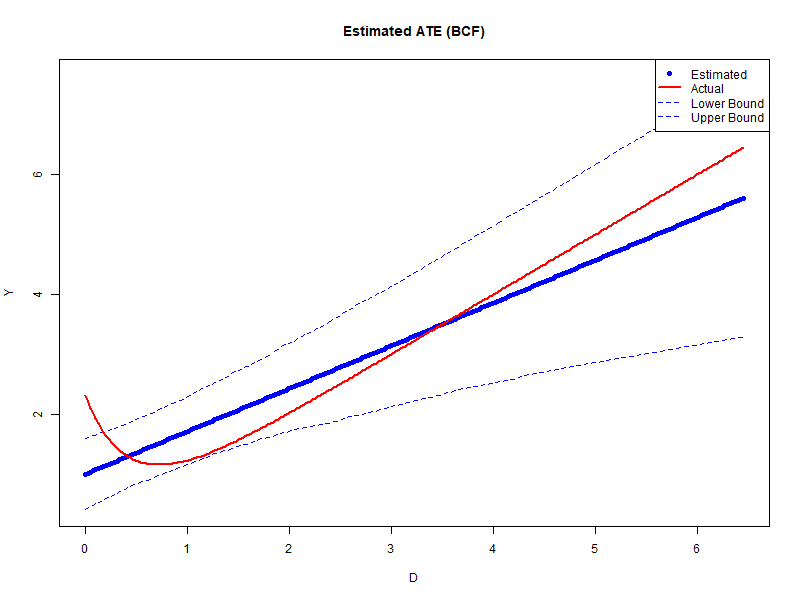}
\end{subfigure}
\hfill
\begin{subfigure}{0.45\textwidth}
    \centering
     \includegraphics[scale=0.2]{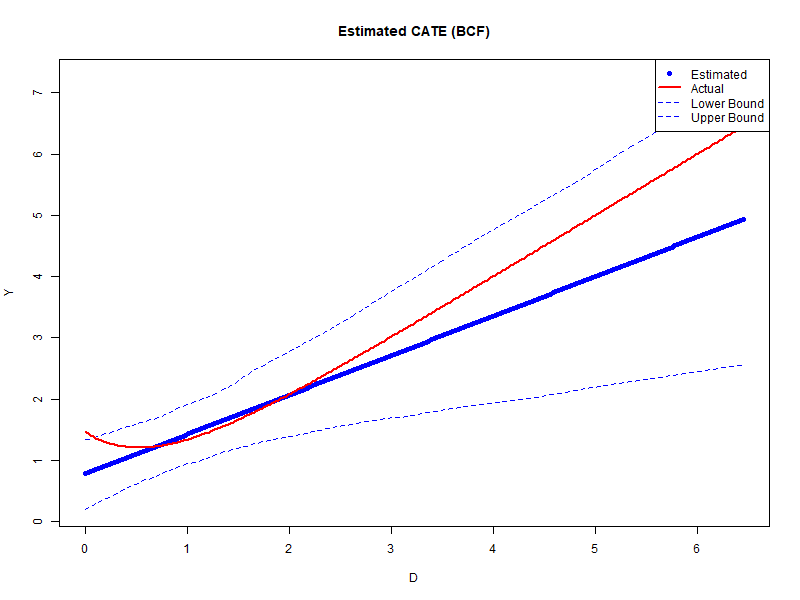}
\end{subfigure}
\caption{BCF ATE and CATE Functions Estimation for N=100 (for CATE, an Example of a Random $\mathbf{x_i}$ of a Random Simulation is used)}\label{Figure7}
\end{figure}

\begin{figure}[H]
\centering
\begin{subfigure}{0.45\textwidth}
    \centering
    \includegraphics[scale=0.2]{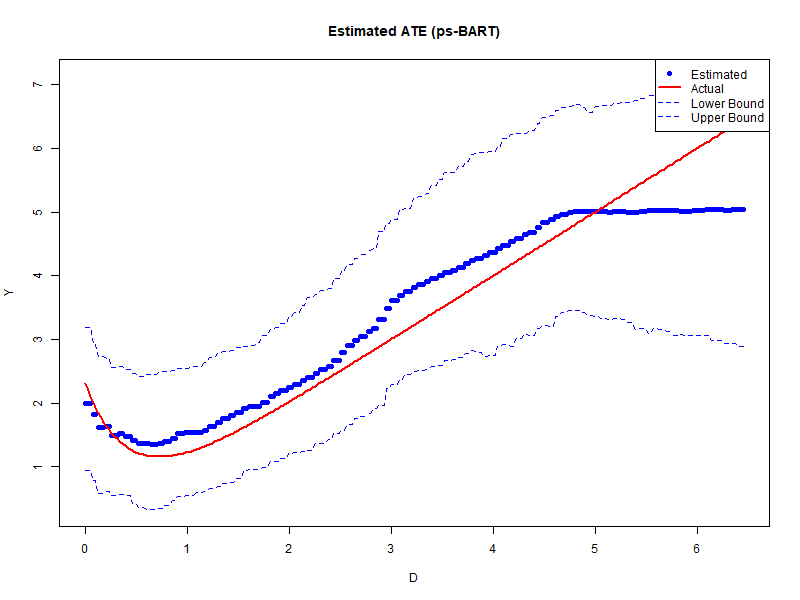}
\end{subfigure}
\hfill
\begin{subfigure}{0.45\textwidth}
    \centering
     \includegraphics[scale=0.2]{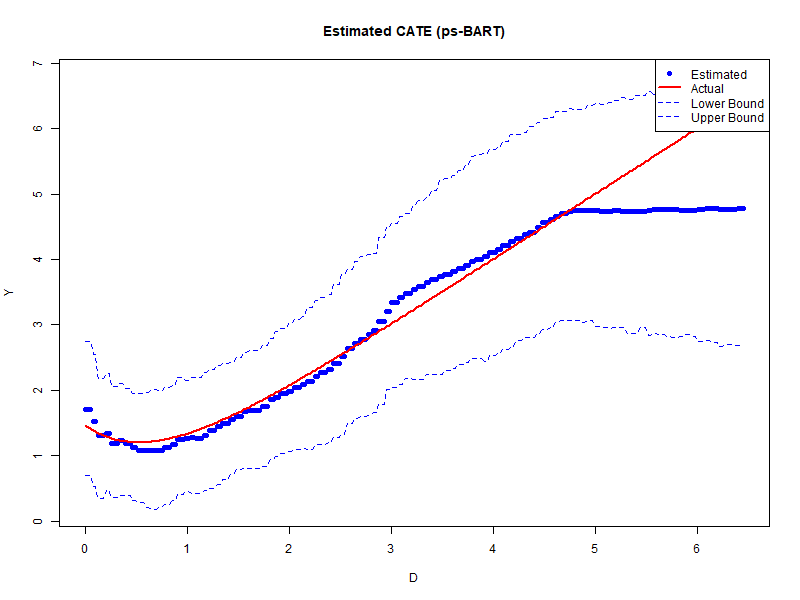}
\end{subfigure}
\caption{ps-BART ATE and CATE Functions Estimation for N=100 (for CATE, an Example of a Random $\mathbf{x_i}$ of a Random Simulation is used)}\label{Figure8}
\end{figure}

\begin{figure}[H]
\centering
\begin{subfigure}{0.45\textwidth}
    \centering
    \includegraphics[scale=0.2]{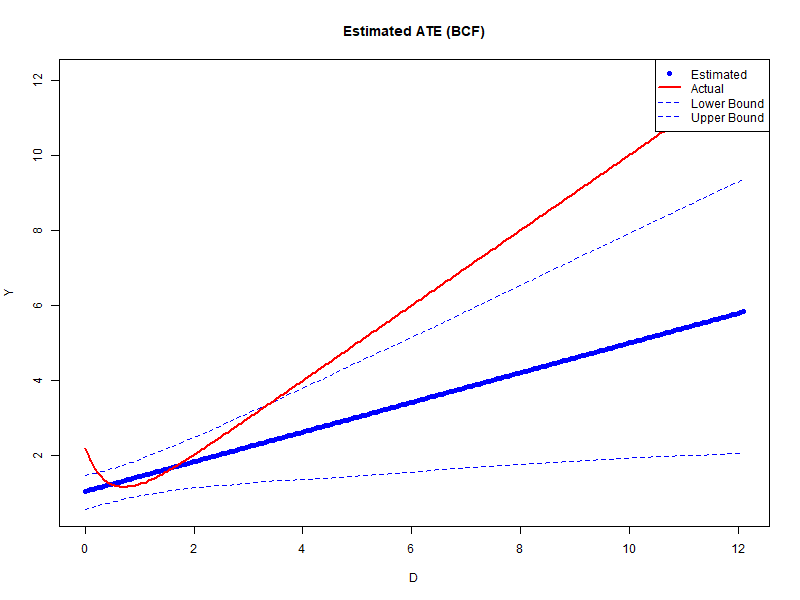}
\end{subfigure}
\hfill
\begin{subfigure}{0.45\textwidth}
    \centering
    \includegraphics[scale=0.2]{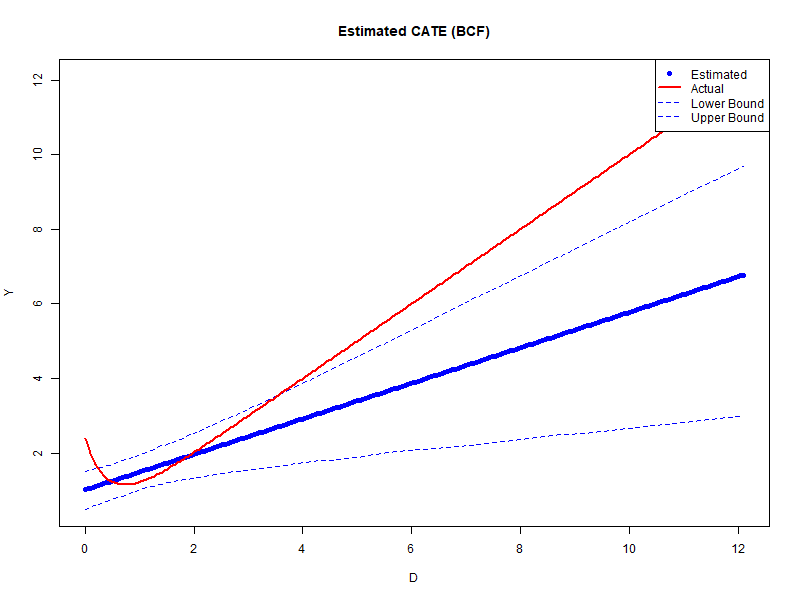}
\end{subfigure}
\caption{BCF ATE and CATE Functions Estimation for N=250 (for CATE, an Example of a Random $\mathbf{x_i}$ of a Random Simulation is used)}\label{Figure9}
\end{figure}

\begin{figure}[H]
\centering
\begin{subfigure}{0.45\textwidth}
    \centering
    \includegraphics[scale=0.2]{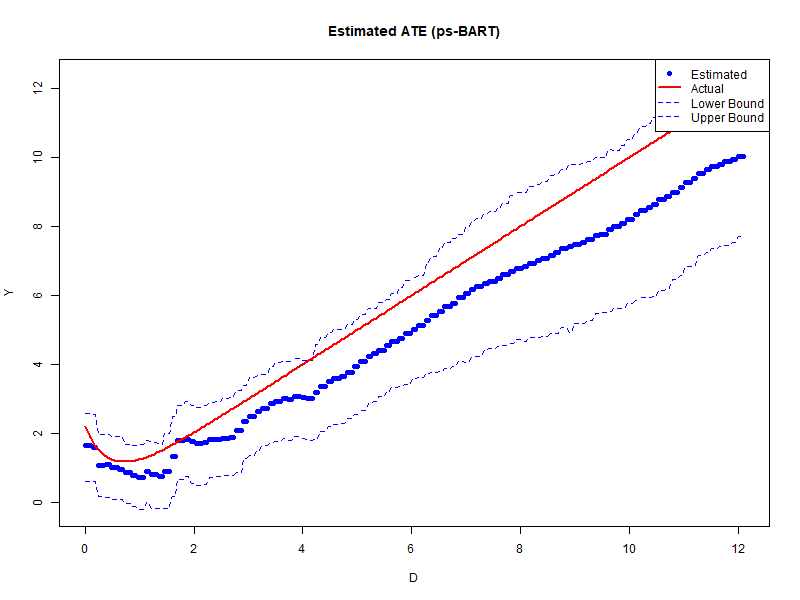}
\end{subfigure}
\hfill
\begin{subfigure}{0.45\textwidth}
    \centering
    \includegraphics[scale=0.2]{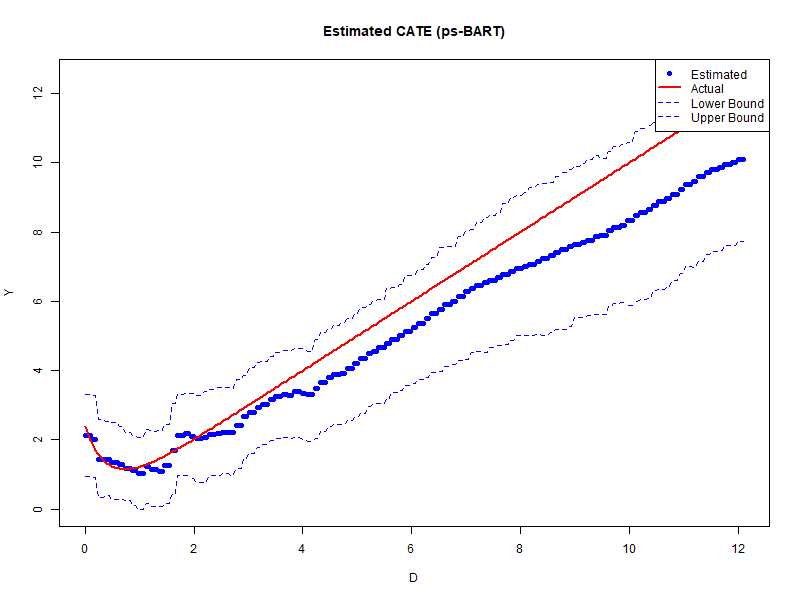}
\end{subfigure}
\caption{ps-BART ATE and CATE Functions Estimation for N=250 (for CATE, an Example of a Random $\mathbf{x_i}$ of a Random Simulation is used)}\label{Figure10}
\end{figure}

\begin{figure}[H]
\centering
\begin{subfigure}{0.45\textwidth}
    \centering
    \includegraphics[scale=0.2]{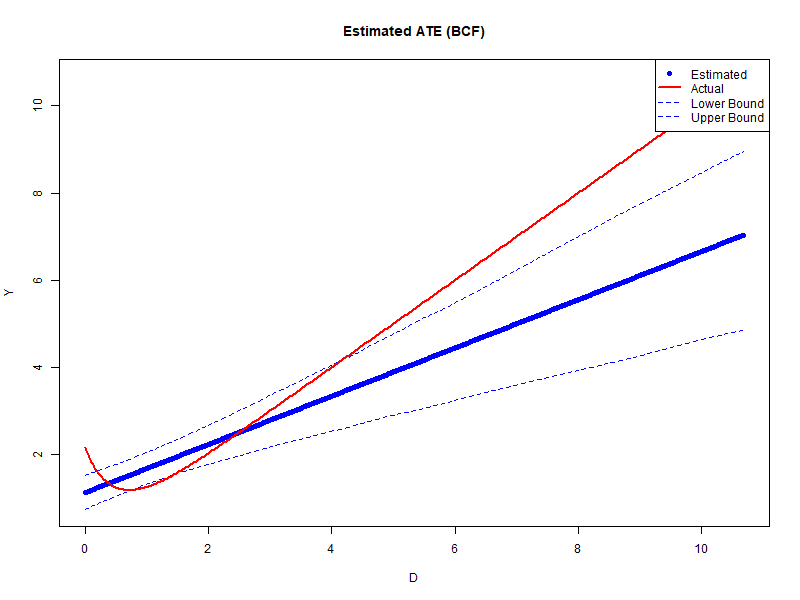}
\end{subfigure}
\hfill
\begin{subfigure}{0.45\textwidth}
    \centering
    \includegraphics[scale=0.2]{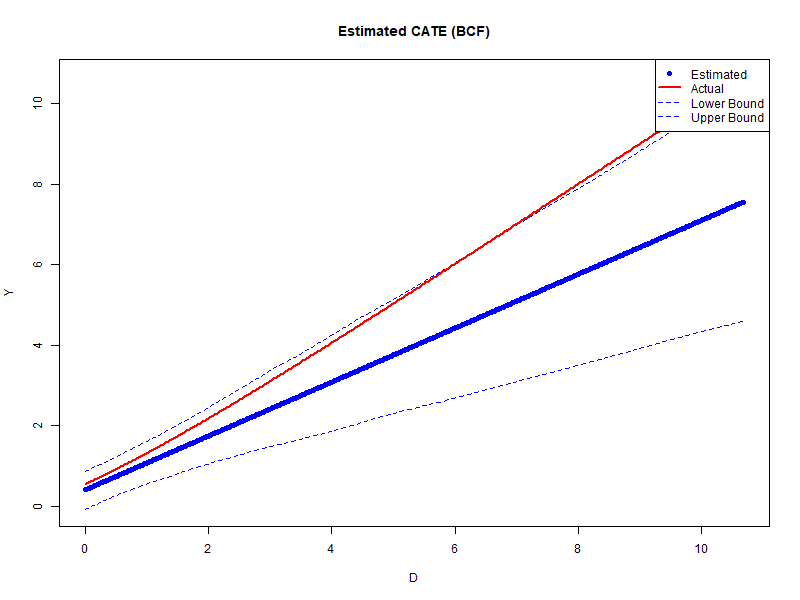}
\end{subfigure}
\caption{BCF ATE and CATE Functions Estimation for N=500 (for CATE, an Example of a Random $\mathbf{x_i}$ of a Random Simulation is used)}\label{Figure11}
\end{figure}

\begin{figure}[H]
\centering
\begin{subfigure}{0.45\textwidth}
    \centering
    \includegraphics[scale=0.2]{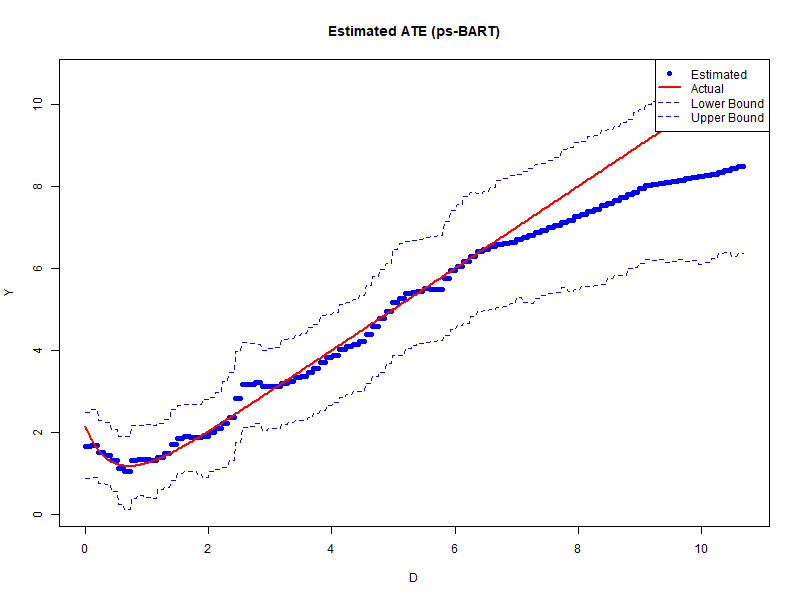}
\end{subfigure}
\hfill
\begin{subfigure}{0.45\textwidth}
    \centering
    \includegraphics[scale=0.2]{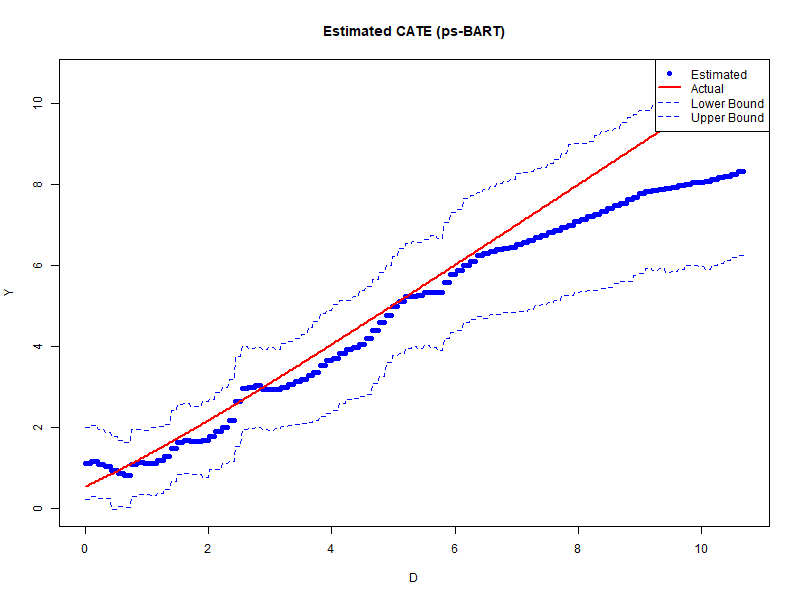}
\end{subfigure}
\caption{ps-BART ATE and CATE Functions Estimation for N=500 (for CATE, an Example of a Random $\mathbf{x_i}$ of a Random Simulation is used)}\label{Figure12}
\end{figure}

Figures \ref{Figure7}, \ref{Figure8}, \ref{Figure9}, \ref{Figure10}, \ref{Figure11}, and \ref{Figure12} graphically illustrate the results of Table \ref{Table5}. The misspecification of the BCF model for the underlying DGPs can be visually seen in these figures.

The statistical tests results for $n=100$, $n=250$, and $n=500$ can be found in Tables \ref{Table6}, \ref{Table7}, and \ref{Table8} respectively. It can be inferred that the superiority of the ps-BART model is statistically significant, while remaining less robust than the BCF model for ATE function point-wise estimation.

\begin{table}[H]
\centering
\caption{Metric Results}\label{Table5}
\begin{tabular}{llcc}
\hline
$n$ & \textbf{Metric} & \textbf{BCF (Mean $\pm$ SD)} & \textbf{ps-BART (Mean $\pm$ SD)} \\
\hline
100 & RMSE\(_{ATE}\) & $0.572 \pm 0.135$ & $0.471 \pm 0.180$ \\
& MAE\(_{ATE}\)  & $0.429 \pm 0.097$ & $0.355 \pm 0.181$ \\
& MAPE\(_{ATE}\) & $0.241 \pm 0.063$ & $0.205 \pm 0.120$ \\
& Len\(_{ATE}\) & $1.356 \pm 0.249$ & $2.155 \pm 0.391$ \\
& Cover\(_{ATE}\) & $0.788 \pm 0.109$ & $0.978 \pm 0.072$ \\
& RMSE\(_{CATE}\) & $1.289 \pm 0.748$ & $1.083 \pm 0.560$ \\
& MAE\(_{CATE}\)  & $1.040 \pm 0.616$ & $0.866 \pm 0.444$ \\
& MAPE\(_{CATE}\) & $0.231 \pm 0.069$ & $0.215 \pm 0.055$ \\
& Len\(_{CATE}\) & $3.572 \pm 1.526$ & $3.432 \pm 1.075$ \\
& Cover\(_{CATE}\) & $0.867 \pm 0.137$ & $0.900 \pm 0.116$ \\
& SEC\(_{ATE}\)  & $0.038 \pm 0.040$ & $0.006 \pm 0.031$ \\
& AEC\(_{ATE}\)  & $0.168 \pm 0.099$ & $0.053 \pm 0.057$ \\
& SEC\(_{CATE}\)  & $0.025 \pm 0.050$ & $0.016 \pm 0.029$ \\
& AEC\(_{CATE}\)  & $0.099 \pm 0.125$ & $0.087 \pm 0.091$ \\
\hline
250 & RMSE\(_{ATE}\) & $0.539 \pm 0.087$ & $0.369 \pm 0.146$ \\
& MAE\(_{ATE}\)  & $0.413 \pm 0.043$ & $0.289 \pm 0.150$ \\
& MAPE\(_{ATE}\) & $0.235 \pm 0.025$ & $0.174 \pm 0.101$ \\
& Len\(_{ATE}\) & $0.980 \pm 0.131$ & $1.836 \pm 0.226$ \\
& Cover\(_{ATE}\) & $0.640 \pm 0.094$ & $0.986 \pm 0.048$ \\
& RMSE\(_{CATE}\) & $1.669 \pm 0.765$ & $0.960 \pm 0.405$ \\
& MAE\(_{CATE}\)  & $1.348 \pm 0.634$ & $0.742 \pm 0.299$ \\
& MAPE\(_{CATE}\) & $0.231 \pm 0.059$ & $0.156 \pm 0.032$ \\
& Len\(_{CATE}\) & $3.082 \pm 0.970$ & $3.370 \pm 0.857$ \\
& Cover\(_{CATE}\) & $0.664 \pm 0.183$ & $0.943 \pm 0.071$ \\
& SEC\(_{ATE}\)  & $0.105 \pm 0.061$ & $0.004 \pm 0.011$ \\
& AEC\(_{ATE}\)  & $0.310 \pm 0.094$ & $0.051 \pm 0.032$ \\
& SEC\(_{CATE}\)  & $0.115 \pm 0.115$ & $0.005 \pm 0.013$ \\
& AEC\(_{CATE}\)  & $0.287 \pm 0.182$ & $0.051 \pm 0.049$ \\
\hline
500 & RMSE\(_{ATE}\) & $0.552 \pm 0.080$ & $0.316 \pm 0.114$ \\
& MAE\(_{ATE}\)  & $0.410 \pm 0.031$ & $0.251 \pm 0.109$ \\
& MAPE\(_{ATE}\) & $0.230 \pm 0.019$ & $0.149 \pm 0.071$ \\
& Len\(_{ATE}\) & $0.785 \pm 0.085$ & $1.572 \pm 0.170$ \\
& Cover\(_{ATE}\) & $0.531 \pm 0.097$ & $0.988 \pm 0.036$ \\
& RMSE\(_{CATE}\) & $2.108 \pm 1.080$ & $0.914 \pm 0.715$ \\
& MAE\(_{CATE}\)  & $1.714 \pm 0.912$ & $0.712 \pm 0.578$ \\
& MAPE\(_{CATE}\) & $0.227 \pm 0.043$ & $0.123 \pm 0.028$ \\
& Len\(_{CATE}\) & $2.971 \pm 1.162$ & $3.662 \pm 1.678$ \\
& Cover\(_{CATE}\) & $0.476 \pm 0.137$ & $0.970 \pm 0.038$ \\
& SEC\(_{ATE}\)  & $0.185 \pm 0.082$ & $0.003 \pm 0.004$ \\
& AEC\(_{ATE}\)  & $0.419 \pm 0.097$ & $0.048 \pm 0.021$ \\
& SEC\(_{CATE}\)  & $0.243 \pm 0.117$ & $0.002 \pm 0.003$ \\
& AEC\(_{CATE}\)  & $0.474 \pm 0.137$ & $0.037 \pm 0.022$ \\
\hline
\end{tabular}
\end{table}

\begin{landscape}
\begin{table}[h!]
\centering
\caption{Statistical Test Results: p-values for Different Metrics (n=100)}\label{Table6}
\begin{tabular}{lccccc}
\hline
\textbf{Metric} & \textbf{Fligner-Policello Test} & \textbf{Mann-Whitney U Test} & \textbf{Kruskal-Wallis H Test} & \textbf{Levene's Test} & \textbf{Brown-Forsythe Test} \\
\hline
RMSE\(_{ATE}\)  & $1.25 \times 10^{-10}$ & \textit{N/A} & \textit{N/A} & $0.0172$ & $0.0567$ \\
MAE\(_{ATE}\)   & $6.80 \times 10^{-13}$ & \textit{N/A} & \textit{N/A} & $3.52 \times 10^{-5}$ & $0.0046$ \\
MAPE\(_{ATE}\)  & $4.63 \times 10^{-10}$ & \textit{N/A} & \textit{N/A} & $5.83 \times 10^{-6}$ & $0.00091$ \\
Len\(_{ATE}\)   & $0$ & \textit{N/A} & \textit{N/A} & $4.73 \times 10^{-5}$ & $9.86 \times 10^{-5}$ \\
RMSE\(_{CATE}\) & \textit{N/A} & $0.0150$ & $0.0149$ & $0.2229$ & $0.2040$ \\
MAE\(_{CATE}\)  & \textit{N/A} & $0.0148$ & $0.0147$ & $0.1309$ & $0.1315$ \\
MAPE\(_{CATE}\) & $0.1191$ & \textit{N/A} & \textit{N/A} & $0.0385$ & $0.0506$ \\
Len\(_{CATE}\)  & $0.9269$ & \textit{N/A} & \textit{N/A} & $0.0186$ & $0.0656$ \\
SEC\(_{ATE}\)   & $1.71 \times 10^{-30}$ & \textit{N/A} & \textit{N/A} & $2.13 \times 10^{-7}$ & $2.61 \times 10^{-7}$ \\
AEC\(_{ATE}\)   & $1.71 \times 10^{-30}$ & \textit{N/A} & \textit{N/A} & $5.51 \times 10^{-13}$ & $1.84 \times 10^{-13}$ \\
SEC\(_{CATE}\)  & $0.0482$ & \textit{N/A} & \textit{N/A} & $0.00124$ & $0.0734$ \\
AEC\(_{CATE}\)  & $0.0482$ & \textit{N/A} & \textit{N/A} & $0.00060$ & $0.0702$ \\
\hline
\end{tabular}
\end{table}
\end{landscape}

\begin{landscape}
\begin{table}[h!]
\centering
\caption{Statistical Test Results: p-values for Different Metrics (n=250)}\label{Table7}
\begin{tabular}{lccccc}
\hline
\textbf{Metric} & \textbf{Fligner-Policello Test} & \textbf{Mann-Whitney U Test} & \textbf{Kruskal-Wallis H Test} & \textbf{Levene's Test} & \textbf{Brown-Forsythe Test} \\
\hline
RMSE\(_{ATE}\)  & $5.88 \times 10^{-31}$ & \textit{N/A} & \textit{N/A} & $5.57 \times 10^{-5}$ & $0.00282$ \\
MAE\(_{ATE}\)   & $4.99 \times 10^{-22}$ & \textit{N/A} & \textit{N/A} & $3.61 \times 10^{-12}$ & $5.03 \times 10^{-7}$ \\
MAPE\(_{ATE}\)  & $1.95 \times 10^{-16}$ & \textit{N/A} & \textit{N/A} & $1.53 \times 10^{-13}$ & $8.98 \times 10^{-8}$ \\
Len\(_{ATE}\)   & $0$ & \textit{N/A} & \textit{N/A} & $2.55 \times 10^{-5}$ & $2.54 \times 10^{-5}$ \\
RMSE\(_{CATE}\) & $7.04 \times 10^{-25}$ & \textit{N/A} & \textit{N/A} & $2.70 \times 10^{-6}$ & $9.54 \times 10^{-6}$ \\
MAE\(_{CATE}\)  & $1.18 \times 10^{-28}$ & \textit{N/A} & \textit{N/A} & $1.31 \times 10^{-7}$ & $8.62 \times 10^{-7}$ \\
MAPE\(_{CATE}\) & $2.81 \times 10^{-56}$ & \textit{N/A} & \textit{N/A} & $3.45 \times 10^{-8}$ & $9.24 \times 10^{-8}$ \\
Len\(_{CATE}\)  & \textit{N/A} & $0.00363$ & $0.00361$ & $0.2804$ & $0.3984$ \\
SEC\(_{ATE}\)   & $0$ & \textit{N/A} & \textit{N/A} & $1.34 \times 10^{-26}$ & $1.26 \times 10^{-22}$ \\
AEC\(_{ATE}\)   & $0$ & \textit{N/A} & \textit{N/A} & $6.41 \times 10^{-22}$ & $2.66 \times 10^{-21}$ \\
SEC\(_{CATE}\)  & $3.52 \times 10^{-61}$ & \textit{N/A} & \textit{N/A} & $3.88 \times 10^{-31}$ & $2.58 \times 10^{-22}$ \\
AEC\(_{CATE}\)  & $3.52 \times 10^{-61}$ & \textit{N/A} & \textit{N/A} & $5.61 \times 10^{-27}$ & $2.89 \times 10^{-27}$ \\
\hline
\end{tabular}
\end{table}
\end{landscape}

\begin{landscape}
\begin{table}[h!]
\centering
\caption{Statistical Test Results: p-values for Different Metrics (n=500)}\label{Table8}
\begin{tabular}{lccccc}
\hline
\textbf{Metric} & \textbf{Fligner-Policello Test} & \textbf{Mann-Whitney U Test} & \textbf{Kruskal-Wallis H Test} & \textbf{Levene's Test} & \textbf{Brown-Forsythe Test} \\
\hline
RMSE\(_{ATE}\)  & $4.81 \times 10^{-112}$ & \textit{N/A} & \textit{N/A} & $4.24 \times 10^{-3}$ & $8.76 \times 10^{-3}$ \\
MAE\(_{ATE}\)   & $2.56 \times 10^{-41}$  & \textit{N/A} & \textit{N/A} & $5.39 \times 10^{-15}$ & $8.01 \times 10^{-12}$ \\
MAPE\(_{ATE}\)  & $3.05 \times 10^{-26}$  & \textit{N/A} & \textit{N/A} & $2.49 \times 10^{-17}$ & $9.22 \times 10^{-12}$ \\
Len\(_{ATE}\)   & $0$                     & \textit{N/A} & \textit{N/A} & $4.53 \times 10^{-7}$  & $5.31 \times 10^{-7}$  \\
RMSE\(_{CATE}\) & $3.88 \times 10^{-98}$  & \textit{N/A} & \textit{N/A} & $8.44 \times 10^{-4}$  & $2.77 \times 10^{-3}$  \\
MAE\(_{CATE}\)  & $4.37 \times 10^{-109}$ & \textit{N/A} & \textit{N/A} & $2.66 \times 10^{-4}$  & $1.02 \times 10^{-3}$  \\
MAPE\(_{CATE}\) & $0$                     & \textit{N/A} & \textit{N/A} & $1.98 \times 10^{-4}$  & $1.91 \times 10^{-4}$  \\
Len\(_{CATE}\)  & \textit{N/A}            & $8.96 \times 10^{-6}$  & $8.91 \times 10^{-6}$ & $0.2330$  & $0.3937$ \\
SEC\(_{ATE}\)   & $0$                     & \textit{N/A} & \textit{N/A} & $1.31 \times 10^{-37}$ & $1.01 \times 10^{-37}$ \\
AEC\(_{ATE}\)   & $0$                     & \textit{N/A} & \textit{N/A} & $1.68 \times 10^{-32}$ & $4.74 \times 10^{-31}$ \\
SEC\(_{CATE}\)  & $0$                     & \textit{N/A} & \textit{N/A} & $2.19 \times 10^{-27}$ & $2.38 \times 10^{-27}$ \\
AEC\(_{CATE}\)  & $0$                     & \textit{N/A} & \textit{N/A} & $1.05 \times 10^{-18}$ & $1.48 \times 10^{-17}$ \\
\hline
\end{tabular}
\end{table}
\end{landscape}

\subsubsection{$\alpha=4$}

The performance measures results for $\alpha=4$ can be seen in \ref{Table9}. Concerning the point-wise estimation performance metrics, the ps-BART model is clearly superior to the BCF model, albeit the latter still remains more robust for ATE function estimation. Moving to the uncertainty estimation performance measures, it can be concluded that the ps-BART model is significantly better in both ATE and CATE function estimation. Anew, such results were already expected for $\alpha=4$.

\begin{figure}[H]
\centering
\begin{subfigure}{0.45\textwidth}
    \centering
     \includegraphics[scale=0.2]{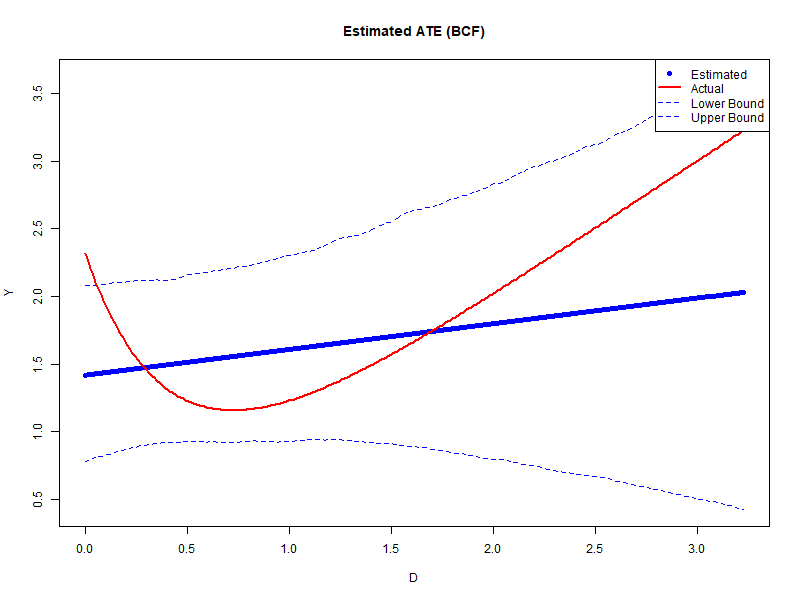}
\end{subfigure}
\hfill
\begin{subfigure}{0.45\textwidth}
    \centering
     \includegraphics[scale=0.2]{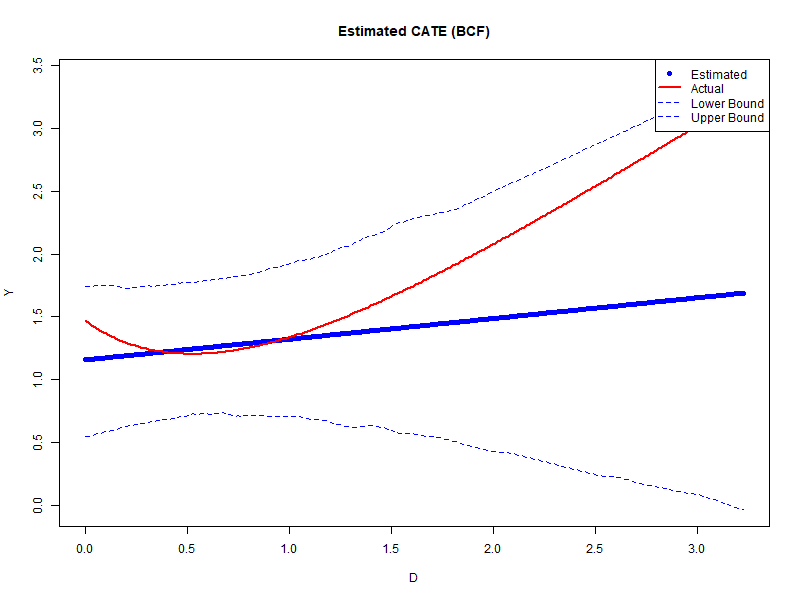}
\end{subfigure}
\caption{BCF ATE and CATE Functions Estimation for N=100 (for CATE, an Example of a Random $\mathbf{x_i}$ of a Random Simulation is used)}\label{Figure13}
\end{figure}

\begin{figure}[H]
\centering
\begin{subfigure}{0.45\textwidth}
    \centering
    \includegraphics[scale=0.2]{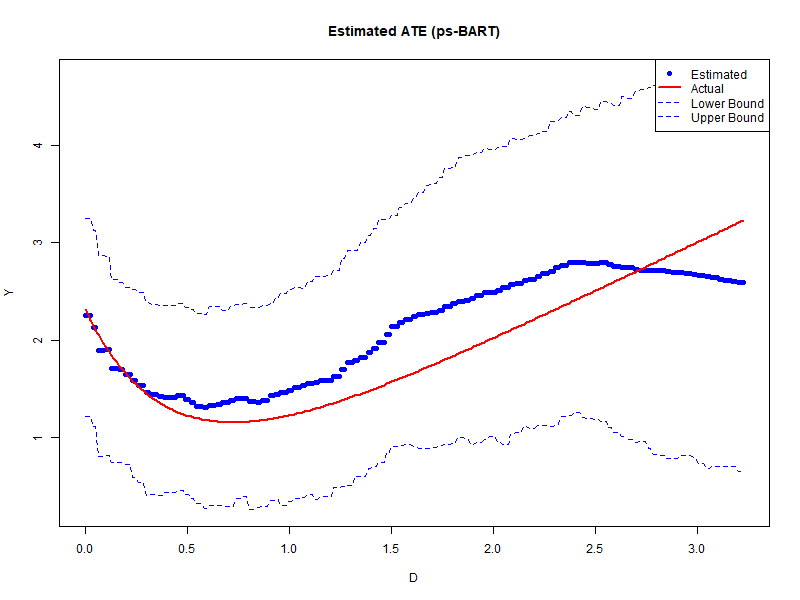}
\end{subfigure}
\hfill
\begin{subfigure}{0.45\textwidth}
    \centering
     \includegraphics[scale=0.2]{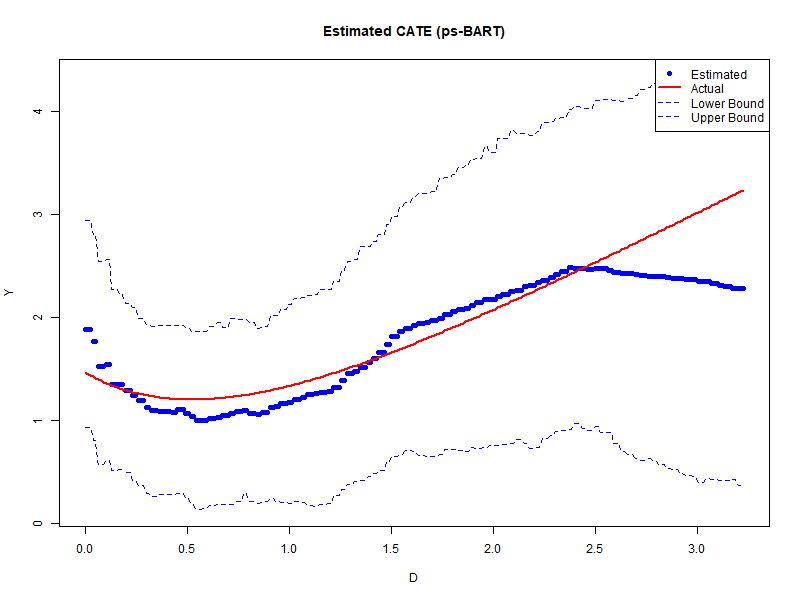}
\end{subfigure}
\caption{ps-BART ATE and CATE Functions Estimation for N=100 (for CATE, an Example of a Random $\mathbf{x_i}$ of a Random Simulation is used)}\label{Figure14}
\end{figure}

\begin{figure}[H]
\centering
\begin{subfigure}{0.45\textwidth}
    \centering
    \includegraphics[scale=0.2]{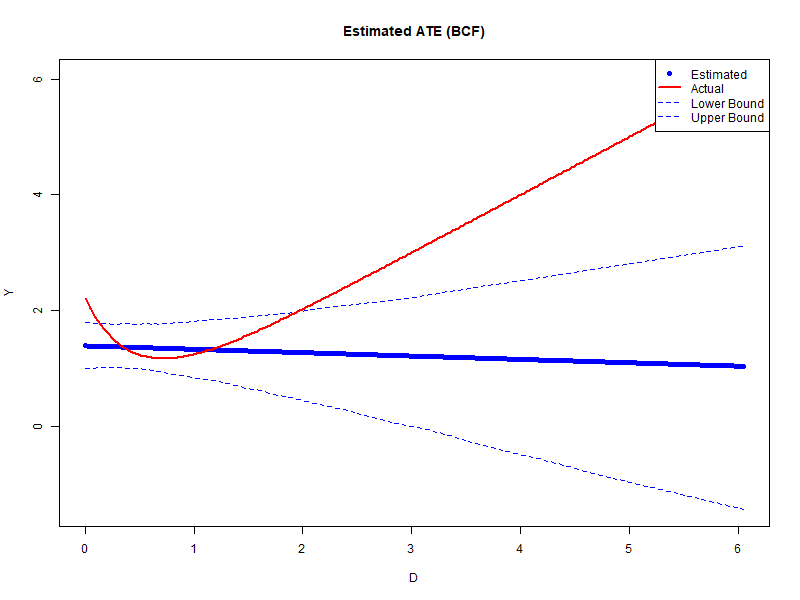}
\end{subfigure}
\hfill
\begin{subfigure}{0.45\textwidth}
    \centering
    \includegraphics[scale=0.2]{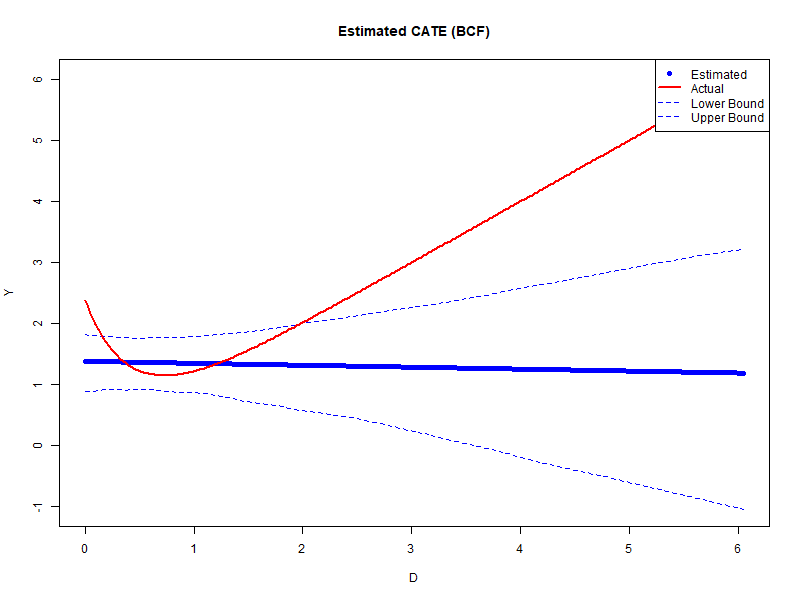}
\end{subfigure}
\caption{BCF ATE and CATE Functions Estimation for N=250 (for CATE, an Example of a Random $\mathbf{x_i}$ of a Random Simulation is used)}\label{Figure15}
\end{figure}

\begin{figure}[H]
\centering
\begin{subfigure}{0.45\textwidth}
    \centering
    \includegraphics[scale=0.2]{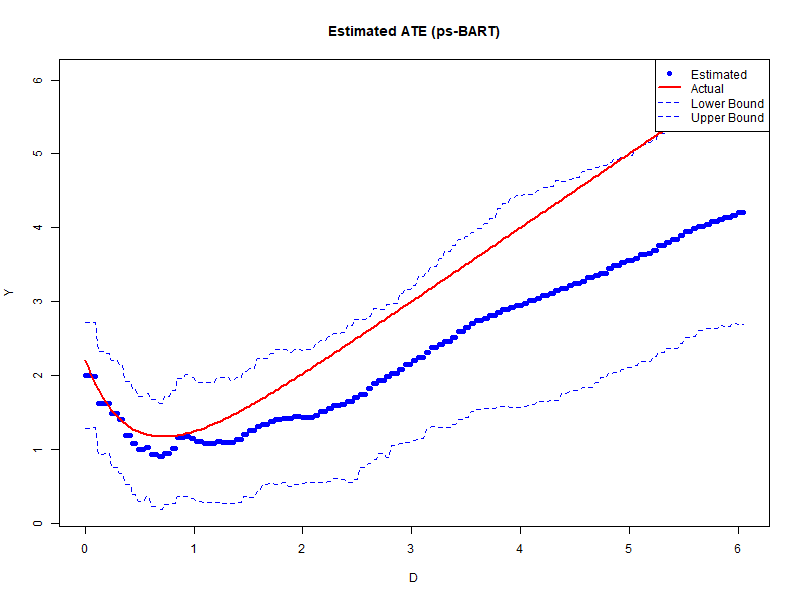}
\end{subfigure}
\hfill
\begin{subfigure}{0.45\textwidth}
    \centering
    \includegraphics[scale=0.2]{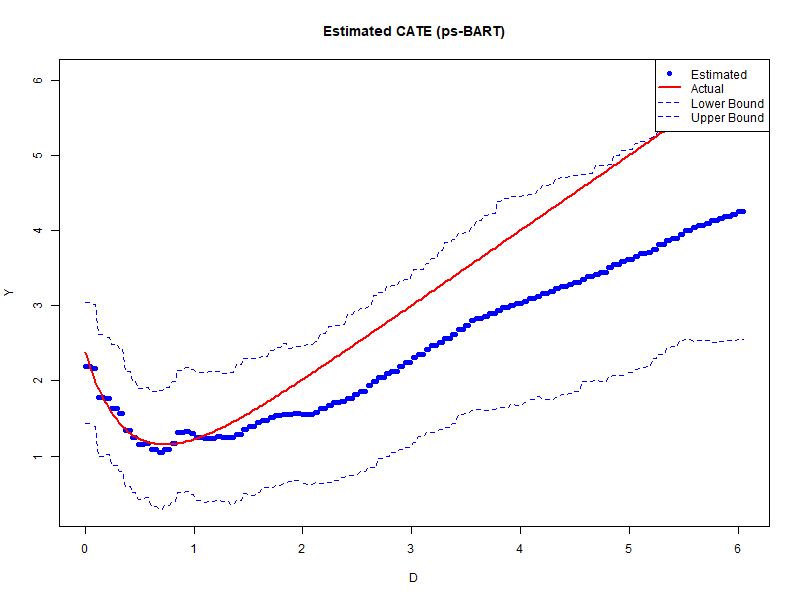}
\end{subfigure}
\caption{ps-BART ATE and CATE Functions Estimation for N=250 (for CATE, an Example of a Random $\mathbf{x_i}$ of a Random Simulation is used)}\label{Figure16}
\end{figure}

\begin{figure}[H]
\centering
\begin{subfigure}{0.45\textwidth}
    \centering
    \includegraphics[scale=0.2]{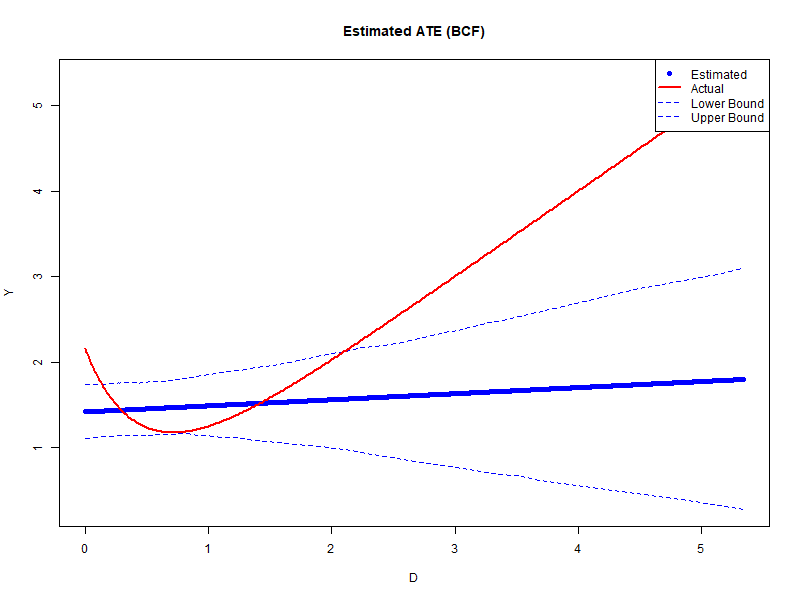}
\end{subfigure}
\hfill
\begin{subfigure}{0.45\textwidth}
    \centering
    \includegraphics[scale=0.2]{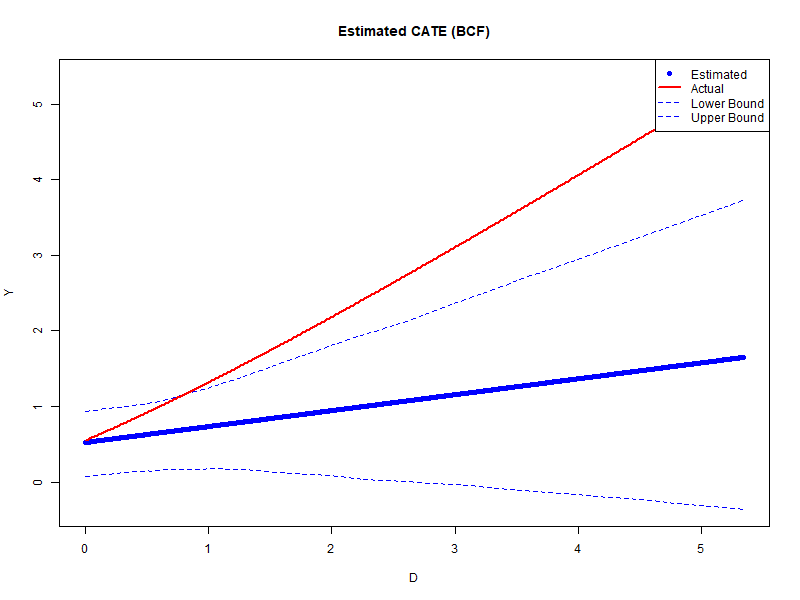}
\end{subfigure}
\caption{BCF ATE and CATE Functions Estimation for N=500 (for CATE, an Example of a Random $\mathbf{x_i}$ of a Random Simulation is used)}\label{Figure17}
\end{figure}

\begin{figure}[H]
\centering
\begin{subfigure}{0.45\textwidth}
    \centering
    \includegraphics[scale=0.2]{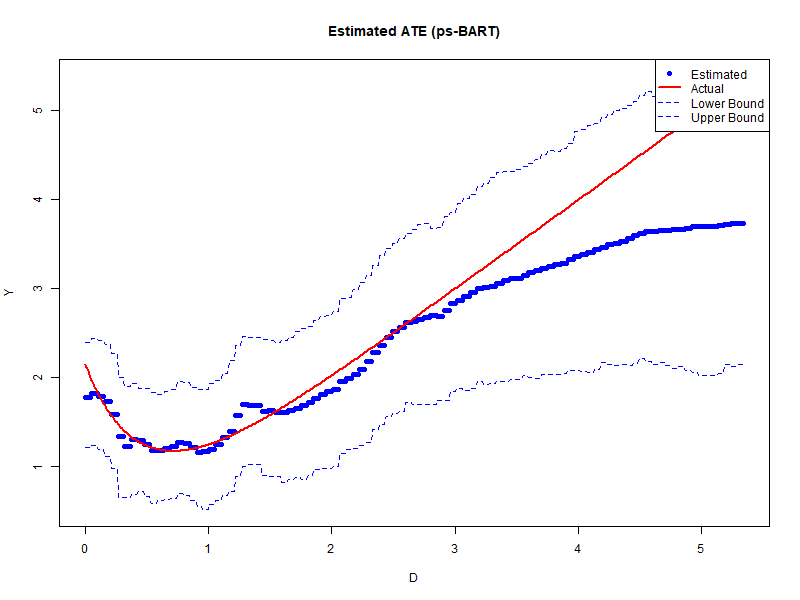}
\end{subfigure}
\hfill
\begin{subfigure}{0.45\textwidth}
    \centering
    \includegraphics[scale=0.2]{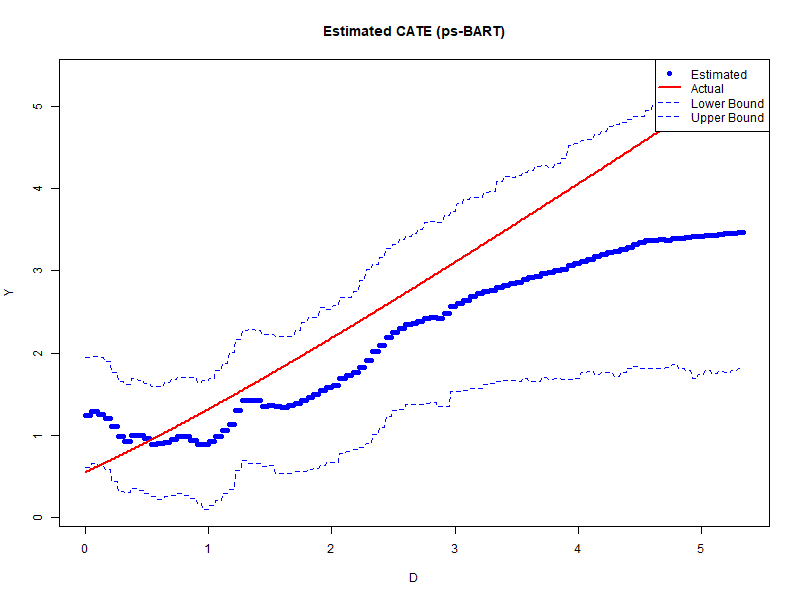}
\end{subfigure}
\caption{ps-BART ATE and CATE Functions Estimation for N=500 (for CATE, an Example of a Random $\mathbf{x_i}$ of a Random Simulation is used)}\label{Figure18}
\end{figure}

Figures \ref{Figure13}, \ref{Figure14}, \ref{Figure15}, \ref{Figure16}, \ref{Figure17}, and \ref{Figure18} graphically illustrate the results of Table \ref{Table9}. The misspecification of the BCF model for the underlying DGPs can be visually seen in these figures.

Tables \ref{Table10}, \ref{Table11}, and \ref{Table12} present the statistical tests results for $n=100$, $n=250$, and $n=500$ respectively. It can be concluded that the ps-BART model is statistically significantly superior to the BCF model, while remaining less robust than the benchmark model for ATE function point-wise estimation (with the expection of RMSE\(_{ATE}\)).

\begin{table}[H]
\centering
\caption{Metric Results}\label{Table9}
\begin{tabular}{llcc}
\hline
$n$ & \textbf{Metric} & \textbf{BCF (Mean $\pm$ SD)} & \textbf{ps-BART (Mean $\pm$ SD)} \\
\hline
100 & RMSE\(_{ATE}\) & $0.459 \pm 0.111$ & $0.355 \pm 0.139$ \\
& MAE\(_{ATE}\)  & $0.351 \pm 0.092$ & $0.276 \pm 0.141$ \\
& MAPE\(_{ATE}\) & $0.207 \pm 0.058$ & $0.169 \pm 0.096$ \\
& Len\(_{ATE}\) & $1.174 \pm 0.218$ & $1.725 \pm 0.230$ \\
& Cover\(_{ATE}\) & $0.804 \pm 0.157$ & $0.977 \pm 0.078$ \\
& RMSE\(_{CATE}\) & $1.054 \pm 0.536$ & $0.831 \pm 0.375$ \\
& MAE\(_{CATE}\)  & $0.841 \pm 0.435$ & $0.685 \pm 0.308$ \\
& MAPE\(_{CATE}\) & $0.328 \pm 0.093$ & $0.296 \pm 0.073$ \\
& Len\(_{CATE}\) & $2.440 \pm 0.872$ & $2.574 \pm 0.560$ \\
& Cover\(_{CATE}\) & $0.805 \pm 0.157$ & $0.885 \pm 0.121$ \\
& SEC\(_{ATE}\)  & $0.046 \pm 0.058$ & $0.007 \pm 0.032$ \\
& AEC\(_{ATE}\)  & $0.169 \pm 0.131$ & $0.057 \pm 0.060$ \\
& SEC\(_{CATE}\)  & $0.046 \pm 0.067$ & $0.019 \pm 0.035$ \\
& AEC\(_{CATE}\)  & $0.151 \pm 0.152$ & $0.092 \pm 0.102$ \\
\hline
250 & RMSE\(_{ATE}\) & $0.441 \pm 0.086$ & $0.282 \pm 0.099$ \\
& MAE\(_{ATE}\)  & $0.324 \pm 0.053$ & $0.221 \pm 0.101$ \\
& MAPE\(_{ATE}\) & $0.190 \pm 0.027$ & $0.138 \pm 0.067$ \\
& Len\(_{ATE}\) & $0.889 \pm 0.108$ & $1.446 \pm 0.145$ \\
& Cover\(_{ATE}\) & $0.735 \pm 0.128$ & $0.993 \pm 0.021$ \\
& RMSE\(_{CATE}\) & $1.543 \pm 0.652$ & $0.844 \pm 0.322$ \\
& MAE\(_{CATE}\)  & $1.233 \pm 0.534$ & $0.685 \pm 0.258$ \\
& MAPE\(_{CATE}\) & $0.380 \pm 0.095$ & $0.249 \pm 0.057$ \\
& Len\(_{CATE}\) & $2.199 \pm 0.602$ & $2.469 \pm 0.440$ \\
& Cover\(_{CATE}\) & $0.528 \pm 0.164$ & $0.880 \pm 0.119$ \\
& SEC\(_{ATE}\)  & $0.062 \pm 0.053$ & $0.002 \pm 0.001$ \\
& AEC\(_{ATE}\)  & $0.222 \pm 0.115$ & $0.047 \pm 0.009$ \\
& SEC\(_{CATE}\)  & $0.205 \pm 0.128$ & $0.019 \pm 0.039$ \\
& AEC\(_{CATE}\)  & $0.422 \pm 0.164$ & $0.096 \pm 0.099$ \\
\hline
500 & RMSE\(_{ATE}\) & $0.456 \pm 0.081$ & $0.255 \pm 0.088$ \\
& MAE\(_{ATE}\)  & $0.322 \pm 0.038$ & $0.207 \pm 0.089$ \\
& MAPE\(_{ATE}\) & $0.186 \pm 0.021$ & $0.129 \pm 0.058$ \\
& Len\(_{ATE}\) & $0.719 \pm 0.072$ & $1.293 \pm 0.132$ \\
& Cover\(_{ATE}\) & $0.625 \pm 0.088$ & $0.991 \pm 0.036$ \\
& RMSE\(_{CATE}\) & $2.096 \pm 1.088$ & $0.808 \pm 0.494$ \\
& MAE\(_{CATE}\)  & $1.689 \pm 0.911$ & $0.652 \pm 0.401$ \\
& MAPE\(_{CATE}\) & $0.409 \pm 0.080$ & $0.198 \pm 0.043$ \\
& Len\(_{CATE}\) & $2.208 \pm 0.819$ & $2.577 \pm 0.771$ \\
& Cover\(_{CATE}\) & $0.356 \pm 0.110$ & $0.917 \pm 0.084$ \\
& SEC\(_{ATE}\)  & $0.113 \pm 0.056$ & $0.003 \pm 0.005$ \\
& AEC\(_{ATE}\)  & $0.325 \pm 0.088$ & $0.050 \pm 0.022$ \\
& SEC\(_{CATE}\)  & $0.364 \pm 0.117$ & $0.008 \pm 0.018$ \\
& AEC\(_{CATE}\)  & $0.594 \pm 0.110$ & $0.063 \pm 0.064$ \\
\hline
\end{tabular}
\end{table}

\begin{landscape}
\begin{table}[h!]
\centering
\caption{Statistical Test Results: p-values for Different Metrics (n=100)}\label{Table10}
\begin{tabular}{lccccc}
\hline
\textbf{Metric} & \textbf{Fligner-Policello Test} & \textbf{Mann-Whitney U Test} & \textbf{Kruskal-Wallis H Test} & \textbf{Levene's Test} & \textbf{Brown-Forsythe Test} \\
\hline
RMSE\(_{ATE}\)  & \textit{N/A}              & $1.91 \times 10^{-12}$ & $1.90 \times 10^{-12}$ & $0.5227$                & $0.7332$ \\
MAE\(_{ATE}\)   & \textit{N/A}              & $2.94 \times 10^{-11}$ & $2.91 \times 10^{-11}$ & $0.1015$                & $0.2298$ \\
MAPE\(_{ATE}\)  & $2.15 \times 10^{-12}$    & \textit{N/A}           & \textit{N/A}           & $0.0104$                & $0.0655$ \\
Len\(_{ATE}\)   & \textit{N/A}              & $2.12 \times 10^{-29}$ & $2.09 \times 10^{-29}$ & $0.3127$                & $0.3397$ \\
RMSE\(_{CATE}\) & \textit{N/A}              & $2.89 \times 10^{-4}$  & $2.88 \times 10^{-4}$  & $0.0650$                & $0.1165$ \\
MAE\(_{CATE}\)  & \textit{N/A}              & $2.56 \times 10^{-3}$  & $2.55 \times 10^{-3}$  & $0.0883$                & $0.1377$ \\
MAPE\(_{CATE}\) & $0.0171$                  & \textit{N/A}           & \textit{N/A}           & $0.0132$                & $0.0199$ \\
Len\(_{CATE}\)  & $0.0024$                  & \textit{N/A}           & \textit{N/A}           & $0.0015$                & $0.0125$ \\
SEC\(_{ATE}\)   & $1.82 \times 10^{-10}$    & \textit{N/A}           & \textit{N/A}           & $1.26 \times 10^{-11}$  & $4.07 \times 10^{-9}$ \\
AEC\(_{ATE}\)   & $1.82 \times 10^{-10}$    & \textit{N/A}           & \textit{N/A}           & $8.89 \times 10^{-19}$  & $5.13 \times 10^{-19}$ \\
SEC\(_{CATE}\)  & $0.3268$                  & \textit{N/A}           & \textit{N/A}           & $1.10 \times 10^{-6}$   & $1.82 \times 10^{-4}$ \\
AEC\(_{CATE}\)  & $0.3268$                  & \textit{N/A}           & \textit{N/A}           & $1.44 \times 10^{-7}$   & $2.53 \times 10^{-6}$ \\
\hline
\end{tabular}
\end{table}
\end{landscape}

\begin{landscape}
\begin{table}[h!]
\centering
\caption{Statistical Test Results: p-values for Different Metrics (n=250)}\label{Table11}
\begin{tabular}{lccccc}
\hline
\textbf{Metric} & \textbf{Fligner-Policello Test} & \textbf{Mann-Whitney U Test} & \textbf{Kruskal-Wallis H Test} & \textbf{Levene's Test} & \textbf{Brown-Forsythe Test} \\
\hline
RMSE\(_{ATE}\)  & \textit{N/A}              & $1.64 \times 10^{-20}$ & $1.62 \times 10^{-20}$ & $0.1460$                & $0.2392$ \\
MAE\(_{ATE}\)   & $2.12 \times 10^{-21}$    & \textit{N/A}           & \textit{N/A}           & $4.38 \times 10^{-8}$   & $1.18 \times 10^{-5}$ \\
MAPE\(_{ATE}\)  & $1.45 \times 10^{-14}$    & \textit{N/A}           & \textit{N/A}           & $1.62 \times 10^{-12}$  & $2.69 \times 10^{-8}$ \\
Len\(_{ATE}\)   & $0$                       & \textit{N/A}           & \textit{N/A}           & $0.0097$                & $0.0114$ \\
RMSE\(_{CATE}\) & $1.21 \times 10^{-42}$    & \textit{N/A}           & \textit{N/A}           & $6.22 \times 10^{-6}$   & $3.74 \times 10^{-5}$ \\
MAE\(_{CATE}\)  & $2.57 \times 10^{-38}$    & \textit{N/A}           & \textit{N/A}           & $4.59 \times 10^{-6}$   & $2.89 \times 10^{-5}$ \\
MAPE\(_{CATE}\) & $1.76 \times 10^{-63}$    & \textit{N/A}           & \textit{N/A}           & $7.88 \times 10^{-5}$   & $1.74 \times 10^{-4}$ \\
Len\(_{CATE}\)  & $1.26 \times 10^{-6}$     & \textit{N/A}           & \textit{N/A}           & $0.0055$                & $0.0138$ \\
SEC\(_{ATE}\)   & $1.40 \times 10^{-42}$    & \textit{N/A}           & \textit{N/A}           & $1.15 \times 10^{-31}$  & $1.81 \times 10^{-24}$ \\
AEC\(_{ATE}\)   & $1.40 \times 10^{-42}$    & \textit{N/A}           & \textit{N/A}           & $1.50 \times 10^{-30}$  & $1.52 \times 10^{-30}$ \\
SEC\(_{CATE}\)  & $9.87 \times 10^{-115}$   & \textit{N/A}           & \textit{N/A}           & $6.10 \times 10^{-21}$  & $1.69 \times 10^{-20}$ \\
AEC\(_{CATE}\)  & $9.87 \times 10^{-115}$   & \textit{N/A}           & \textit{N/A}           & $8.48 \times 10^{-6}$   & $4.74 \times 10^{-6}$ \\
\hline
\end{tabular}
\end{table}
\end{landscape}

\begin{landscape}
\begin{table}[h!]
\centering
\caption{Statistical Test Results: p-values for Different Metrics (n=500)}\label{Table12}
\begin{tabular}{lccccc}
\hline
\textbf{Metric} & \textbf{Fligner-Policello Test} & \textbf{Mann-Whitney U Test} & \textbf{Kruskal-Wallis H Test} & \textbf{Levene's Test} & \textbf{Brown-Forsythe Test} \\
\hline
RMSE\(_{ATE}\)  & \textit{N/A}              & $1.64 \times 10^{-20}$ & $1.62 \times 10^{-20}$ & $0.1460$                & $0.2392$ \\
MAE\(_{ATE}\)   & $2.12 \times 10^{-21}$    & \textit{N/A}           & \textit{N/A}           & $4.38 \times 10^{-8}$   & $1.18 \times 10^{-5}$ \\
MAPE\(_{ATE}\)  & $1.45 \times 10^{-14}$    & \textit{N/A}           & \textit{N/A}           & $1.62 \times 10^{-12}$  & $2.69 \times 10^{-8}$ \\
Len\(_{ATE}\)   & $0$                       & \textit{N/A}           & \textit{N/A}           & $0.0097$                & $0.0114$ \\
RMSE\(_{CATE}\) & $1.21 \times 10^{-42}$    & \textit{N/A}           & \textit{N/A}           & $6.22 \times 10^{-6}$   & $3.74 \times 10^{-5}$ \\
MAE\(_{CATE}\)  & $2.57 \times 10^{-38}$    & \textit{N/A}           & \textit{N/A}           & $4.59 \times 10^{-6}$   & $2.89 \times 10^{-5}$ \\
MAPE\(_{CATE}\) & $1.76 \times 10^{-63}$    & \textit{N/A}           & \textit{N/A}           & $7.88 \times 10^{-5}$   & $1.74 \times 10^{-4}$ \\
Len\(_{CATE}\)  & $1.26 \times 10^{-6}$     & \textit{N/A}           & \textit{N/A}           & $0.0055$                & $0.0138$ \\
SEC\(_{ATE}\)   & $1.40 \times 10^{-42}$    & \textit{N/A}           & \textit{N/A}           & $1.15 \times 10^{-31}$  & $1.81 \times 10^{-24}$ \\
AEC\(_{ATE}\)   & $1.40 \times 10^{-42}$    & \textit{N/A}           & \textit{N/A}           & $1.50 \times 10^{-30}$  & $1.52 \times 10^{-30}$ \\
SEC\(_{CATE}\)  & $9.87 \times 10^{-115}$   & \textit{N/A}           & \textit{N/A}           & $6.10 \times 10^{-21}$  & $1.69 \times 10^{-20}$ \\
AEC\(_{CATE}\)  & $9.87 \times 10^{-115}$   & \textit{N/A}           & \textit{N/A}           & $8.48 \times 10^{-6}$   & $4.74 \times 10^{-6}$ \\
\hline
\end{tabular}
\end{table}
\end{landscape}

\subsubsection{$\alpha=8$}

The results for $\alpha=4$ are presented in \ref{Table13}. Regarding the point-wise estimation performance measures, the ps-BART model is still superior to the BCF model, though to a less extent considering ATE function estimation and BCF remaining more robust for ATE function estimation. Moving to the uncertainty estimation performance metrics, it can be concluded that the ps-BART model is superior in both ATE and CATE function estimation. Interestingly the jump from 4 to 8 for the $\alpha$ value had a more negative impact on the relative performance of the ps-BART model than the BCF model for ATE function estimation, showing a limitation of the proposed model for highly nonlinear relationships between the treatment and the outcome.

\begin{figure}[H]
\centering
\begin{subfigure}{0.45\textwidth}
    \centering
     \includegraphics[scale=0.2]{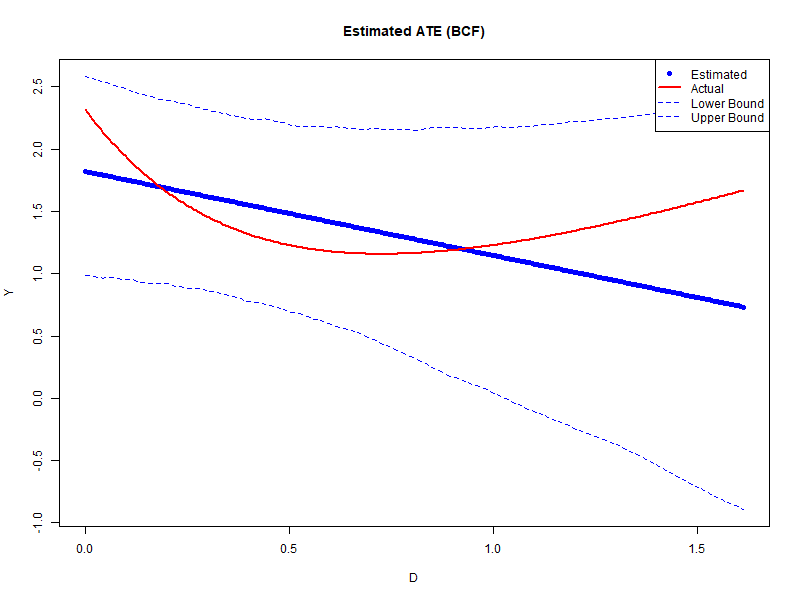}
\end{subfigure}
\hfill
\begin{subfigure}{0.45\textwidth}
    \centering
     \includegraphics[scale=0.2]{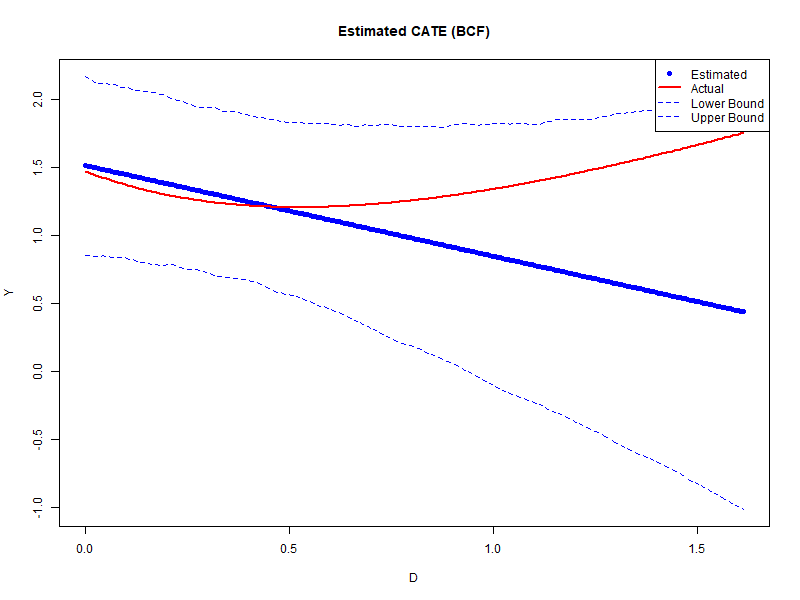}
\end{subfigure}
\caption{BCF ATE and CATE Functions Estimation for N=100 (for CATE, an Example of a Random $\mathbf{x_i}$ of a Random Simulation is used)}\label{Figure19}
\end{figure}

\begin{figure}[H]
\centering
\begin{subfigure}{0.45\textwidth}
    \centering
    \includegraphics[scale=0.2]{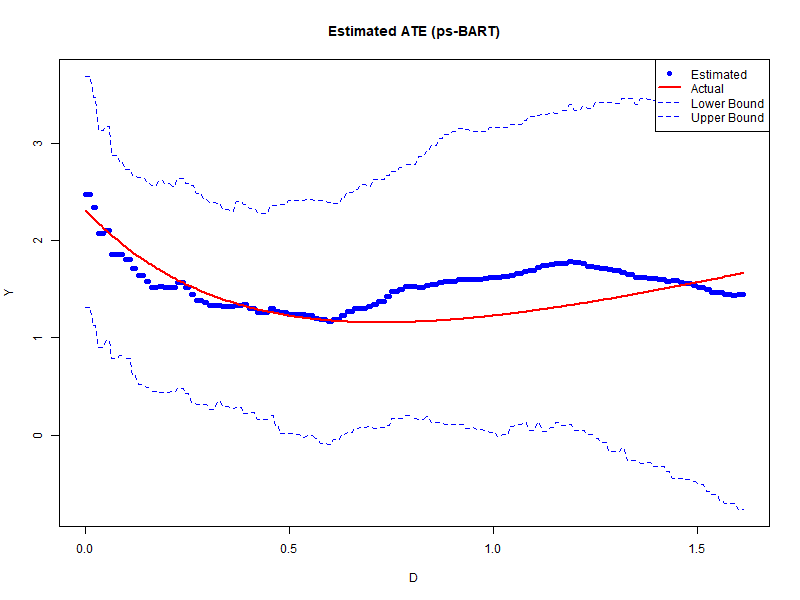}
\end{subfigure}
\hfill
\begin{subfigure}{0.45\textwidth}
    \centering
     \includegraphics[scale=0.2]{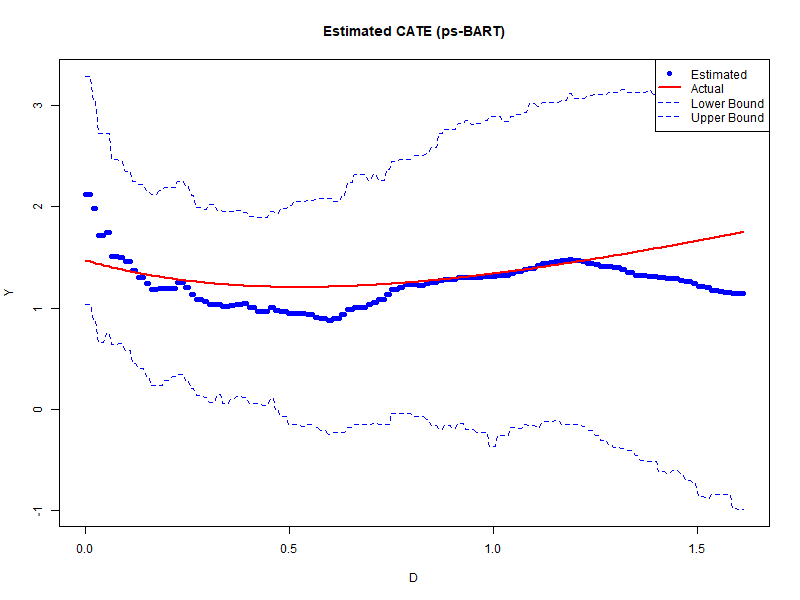}
\end{subfigure}
\caption{ps-BART ATE and CATE Functions Estimation for N=100 (for CATE, an Example of a Random $\mathbf{x_i}$ of a Random Simulation is used)}\label{Figure20}
\end{figure}

\begin{figure}[H]
\centering
\begin{subfigure}{0.45\textwidth}
    \centering
    \includegraphics[scale=0.2]{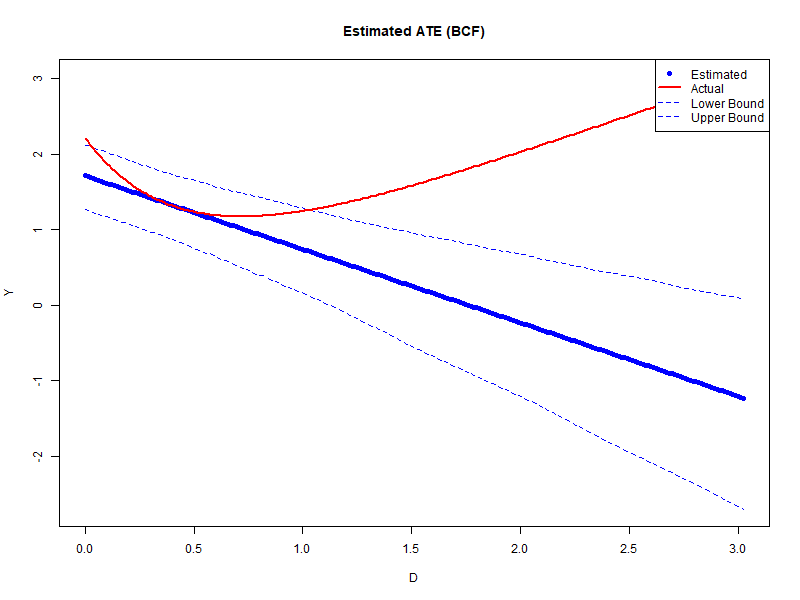}
\end{subfigure}
\hfill
\begin{subfigure}{0.45\textwidth}
    \centering
    \includegraphics[scale=0.2]{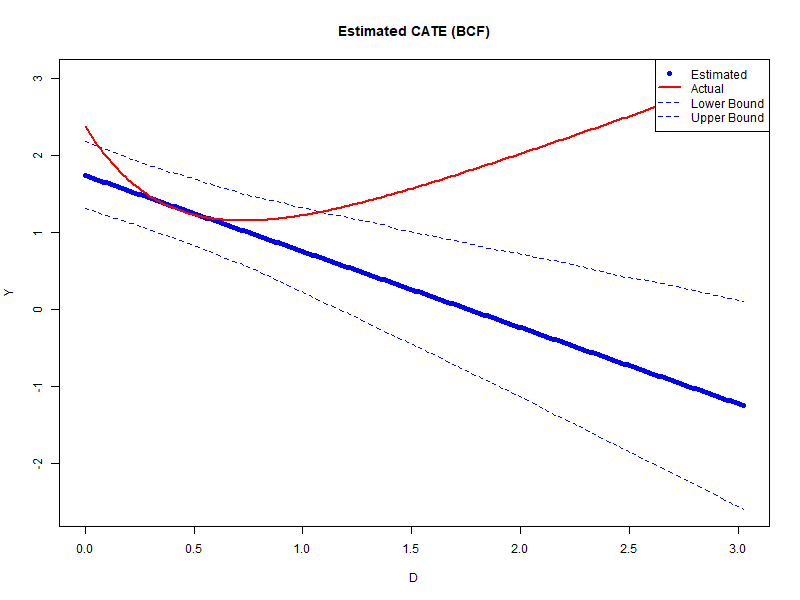}
\end{subfigure}
\caption{BCF ATE and CATE Functions Estimation for N=250 (for CATE, an Example of a Random $\mathbf{x_i}$ of a Random Simulation is used)}\label{Figure21}
\end{figure}

\begin{figure}[H]
\centering
\begin{subfigure}{0.45\textwidth}
    \centering
    \includegraphics[scale=0.2]{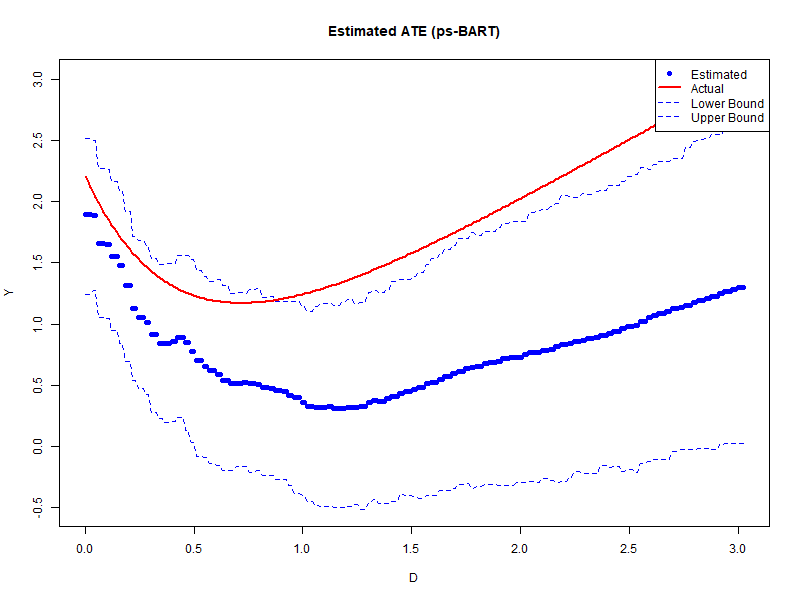}
\end{subfigure}
\hfill
\begin{subfigure}{0.45\textwidth}
    \centering
    \includegraphics[scale=0.2]{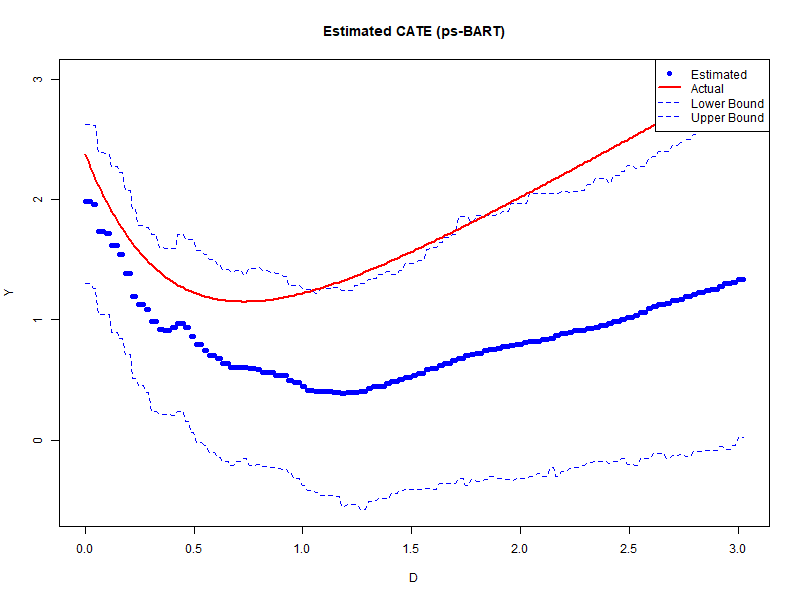}
\end{subfigure}
\caption{ps-BART ATE and CATE Functions Estimation for N=250 (for CATE, an Example of a Random $\mathbf{x_i}$ of a Random Simulation is used)}\label{Figure22}
\end{figure}

\begin{figure}[H]
\centering
\begin{subfigure}{0.45\textwidth}
    \centering
    \includegraphics[scale=0.2]{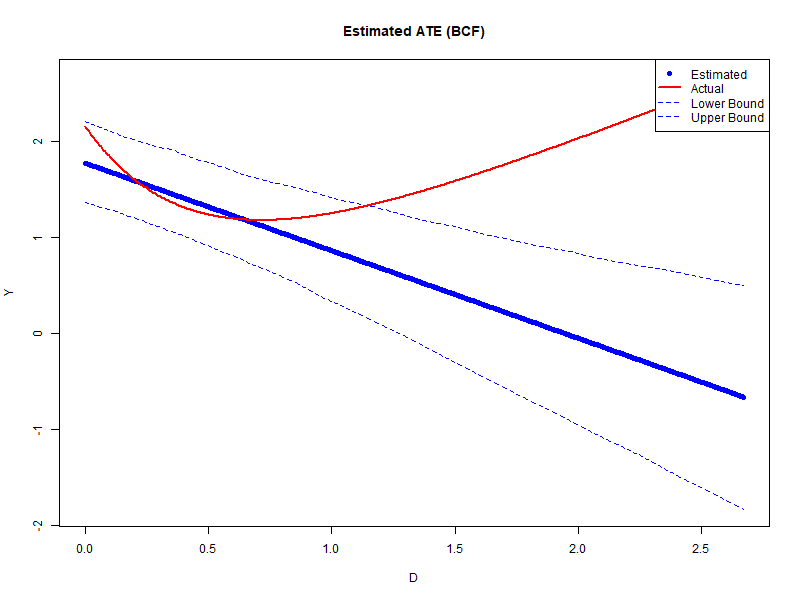}
\end{subfigure}
\hfill
\begin{subfigure}{0.45\textwidth}
    \centering
    \includegraphics[scale=0.2]{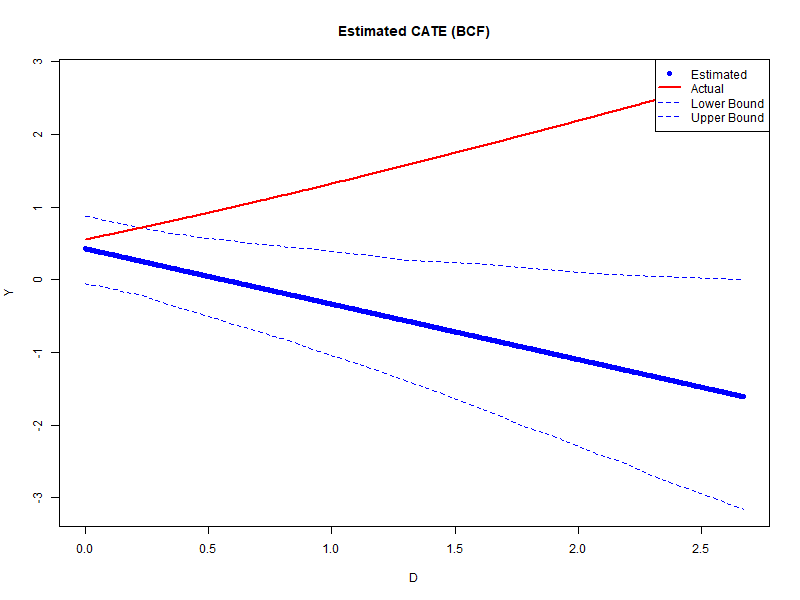}
\end{subfigure}
\caption{BCF ATE and CATE Functions Estimation for N=500 (for CATE, an Example of a Random $\mathbf{x_i}$ of a Random Simulation is used)}\label{Figure23}
\end{figure}

\begin{figure}[H]
\centering
\begin{subfigure}{0.45\textwidth}
    \centering
    \includegraphics[scale=0.2]{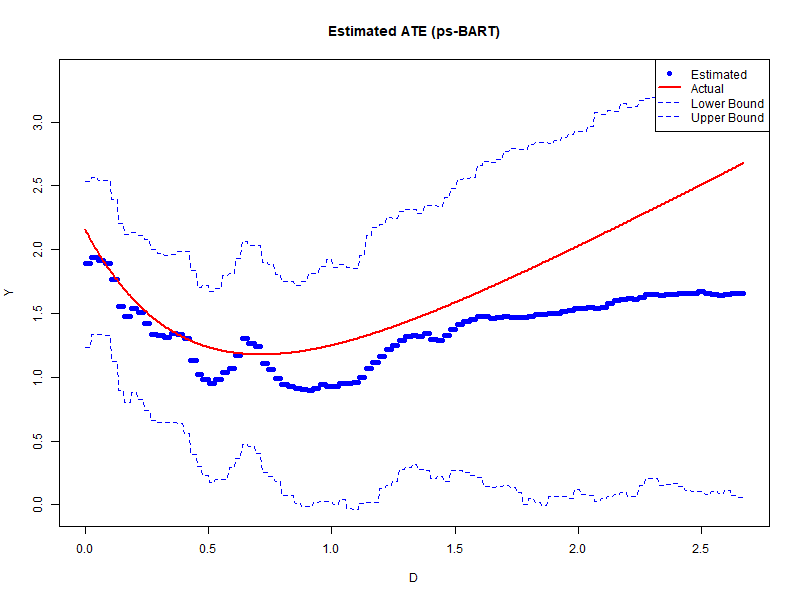}
\end{subfigure}
\hfill
\begin{subfigure}{0.45\textwidth}
    \centering
    \includegraphics[scale=0.2]{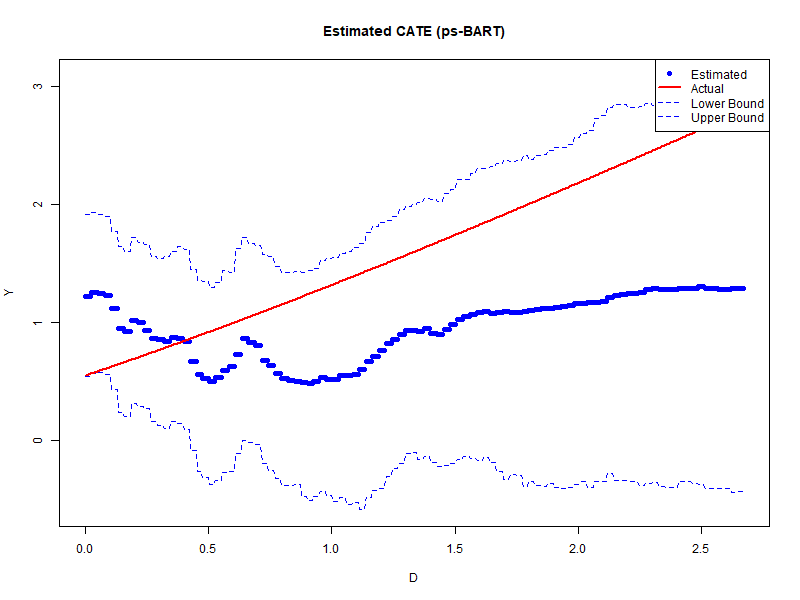}
\end{subfigure}
\caption{ps-BART ATE and CATE Functions Estimation for N=500 (for CATE, an Example of a Random $\mathbf{x_i}$ of a Random Simulation is used)}\label{Figure24}
\end{figure}

Figures \ref{Figure19}, \ref{Figure20}, \ref{Figure21}, \ref{Figure22}, \ref{Figure23}, and \ref{Figure24} graphically illustrate the results of Table \ref{Table13}. The misspecification of the BCF model for the underlying DGPs can be visually seen in these figures.

Tables \ref{Table14}, \ref{Table15}, and \ref{Table16} show the statistical tests results for $n=100$, $n=250$, and $n=500$ respectively. It can be seen that the proposed model is statistically significantly superior to the BCF model, while remaining less robust than the benchmark model for ATE function point-wise estimation for $n=250$ and $n=500$.

\begin{table}[H]
\centering
\caption{Metric Results}\label{Table13}
\begin{tabular}{llcc}
\hline
$n$ & \textbf{Metric} & \textbf{BCF (Mean $\pm$ SD)} & \textbf{ps-BART (Mean $\pm$ SD)} \\
\hline
100 & RMSE\(_{ATE}\) & $0.352 \pm 0.118$ & $0.301 \pm 0.158$ \\
& MAE\(_{ATE}\)  & $0.278 \pm 0.115$ & $0.257 \pm 0.163$ \\
& MAPE\(_{ATE}\) & $0.166 \pm 0.070$ & $0.160 \pm 0.102$ \\
& Len\(_{ATE}\) & $1.248 \pm 0.251$ & $1.670 \pm 0.196$ \\
& Cover\(_{ATE}\) & $0.906 \pm 0.161$ & $0.984 \pm 0.073$ \\
& RMSE\(_{CATE}\) & $0.859 \pm 0.371$ & $0.638 \pm 0.234$ \\
& MAE\(_{CATE}\)  & $0.694 \pm 0.299$ & $0.546 \pm 0.204$ \\
& MAPE\(_{CATE}\) & $0.440 \pm 0.125$ & $0.381 \pm 0.101$ \\
& Len\(_{CATE}\) & $2.021 \pm 0.537$ & $2.451 \pm 0.352$ \\
& Cover\(_{CATE}\) & $0.795 \pm 0.130$ & $0.940 \pm 0.088$ \\
& SEC\(_{ATE}\)  & $0.028 \pm 0.060$ & $0.006 \pm 0.030$ \\
& AEC\(_{ATE}\)  & $0.106 \pm 0.129$ & $0.058 \pm 0.055$ \\
& SEC\(_{CATE}\)  & $0.041 \pm 0.050$ & $0.008 \pm 0.033$ \\
& AEC\(_{CATE}\)  & $0.157 \pm 0.128$ & $0.047 \pm 0.075$ \\
\hline
250 & RMSE\(_{ATE}\) & $0.327 \pm 0.090$ & $0.259 \pm 0.116$ \\
& MAE\(_{ATE}\)  & $0.226 \pm 0.081$ & $0.216 \pm 0.118$ \\
& MAPE\(_{ATE}\) & $0.134 \pm 0.045$ & $0.136 \pm 0.073$ \\
& Len\(_{ATE}\) & $0.935 \pm 0.111$ & $1.411 \pm 0.139$ \\
& Cover\(_{ATE}\) & $0.905 \pm 0.148$ & $0.991 \pm 0.034$ \\
& RMSE\(_{CATE}\) & $1.344 \pm 0.487$ & $0.706 \pm 0.214$ \\
& MAE\(_{CATE}\)  & $1.076 \pm 0.393$ & $0.611 \pm 0.187$ \\
& MAPE\(_{CATE}\) & $0.575 \pm 0.139$ & $0.375 \pm 0.084$ \\
& Len\(_{CATE}\) & $1.764 \pm 0.384$ & $2.334 \pm 0.300$ \\
& Cover\(_{CATE}\) & $0.501 \pm 0.128$ & $0.904 \pm 0.109$ \\
& SEC\(_{ATE}\)  & $0.024 \pm 0.055$ & $0.003 \pm 0.004$ \\
& AEC\(_{ATE}\)  & $0.093 \pm 0.124$ & $0.050 \pm 0.018$ \\
& SEC\(_{CATE}\)  & $0.218 \pm 0.109$ & $0.014 \pm 0.039$ \\
& AEC\(_{CATE}\)  & $0.449 \pm 0.128$ & $0.070 \pm 0.095$ \\
\hline
500 & RMSE\(_{ATE}\) & $0.341 \pm 0.079$ & $0.233 \pm 0.093$ \\
& MAE\(_{ATE}\)  & $0.224 \pm 0.061$ & $0.198 \pm 0.096$ \\
& MAPE\(_{ATE}\) & $0.130 \pm 0.034$ & $0.124 \pm 0.060$ \\
& Len\(_{ATE}\) & $0.775 \pm 0.078$ & $1.218 \pm 0.100$ \\
& Cover\(_{ATE}\) & $0.865 \pm 0.144$ & $0.992 \pm 0.036$ \\
& RMSE\(_{CATE}\) & $1.883 \pm 0.871$ & $0.700 \pm 0.337$ \\
& MAE\(_{CATE}\)  & $1.514 \pm 0.720$ & $0.605 \pm 0.286$ \\
& MAPE\(_{CATE}\) & $0.662 \pm 0.121$ & $0.319 \pm 0.070$ \\
& Len\(_{CATE}\) & $1.736 \pm 0.563$ & $2.293 \pm 0.334$ \\
& Cover\(_{CATE}\) & $0.336 \pm 0.081$ & $0.892 \pm 0.110$ \\
& SEC\(_{ATE}\)  & $0.028 \pm 0.050$ & $0.003 \pm 0.005$ \\
& AEC\(_{ATE}\)  & $0.114 \pm 0.122$ & $0.051 \pm 0.020$ \\
& SEC\(_{CATE}\)  & $0.383 \pm 0.096$ & $0.015 \pm 0.038$ \\
& AEC\(_{CATE}\)  & $0.614 \pm 0.081$ & $0.080 \pm 0.095$ \\
\hline
\end{tabular}
\end{table}

\begin{landscape}
\begin{table}[h!]
\centering
\caption{Statistical Test Results: p-values for Different Metrics (n=100)}\label{Table14}
\begin{tabular}{lccccc}
\hline
\textbf{Metric} & \textbf{Fligner-Policello Test} & \textbf{Mann-Whitney U Test} & \textbf{Kruskal-Wallis H Test} & \textbf{Levene's Test} & \textbf{Brown-Forsythe Test} \\
\hline
RMSE\(_{ATE}\)  & \textit{N/A}                & $2.38 \times 10^{-5}$ & $2.37 \times 10^{-5}$ & $0.2042$               & $0.3304$ \\
MAE\(_{ATE}\)   & \textit{N/A}                & $9.63 \times 10^{-3}$ & $9.60 \times 10^{-3}$ & $0.1057$               & $0.1838$ \\
MAPE\(_{ATE}\)  & $5.02 \times 10^{-2}$       & \textit{N/A}          & \textit{N/A}          & $0.0340$               & $0.0635$ \\
Len\(_{ATE}\)   & $4.92 \times 10^{-85}$      & \textit{N/A}          & \textit{N/A}          & $0.0468$               & $0.0493$ \\
RMSE\(_{CATE}\) & $8.85 \times 10^{-9}$       & \textit{N/A}          & \textit{N/A}          & $7.06 \times 10^{-4}$  & $2.28 \times 10^{-3}$ \\
MAE\(_{CATE}\)  & $1.54 \times 10^{-5}$       & \textit{N/A}          & \textit{N/A}          & $4.50 \times 10^{-3}$  & $6.48 \times 10^{-3}$ \\
MAPE\(_{CATE}\) & $3.57 \times 10^{-4}$       & \textit{N/A}          & \textit{N/A}          & $1.15 \times 10^{-2}$  & $1.18 \times 10^{-2}$ \\
Len\(_{CATE}\)  & $1.34 \times 10^{-18}$      & \textit{N/A}          & \textit{N/A}          & $0.0110$               & $0.0295$ \\
SEC\(_{ATE}\)   & $2.24 \times 10^{-1}$       & \textit{N/A}          & \textit{N/A}          & $8.77 \times 10^{-9}$  & $1.19 \times 10^{-3}$ \\
AEC\(_{ATE}\)   & $2.24 \times 10^{-1}$       & \textit{N/A}          & \textit{N/A}          & $5.48 \times 10^{-14}$ & $2.55 \times 10^{-5}$ \\
SEC\(_{CATE}\)  & $1.53 \times 10^{-10}$      & \textit{N/A}          & \textit{N/A}          & $8.75 \times 10^{-9}$  & $1.78 \times 10^{-8}$ \\
AEC\(_{CATE}\)  & $1.53 \times 10^{-10}$      & \textit{N/A}          & \textit{N/A}          & $6.05 \times 10^{-14}$ & $6.80 \times 10^{-15}$ \\
\hline
\end{tabular}
\end{table}
\end{landscape}

\begin{landscape}
\begin{table}[h!]
\centering
\caption{Statistical Test Results: p-values for Different Metrics (n=250)}\label{Table15}
\begin{tabular}{lccccc}
\hline
\textbf{Metric} & \textbf{Fligner-Policello Test} & \textbf{Mann-Whitney U Test} & \textbf{Kruskal-Wallis H Test} & \textbf{Levene's Test} & \textbf{Brown-Forsythe Test} \\
\hline
RMSE\(_{ATE}\)  & $6.46 \times 10^{-9}$   & \textit{N/A}    & \textit{N/A}   & $0.0375$    & $0.0656$ \\
MAE\(_{ATE}\)   & $0.0485$                & \textit{N/A}    & \textit{N/A}   & $0.0011$    & $0.0066$ \\
MAPE\(_{ATE}\)  & $0.3220$                & \textit{N/A}    & \textit{N/A}   & $0.00015$   & $0.00058$ \\
Len\(_{ATE}\)   & $0$                     & \textit{N/A}    & \textit{N/A}   & $0.0391$    & $0.0472$ \\
RMSE\(_{CATE}\) & $2.88 \times 10^{-99}$  & \textit{N/A}    & \textit{N/A}   & $7.55 \times 10^{-8}$  & $1.25 \times 10^{-7}$ \\
MAE\(_{CATE}\)  & $5.44 \times 10^{-61}$  & \textit{N/A}    & \textit{N/A}   & $8.83 \times 10^{-7}$  & $1.24 \times 10^{-6}$ \\
MAPE\(_{CATE}\) & $7.59 \times 10^{-73}$  & \textit{N/A}    & \textit{N/A}   & $1.83 \times 10^{-5}$  & $2.07 \times 10^{-5}$ \\
Len\(_{CATE}\)  & $6.06 \times 10^{-47}$  & \textit{N/A}    & \textit{N/A}   & $0.0423$    & $0.1053$ \\
SEC\(_{ATE}\)   & $4.75 \times 10^{-6}$   & \textit{N/A}    & \textit{N/A}   & $2.77 \times 10^{-14}$ & $5.75 \times 10^{-5}$ \\
AEC\(_{ATE}\)   & $4.75 \times 10^{-6}$   & \textit{N/A}    & \textit{N/A}   & $9.30 \times 10^{-20}$ & $2.94 \times 10^{-7}$ \\
SEC\(_{CATE}\)  & $0$                     & \textit{N/A}    & \textit{N/A}   & $1.83 \times 10^{-19}$ & $1.91 \times 10^{-20}$ \\
AEC\(_{CATE}\)  & $0$                     & \textit{N/A}    & \textit{N/A}   & $2.70 \times 10^{-4}$  & $1.63 \times 10^{-5}$ \\
\hline
\end{tabular}
\end{table}
\end{landscape}

\begin{landscape}
\begin{table}[h!]
\centering
\caption{Statistical Test Results: p-values for Different Metrics (n=500)}\label{Table16}
\begin{tabular}{lccccc}
\hline
\textbf{Metric} & \textbf{Fligner-Policello Test} & \textbf{Mann-Whitney U Test} & \textbf{Kruskal-Wallis H Test} & \textbf{Levene's Test} & \textbf{Brown-Forsythe Test} \\
\hline
RMSE\(_{ATE}\)  & $7.83 \times 10^{-25}$ & \textit{N/A}    & \textit{N/A}   & $0.0460$    & $0.0711$ \\
MAE\(_{ATE}\)   & $0.00114$              & \textit{N/A}    & \textit{N/A}   & $8.02 \times 10^{-5}$  & $3.79 \times 10^{-4}$ \\
MAPE\(_{ATE}\)  & $0.0354$               & \textit{N/A}    & \textit{N/A}   & $5.38 \times 10^{-7}$  & $7.41 \times 10^{-6}$ \\
Len\(_{ATE}\)   & $0$                    & \textit{N/A}    & \textit{N/A}   & $0.0296$    & $0.0291$ \\
RMSE\(_{CATE}\) & $2.46 \times 10^{-306}$ & \textit{N/A}    & \textit{N/A}   & $2.17 \times 10^{-6}$  & $4.89 \times 10^{-5}$ \\
MAE\(_{CATE}\)  & $1.33 \times 10^{-243}$ & \textit{N/A}    & \textit{N/A}   & $6.25 \times 10^{-6}$  & $1.18 \times 10^{-4}$ \\
MAPE\(_{CATE}\) & $0$                    & \textit{N/A}    & \textit{N/A}   & $4.06 \times 10^{-4}$  & $3.89 \times 10^{-4}$ \\
Len\(_{CATE}\)  & $5.42 \times 10^{-36}$  & \textit{N/A}    & \textit{N/A}   & $1.64 \times 10^{-3}$  & $7.47 \times 10^{-3}$ \\
SEC\(_{ATE}\)   & $0.5357$               & \textit{N/A}    & \textit{N/A}   & $3.68 \times 10^{-18}$ & $3.19 \times 10^{-7}$ \\
AEC\(_{ATE}\)   & $0.5357$               & \textit{N/A}    & \textit{N/A}   & $4.25 \times 10^{-27}$ & $9.43 \times 10^{-12}$ \\
SEC\(_{CATE}\)  & $0$                    & \textit{N/A}    & \textit{N/A}   & $9.01 \times 10^{-13}$ & $5.98 \times 10^{-14}$ \\
AEC\(_{CATE}\)  & \textit{N/A}           & $3.90 \times 10^{-34}$ & $3.84 \times 10^{-34}$ & $0.2891$    & $0.9279$ \\
\hline
\end{tabular}
\end{table}
\end{landscape}

\subsubsection{Conclusion}

It can be concluded that based on the results of the 1st set of DGPs, the proposed model is statistically significantly better at ATE and CATE functions estimation than the benchmark model, considering both dimensions of point-wise and uncertainty estimation. Even when the degree of nonlinearity is not substantial, the ps-BART model is statistically significantly superior for $n \geq 250$. Finally, it can also be concluded that the BCF model is seemingly statistically significantly more robust than the proposed model for ATE function estimation in the domain of point-wise estimation. Besides this last result, the results of this 1st set of DGPs were already expected given the misspecification of the BCF model for the considered DGPs.

\subsection{2nd Set of DGPs}

\subsubsection{Specification 1}

Table \ref{Table17} presents the performance metrics results for the 2nd Set of DGPs with Specification 1. It can be seen that the ps-BART model becomes clearly superior in both point-wise estimation and uncertainty estimation, while also being more robust than the BCF model for ATE function point-wise estimation this time. The superiority of the ps-BART model is confirmed (i.e., shown to be statistically significant) by the results of the statistical tests found in Tables \ref{Table18}, \ref{Table19}, and \ref{Table20}. Interestingly, the uncertainty estimation performance for CATE function estimation of the proposed model considerably decreased when $n=500$ due to the fact that the model was overconfident.

\begin{figure}[H]
\centering
\begin{subfigure}{0.45\textwidth}
    \centering
     \includegraphics[scale=0.2]{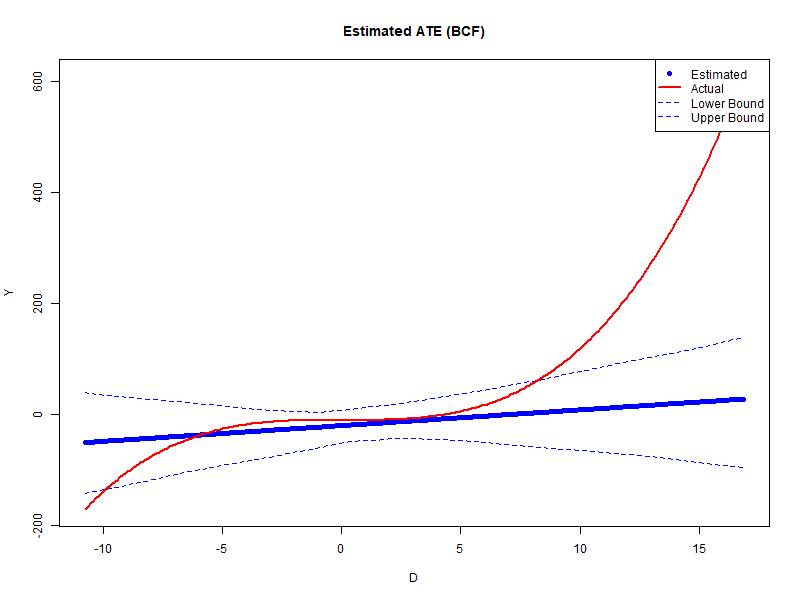}
\end{subfigure}
\hfill
\begin{subfigure}{0.45\textwidth}
    \centering
     \includegraphics[scale=0.2]{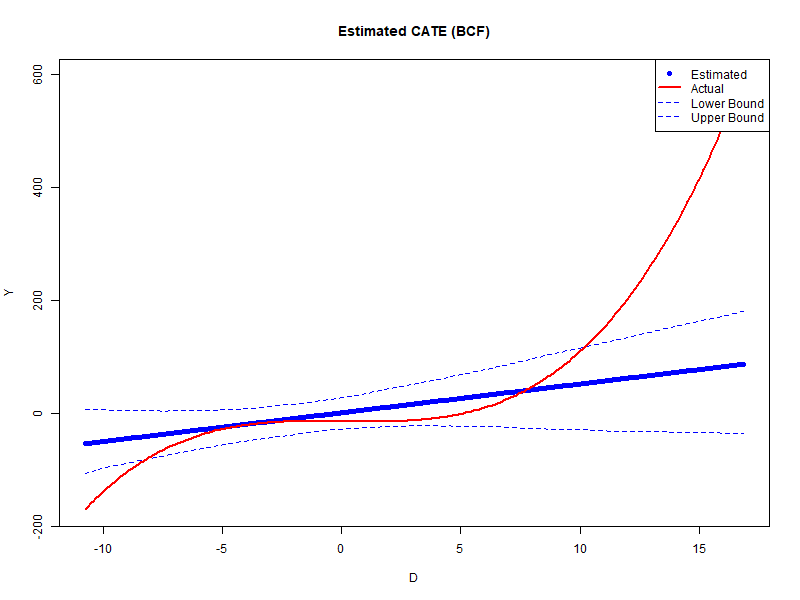}
\end{subfigure}
\caption{BCF ATE and CATE Functions Estimation for N=100 (for CATE, an Example of a Random $\mathbf{x_i}$ of a Random Simulation is used)}\label{Figure25}
\end{figure}

\begin{figure}[H]
\centering
\begin{subfigure}{0.45\textwidth}
    \centering
    \includegraphics[scale=0.2]{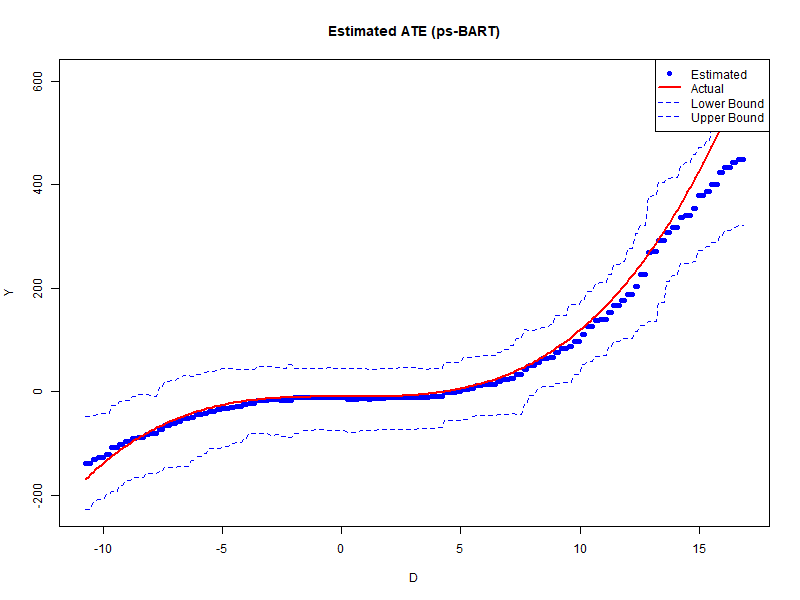}
\end{subfigure}
\hfill
\begin{subfigure}{0.45\textwidth}
    \centering
     \includegraphics[scale=0.2]{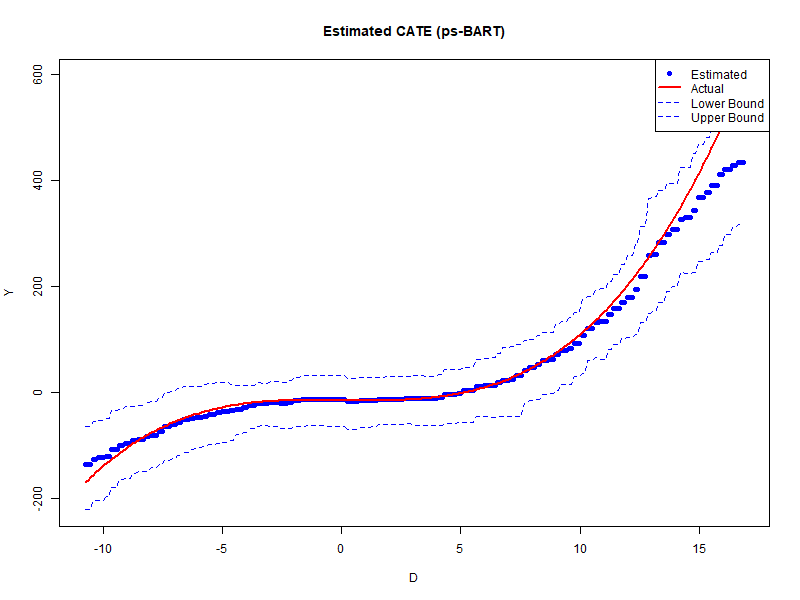}
\end{subfigure}
\caption{ps-BART ATE and CATE Functions Estimation for N=100 (for CATE, an Example of a Random $\mathbf{x_i}$ of a Random Simulation is used)}\label{Figure26}
\end{figure}

\begin{figure}[H]
\centering
\begin{subfigure}{0.45\textwidth}
    \centering
    \includegraphics[scale=0.2]{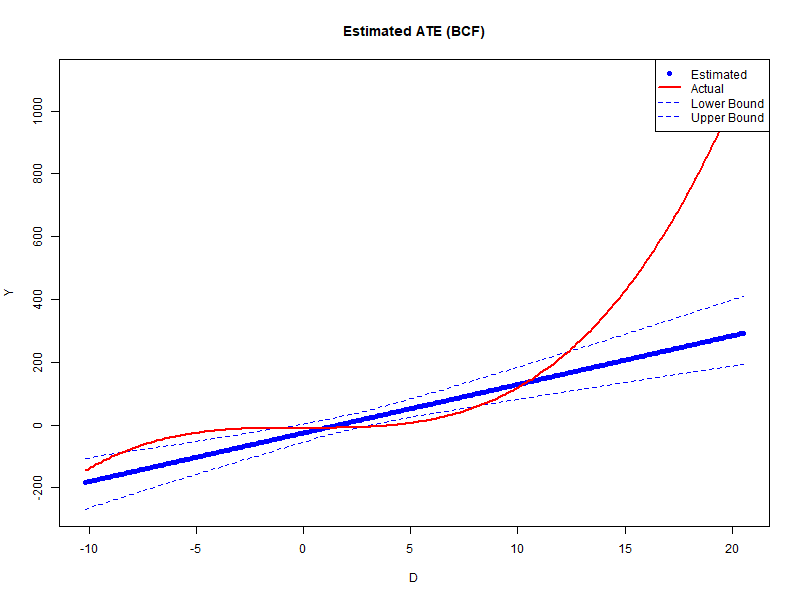}
\end{subfigure}
\hfill
\begin{subfigure}{0.45\textwidth}
    \centering
    \includegraphics[scale=0.2]{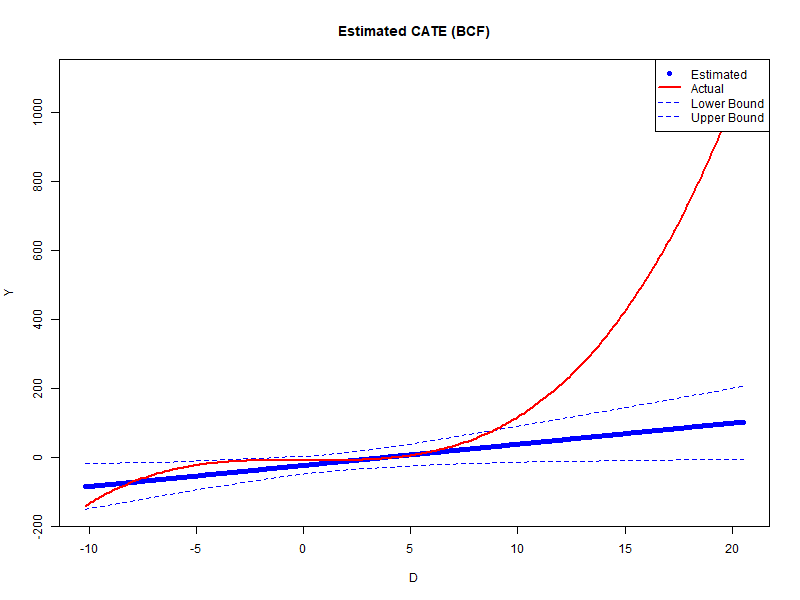}
\end{subfigure}
\caption{BCF ATE and CATE Functions Estimation for N=250 (for CATE, an Example of a Random $\mathbf{x_i}$ of a Random Simulation is used)}\label{Figure27}
\end{figure}

\begin{figure}[H]
\centering
\begin{subfigure}{0.45\textwidth}
    \centering
    \includegraphics[scale=0.2]{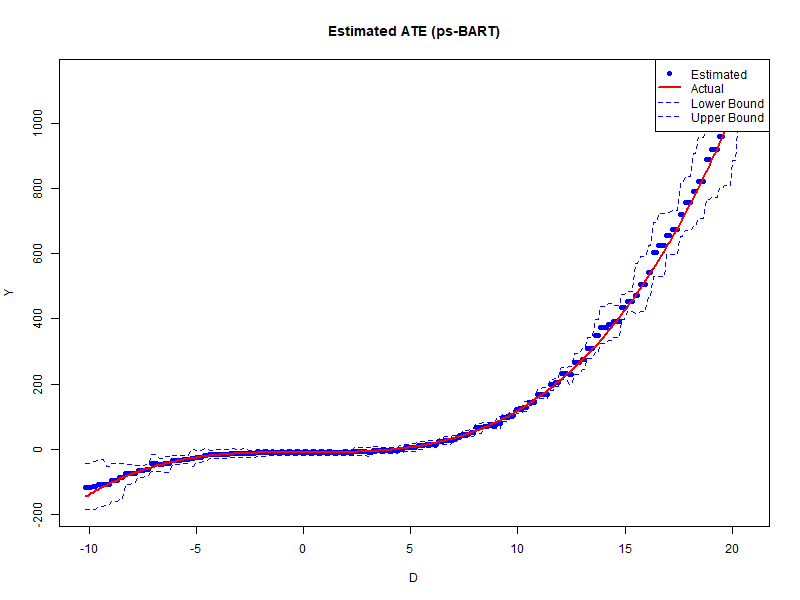}
\end{subfigure}
\hfill
\begin{subfigure}{0.45\textwidth}
    \centering
    \includegraphics[scale=0.2]{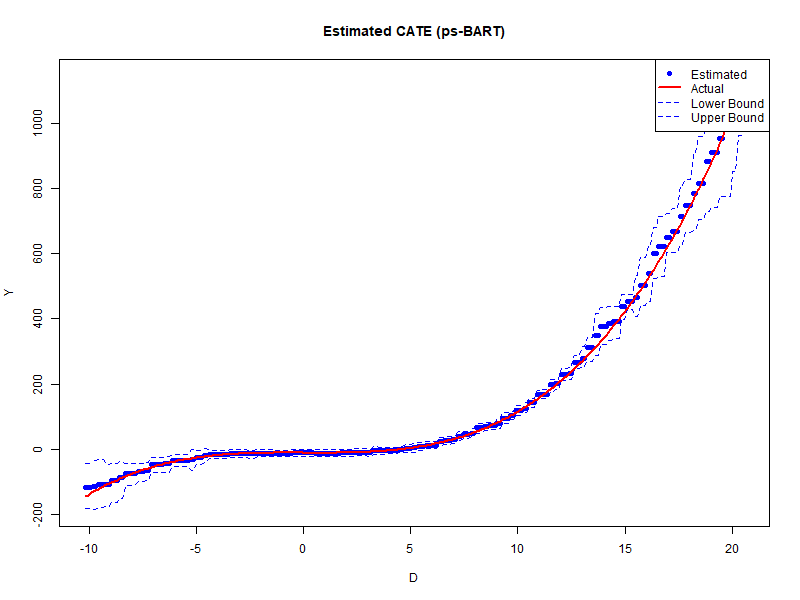}
\end{subfigure}
\caption{ps-BART ATE and CATE Functions Estimation for N=250 (for CATE, an Example of a Random $\mathbf{x_i}$ of a Random Simulation is used)}\label{Figure28}
\end{figure}

\begin{figure}[H]
\centering
\begin{subfigure}{0.45\textwidth}
    \centering
    \includegraphics[scale=0.2]{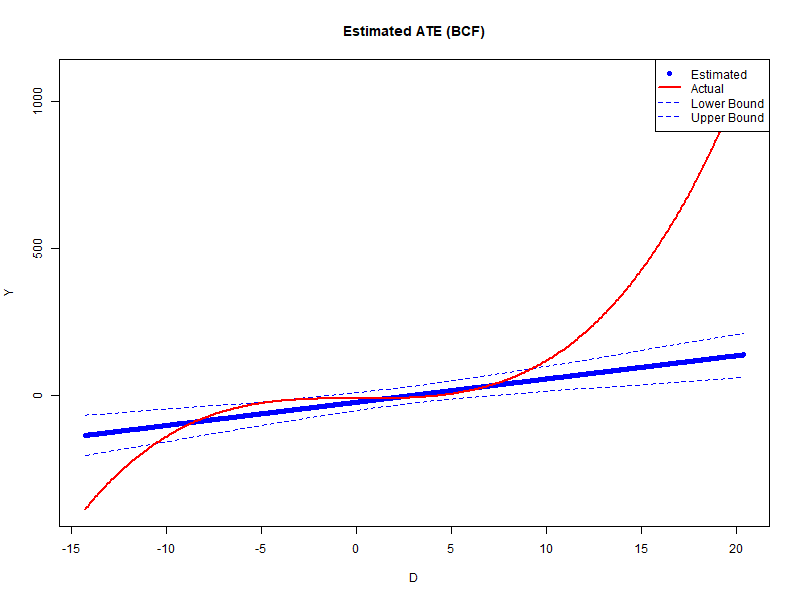}
\end{subfigure}
\hfill
\begin{subfigure}{0.45\textwidth}
    \centering
    \includegraphics[scale=0.2]{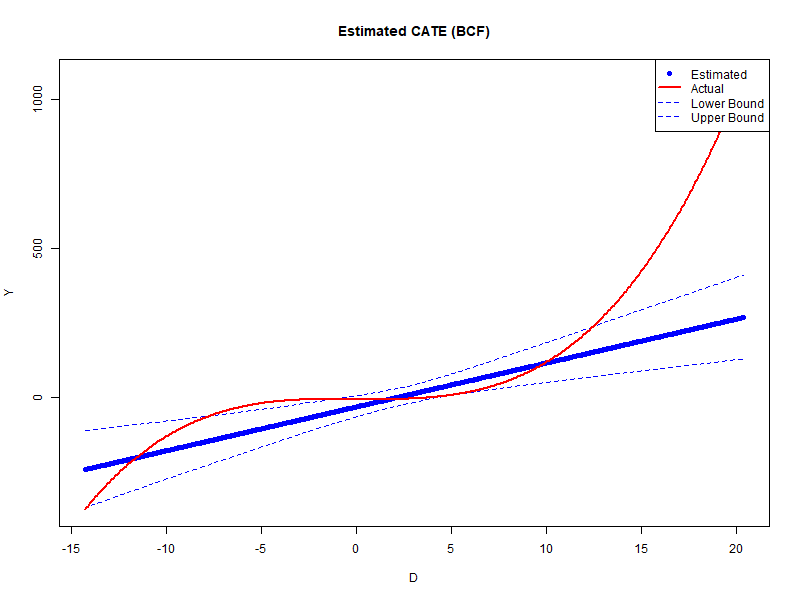}
\end{subfigure}
\caption{BCF ATE and CATE Functions Estimation for N=500 (for CATE, an Example of a Random $\mathbf{x_i}$ of a Random Simulation is used)}\label{Figure29}
\end{figure}

\begin{figure}[H]
\centering
\begin{subfigure}{0.45\textwidth}
    \centering
    \includegraphics[scale=0.2]{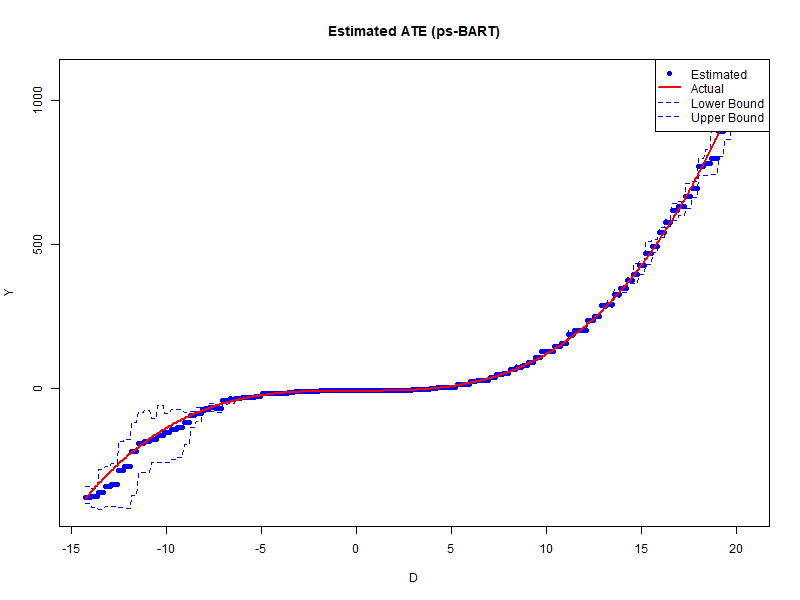}
\end{subfigure}
\hfill
\begin{subfigure}{0.45\textwidth}
    \centering
    \includegraphics[scale=0.2]{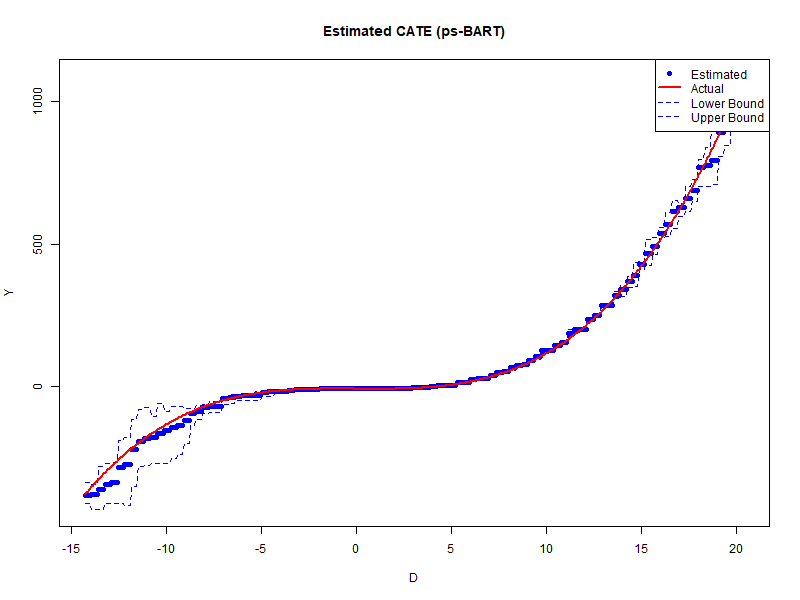}
\end{subfigure}
\caption{ps-BART ATE and CATE Functions Estimation for N=500 (for CATE, an Example of a Random $\mathbf{x_i}$ of a Random Simulation is used)}\label{Figure30}
\end{figure}

Figures \ref{Figure25}, \ref{Figure26}, \ref{Figure27}, \ref{Figure28}, \ref{Figure29}, and \ref{Figure30} graphically illustrate the results of Table \ref{Table17}. The misspecification of the BCF model can be visually seen even more clearly for this DGP.

\begin{table}[H]
\centering
\caption{Metric Results}\label{Table17}
\begin{tabular}{llcc}
\hline
$n$ & \textbf{Metric} & \textbf{BCF (Mean $\pm$ SD)} & \textbf{ps-BART (Mean $\pm$ SD)} \\
\hline
100 & RMSE\(_{ATE}\) & $99 \pm 27$ & $20 \pm 9$ \\
& MAE\(_{ATE}\)  & $51 \pm 12$ & $10 \pm 5$ \\
& MAPE\(_{ATE}\) & $7 \pm 8$ & $2 \pm 3$ \\
& Len\(_{ATE}\) & $101 \pm 29$ & $133 \pm 37$ \\
& Cover\(_{ATE}\) & $0.722 \pm 0.179$ & $0.991 \pm 0.006$ \\
& RMSE\(_{CATE}\) & $205 \pm 89$ & $36 \pm 22$ \\
& MAE\(_{CATE}\)  & $131 \pm 55$ & $21 \pm 12$ \\
& MAPE\(_{CATE}\) & $8 \pm 11$ & $2 \pm 1$ \\
& Len\(_{CATE}\) & $142 \pm 35$ & $159 \pm 61$ \\
& Cover\(_{CATE}\) & $0.536 \pm 0.060$ & $0.970 \pm 0.032$ \\
& SEC\(_{ATE}\)  & $0.084 \pm 0.129$ & $0.002 \pm 0.000$ \\
& AEC\(_{ATE}\)  & $0.229 \pm 0.179$ & $0.041 \pm 0.006$ \\
& SEC\(_{CATE}\)  & $0.175 \pm 0.049$ & $0.001 \pm 0.002$ \\
& AEC\(_{CATE}\)  & $0.414 \pm 0.061$ & $0.033 \pm 0.018$ \\
\hline
250 & RMSE\(_{ATE}\) & $105 \pm 26$ & $8 \pm 7$ \\
& MAE\(_{ATE}\)  & $52 \pm 11$ & $4 \pm 1$ \\
& MAPE\(_{ATE}\) & $6 \pm 8$ & $1 \pm 3$ \\
& Len\(_{ATE}\) & $79 \pm 17$ & $29 \pm 7$ \\
& Cover\(_{ATE}\) & $0.601 \pm 0.187$ & $0.991 \pm 0.008$ \\
& RMSE\(_{CATE}\) & $304 \pm 134$ & $26 \pm 32$ \\
& MAE\(_{CATE}\)  & $192 \pm 84$ & $15 \pm 15$ \\
& MAPE\(_{CATE}\) & $7 \pm 3$ & $1 \pm 4$ \\
& Len\(_{CATE}\) & $133 \pm 26$ & $85 \pm 56$ \\
& Cover\(_{CATE}\) & $0.446 \pm 0.045$ & $0.974 \pm 0.022$ \\
& SEC\(_{ATE}\)  & $0.157 \pm 0.153$ & $0.002 \pm 0.001$ \\
& AEC\(_{ATE}\)  & $0.349 \pm 0.187$ & $0.041 \pm 0.008$ \\
& SEC\(_{CATE}\)  & $0.256 \pm 0.046$ & $0.001 \pm 0.001$ \\
& AEC\(_{CATE}\)  & $0.504 \pm 0.045$ & $0.030 \pm 0.013$ \\
\hline
500 & RMSE\(_{ATE}\) & $103 \pm 17$ & $5 \pm 3$ \\
& MAE\(_{ATE}\)  & $51 \pm 7$ & $3 \pm 1$ \\
& MAPE\(_{ATE}\) & $6 \pm 4$ & $0.3 \pm 0.2$ \\
& Len\(_{ATE}\) & $69 \pm 17$ & $14 \pm 2$ \\
& Cover\(_{ATE}\) & $0.510 \pm 0.181$ & $0.938 \pm 0.029$ \\
& RMSE\(_{CATE}\) & $346 \pm 105$ & $21 \pm 15$ \\
& MAE\(_{CATE}\)  & $216 \pm 65$ & $12 \pm 7$ \\
& MAPE\(_{CATE}\) & $7 \pm 4$ & $1 \pm 0.3$ \\
& Len\(_{CATE}\) & $128 \pm 19$ & $57 \pm 30$ \\
& Cover\(_{CATE}\) & $0.402 \pm 0.038$ & $0.887 \pm 0.029$ \\
& SEC\(_{ATE}\)  & $0.226 \pm 0.152$ & $0.001 \pm 0.002$ \\
& AEC\(_{ATE}\)  & $0.440 \pm 0.181$ & $0.023 \pm 0.022$ \\
& SEC\(_{CATE}\)  & $0.302 \pm 0.041$ & $0.005 \pm 0.006$ \\
& AEC\(_{CATE}\)  & $0.548 \pm 0.038$ & $0.063 \pm 0.029$ \\
\hline
\end{tabular}
\end{table}

\begin{landscape}
\begin{table}[h!]
\centering
\caption{Statistical Test Results: p-values for Different Metrics (n=100)}\label{Table18}
\begin{tabular}{lccccc}
\hline
\textbf{Metric} & \textbf{Fligner-Policello Test} & \textbf{Mann-Whitney U Test} & \textbf{Kruskal-Wallis H Test} & \textbf{Levene's Test} & \textbf{Brown-Forsythe Test} \\
\hline
RMSE\(_{ATE}\)  & $0$                    & \textit{N/A}    & \textit{N/A}   & $1.62 \times 10^{-13}$ & $5.11 \times 10^{-13}$ \\
MAE\(_{ATE}\)   & $0$                    & \textit{N/A}    & \textit{N/A}   & $3.28 \times 10^{-10}$ & $8.74 \times 10^{-9}$  \\
MAPE\(_{ATE}\)  & $6.60 \times 10^{-37}$  & \textit{N/A}    & \textit{N/A}   & $1.99 \times 10^{-6}$  & $9.98 \times 10^{-4}$  \\
Len\(_{ATE}\)   & $2.51 \times 10^{-14}$  & \textit{N/A}    & \textit{N/A}   & $0.0476$               & $0.0567$               \\
RMSE\(_{CATE}\) & $0$                    & \textit{N/A}    & \textit{N/A}   & $1.85 \times 10^{-15}$ & $7.56 \times 10^{-15}$ \\
MAE\(_{CATE}\)  & $0$                    & \textit{N/A}    & \textit{N/A}   & $7.64 \times 10^{-17}$ & $6.08 \times 10^{-16}$ \\
MAPE\(_{CATE}\) & $0$                    & \textit{N/A}    & \textit{N/A}   & $5.72 \times 10^{-3}$  & $0.0414$               \\
Len\(_{CATE}\)  & $0.0824$               & \textit{N/A}    & \textit{N/A}   & $4.09 \times 10^{-4}$  & $1.32 \times 10^{-3}$  \\
SEC\(_{ATE}\)   & $7.56 \times 10^{-54}$  & \textit{N/A}    & \textit{N/A}   & $7.15 \times 10^{-17}$ & $1.28 \times 10^{-8}$  \\
AEC\(_{ATE}\)   & $7.56 \times 10^{-54}$  & \textit{N/A}    & \textit{N/A}   & $5.26 \times 10^{-23}$ & $2.67 \times 10^{-17}$ \\
SEC\(_{CATE}\)  & $0$                    & \textit{N/A}    & \textit{N/A}   & $7.16 \times 10^{-34}$ & $7.15 \times 10^{-34}$ \\
AEC\(_{CATE}\)  & $0$                    & \textit{N/A}    & \textit{N/A}   & $1.82 \times 10^{-19}$ & $9.39 \times 10^{-19}$ \\
\hline
\end{tabular}
\end{table}
\end{landscape}

\begin{landscape}
\begin{table}[h!]
\centering
\caption{Statistical Test Results: p-values for Different Metrics (n=250)}\label{Table19}
\begin{tabular}{lccccc}
\hline
\textbf{Metric} & \textbf{Fligner-Policello Test} & \textbf{Mann-Whitney U Test} & \textbf{Kruskal-Wallis H Test} & \textbf{Levene's Test} & \textbf{Brown-Forsythe Test} \\
\hline
RMSE\(_{ATE}\)  & $0$                    & \textit{N/A}   & \textit{N/A}   & $1.18 \times 10^{-19}$ & $1.01 \times 10^{-18}$ \\
MAE\(_{ATE}\)   & $0$                    & \textit{N/A}   & \textit{N/A}   & $8.31 \times 10^{-18}$ & $1.23 \times 10^{-13}$ \\
MAPE\(_{ATE}\)  & $1.07 \times 10^{-126}$ & \textit{N/A}   & \textit{N/A}   & $2.70 \times 10^{-6}$  & $4.79 \times 10^{-5}$  \\
Len\(_{ATE}\)   & $0$                    & \textit{N/A}   & \textit{N/A}   & $3.77 \times 10^{-11}$ & $8.76 \times 10^{-10}$ \\
RMSE\(_{CATE}\) & $0$                    & \textit{N/A}   & \textit{N/A}   & $2.03 \times 10^{-15}$ & $1.51 \times 10^{-12}$ \\
MAE\(_{CATE}\)  & $0$                    & \textit{N/A}   & \textit{N/A}   & $6.12 \times 10^{-17}$ & $1.24 \times 10^{-13}$ \\
MAPE\(_{CATE}\) & \textit{N/A}           & $2.46 \times 10^{-30}$ & $2.43 \times 10^{-30}$ & $0.1452$               & $0.0545$               \\
Len\(_{CATE}\)  & $1.90 \times 10^{-40}$  & \textit{N/A}   & \textit{N/A}   & $0.0043$               & $0.0391$               \\
SEC\(_{ATE}\)   & $0$                    & \textit{N/A}   & \textit{N/A}   & $1.68 \times 10^{-35}$ & $1.11 \times 10^{-15}$ \\
AEC\(_{ATE}\)   & $0$                    & \textit{N/A}   & \textit{N/A}   & $1.52 \times 10^{-37}$ & $1.86 \times 10^{-22}$ \\
SEC\(_{CATE}\)  & $0$                    & \textit{N/A}   & \textit{N/A}   & $6.75 \times 10^{-30}$ & $4.25 \times 10^{-27}$ \\
AEC\(_{CATE}\)  & $0$                    & \textit{N/A}   & \textit{N/A}   & $1.83 \times 10^{-18}$ & $1.96 \times 10^{-17}$ \\
\hline
\end{tabular}
\end{table}
\end{landscape}

\begin{landscape}
\begin{table}[h!]
\centering
\caption{Statistical Test Results: p-values for Different Metrics (n=500)}\label{Table20}
\begin{tabular}{lccccc}
\hline
\textbf{Metric} & \textbf{Fligner-Policello Test} & \textbf{Mann-Whitney U Test} & \textbf{Kruskal-Wallis H Test} & \textbf{Levene's Test} & \textbf{Brown-Forsythe Test} \\
\hline
RMSE\(_{ATE}\)  & $0$                    & \textit{N/A}    & \textit{N/A}   & $3.52 \times 10^{-26}$ & $3.20 \times 10^{-26}$ \\
MAE\(_{ATE}\)   & $0$                    & \textit{N/A}    & \textit{N/A}   & $3.35 \times 10^{-22}$ & $5.99 \times 10^{-21}$ \\
MAPE\(_{ATE}\)  & $0$                    & \textit{N/A}    & \textit{N/A}   & $4.42 \times 10^{-21}$ & $1.40 \times 10^{-18}$ \\
Len\(_{ATE}\)   & $0$                    & \textit{N/A}    & \textit{N/A}   & $4.33 \times 10^{-12}$ & $9.71 \times 10^{-11}$ \\
RMSE\(_{CATE}\) & $0$                    & \textit{N/A}    & \textit{N/A}   & $3.98 \times 10^{-19}$ & $1.02 \times 10^{-17}$ \\
MAE\(_{CATE}\)  & $0$                    & \textit{N/A}    & \textit{N/A}   & $6.96 \times 10^{-20}$ & $2.97 \times 10^{-18}$ \\
MAPE\(_{CATE}\) & $0$                    & \textit{N/A}    & \textit{N/A}   & $1.47 \times 10^{-10}$ & $2.23 \times 10^{-8}$  \\
Len\(_{CATE}\)  & \textit{N/A}           & $9.45 \times 10^{-30}$ & $9.32 \times 10^{-30}$ & $0.0734$               & $0.197$               \\
SEC\(_{ATE}\)   & $0$                    & \textit{N/A}    & \textit{N/A}   & $8.74 \times 10^{-45}$ & $8.80 \times 10^{-45}$ \\
AEC\(_{ATE}\)   & $0$                    & \textit{N/A}    & \textit{N/A}   & $1.85 \times 10^{-38}$ & $8.05 \times 10^{-33}$ \\
SEC\(_{CATE}\)  & $0$                    & \textit{N/A}    & \textit{N/A}   & $1.98 \times 10^{-21}$ & $1.66 \times 10^{-21}$ \\
AEC\(_{CATE}\)  & $0$                    & \textit{N/A}    & \textit{N/A}   & $0.0170$               & $0.0132$               \\
\hline
\end{tabular}
\end{table}
\end{landscape}

\subsubsection{Specification 2}

The performance measures results for Specification 2 can be found in Table \ref{Table21}. Based on the results, it can be inferred that the ps-BART model is superior in both point-wise estimation and uncertainty estimation, while also being more robust than the benchmark model. The superiority of the ps-BART model is also statistically significant based on the results of the statistical tests found in Tables \ref{Table22}, \ref{Table23}, and \ref{Table24}. Interestingly, the uncertainty estimation performance for both the ATE and CATE functions estimation of the proposed model considerably decreased when $n=500$, which could mean that the model decreases the confidence interval length to a faster rate than it should when $n$ tends to infinity.

\begin{figure}[H]
\centering
\begin{subfigure}{0.45\textwidth}
    \centering
     \includegraphics[scale=0.2]{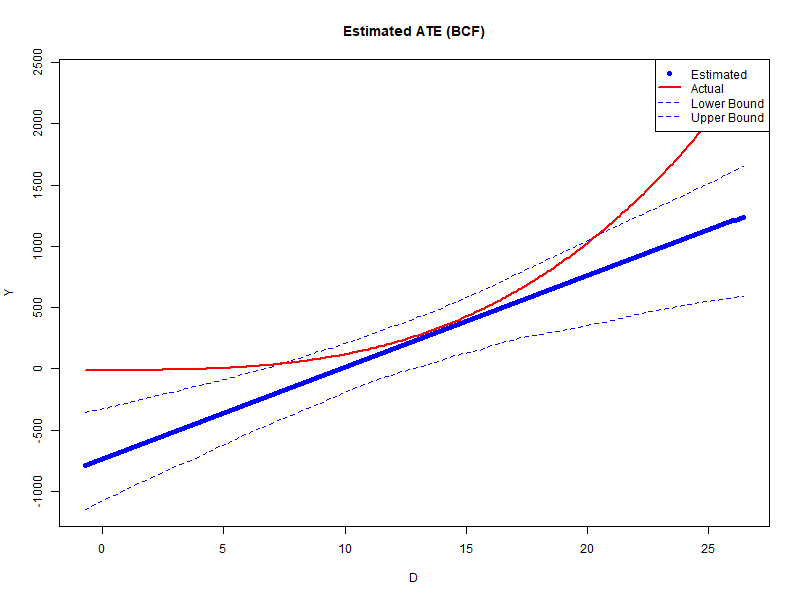}
\end{subfigure}
\hfill
\begin{subfigure}{0.45\textwidth}
    \centering
     \includegraphics[scale=0.2]{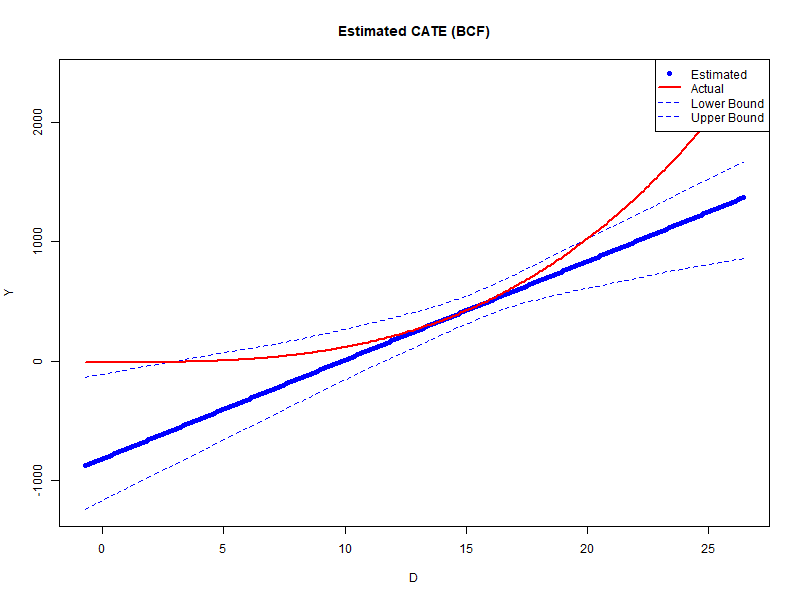}
\end{subfigure}
\caption{BCF ATE and CATE Functions Estimation for N=100 (for CATE, an Example of a Random $\mathbf{x_i}$ of a Random Simulation is used)}\label{Figure31}
\end{figure}

\begin{figure}[H]
\centering
\begin{subfigure}{0.45\textwidth}
    \centering
    \includegraphics[scale=0.2]{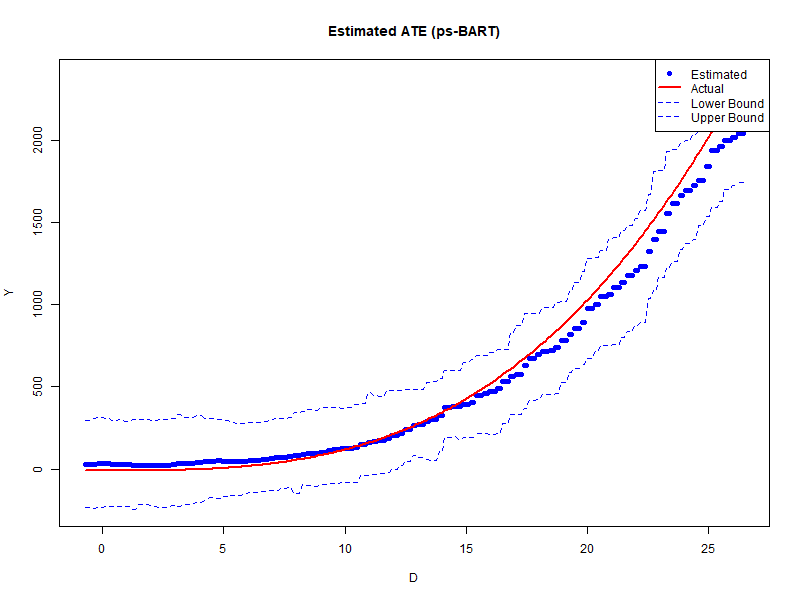}
\end{subfigure}
\hfill
\begin{subfigure}{0.45\textwidth}
    \centering
     \includegraphics[scale=0.2]{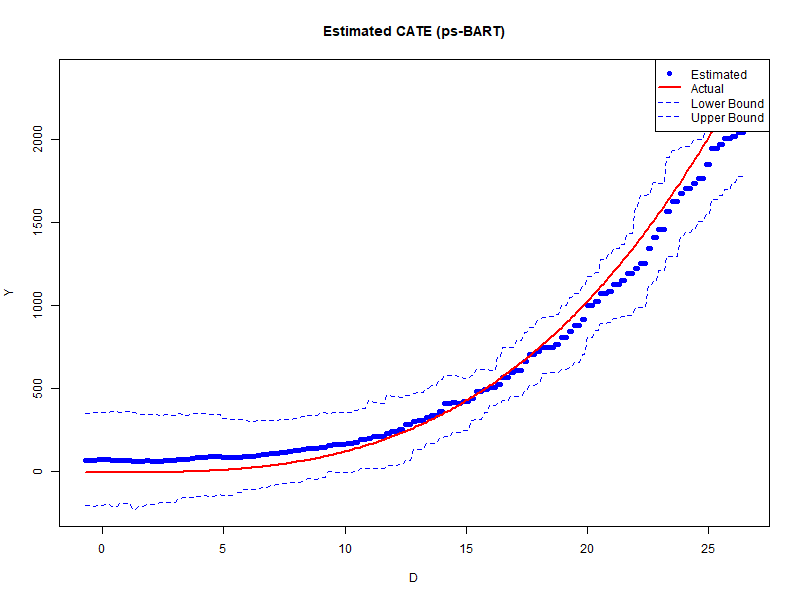}
\end{subfigure}
\caption{ps-BART ATE and CATE Functions Estimation for N=100 (for CATE, an Example of a Random $\mathbf{x_i}$ of a Random Simulation is used)}\label{Figure32}
\end{figure}

\begin{figure}[H]
\centering
\begin{subfigure}{0.45\textwidth}
    \centering
    \includegraphics[scale=0.2]{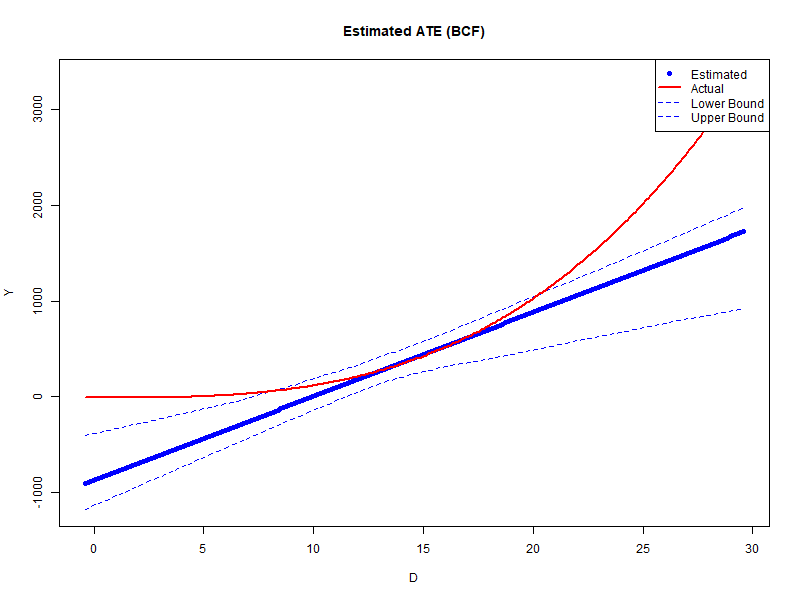}
\end{subfigure}
\hfill
\begin{subfigure}{0.45\textwidth}
    \centering
    \includegraphics[scale=0.2]{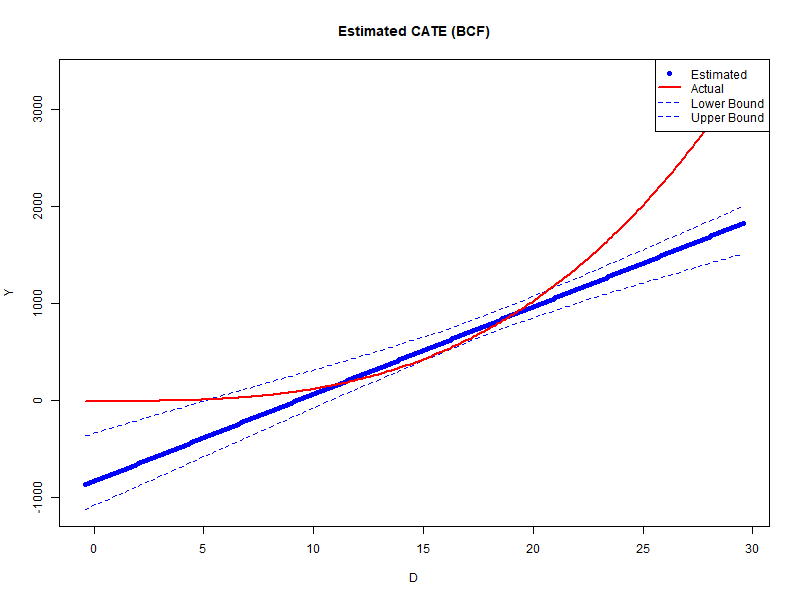}
\end{subfigure}
\caption{BCF ATE and CATE Functions Estimation for N=250 (for CATE, an Example of a Random $\mathbf{x_i}$ of a Random Simulation is used)}\label{Figure33}
\end{figure}

\begin{figure}[H]
\centering
\begin{subfigure}{0.45\textwidth}
    \centering
    \includegraphics[scale=0.2]{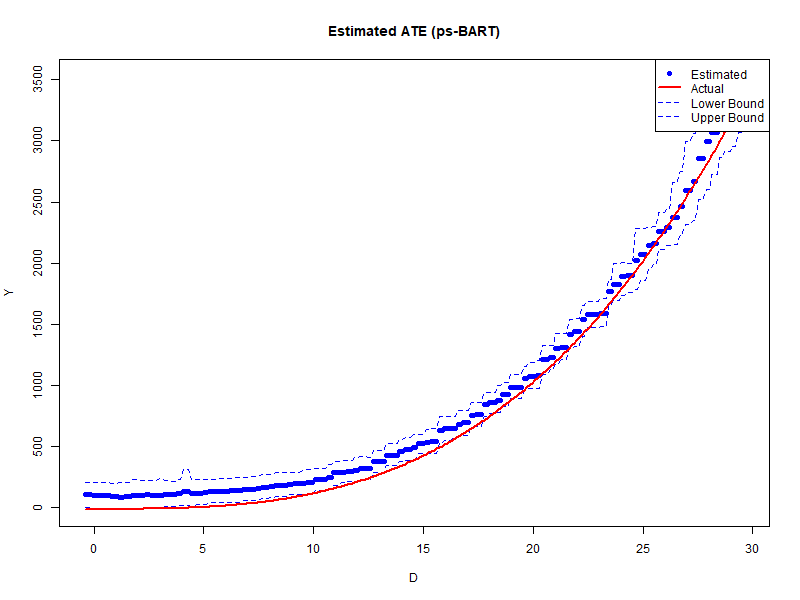}
\end{subfigure}
\hfill
\begin{subfigure}{0.45\textwidth}
    \centering
    \includegraphics[scale=0.2]{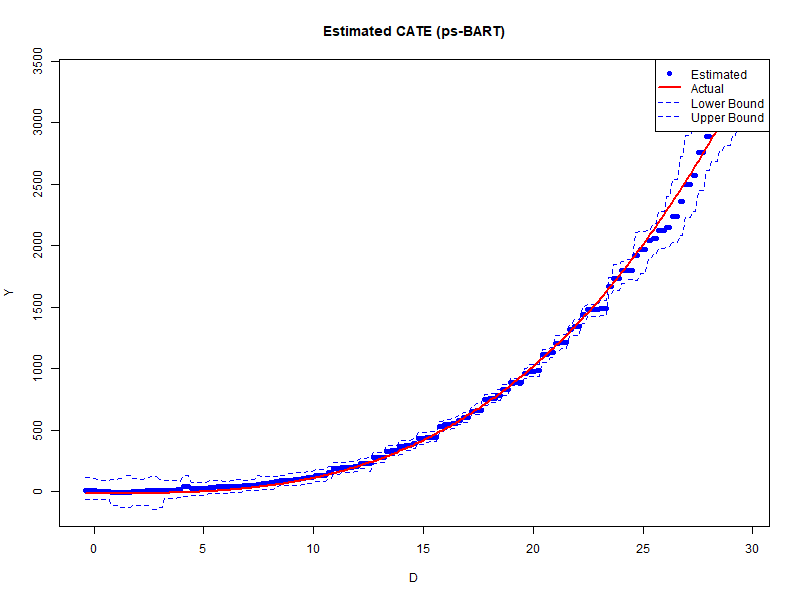}
\end{subfigure}
\caption{ps-BART ATE and CATE Functions Estimation for N=250 (for CATE, an Example of a Random $\mathbf{x_i}$ of a Random Simulation is used)}\label{Figure34}
\end{figure}

\begin{figure}[H]
\centering
\begin{subfigure}{0.45\textwidth}
    \centering
    \includegraphics[scale=0.2]{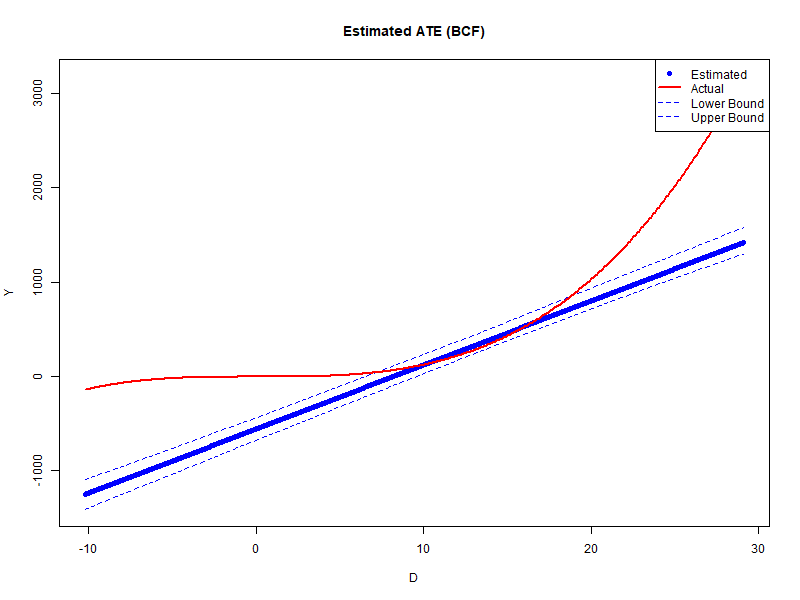}
\end{subfigure}
\hfill
\begin{subfigure}{0.45\textwidth}
    \centering
    \includegraphics[scale=0.2]{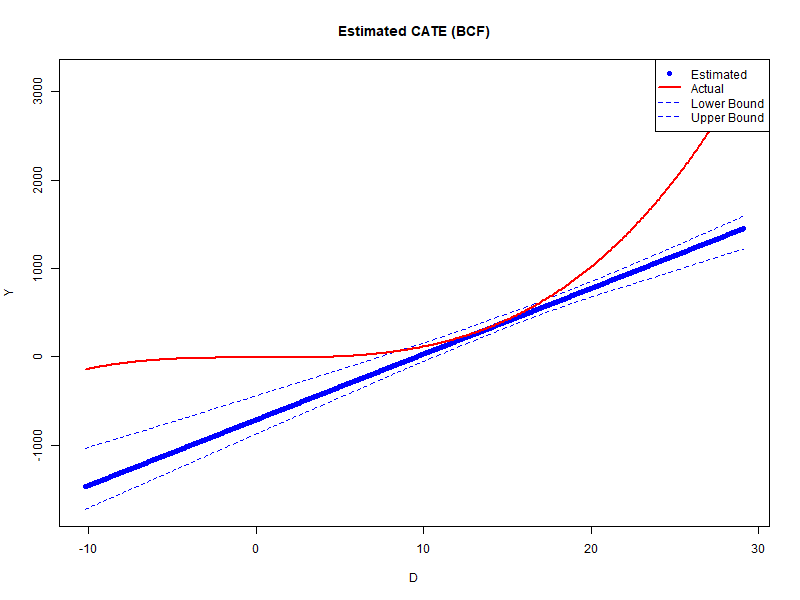}
\end{subfigure}
\caption{BCF ATE and CATE Functions Estimation for N=500 (for CATE, an Example of a Random $\mathbf{x_i}$ of a Random Simulation is used)}\label{Figure35}
\end{figure}

\begin{figure}[H]
\centering
\begin{subfigure}{0.45\textwidth}
    \centering
    \includegraphics[scale=0.2]{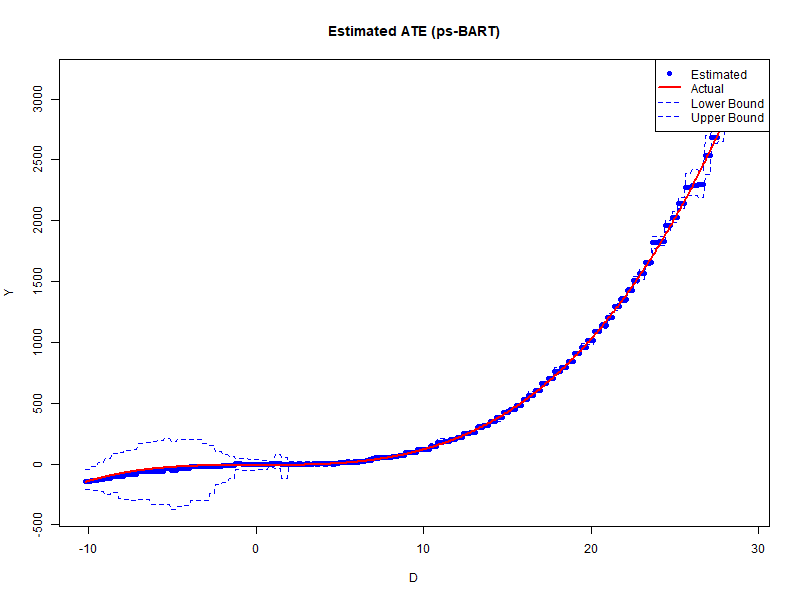}
\end{subfigure}
\hfill
\begin{subfigure}{0.45\textwidth}
    \centering
    \includegraphics[scale=0.2]{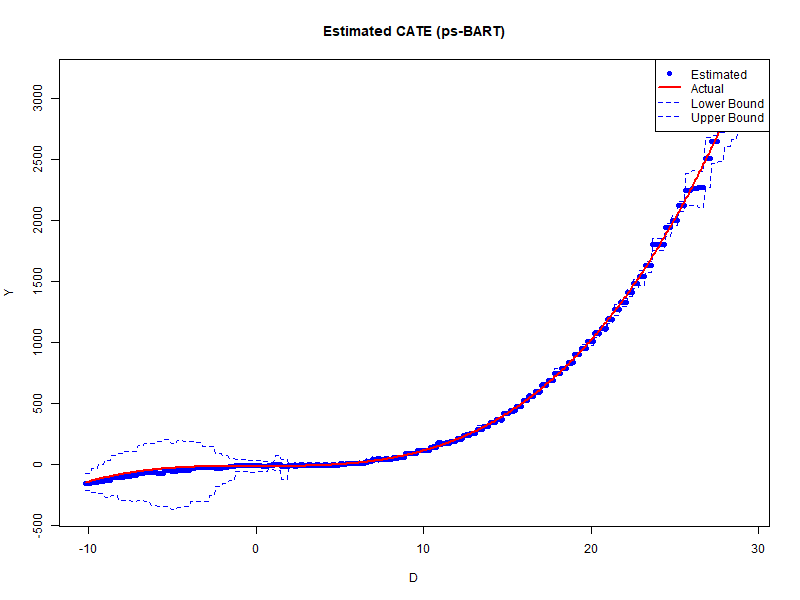}
\end{subfigure}
\caption{ps-BART ATE and CATE Functions Estimation for N=500 (for CATE, an Example of a Random $\mathbf{x_i}$ of a Random Simulation is used)}\label{Figure36}
\end{figure}

Figures \ref{Figure31}, \ref{Figure32}, \ref{Figure33}, \ref{Figure34}, \ref{Figure35}, and \ref{Figure36} graphically illustrate the results of Table \ref{Table21}. The misspecification of the benchmark model can be clearly seen in the figures.

\begin{table}[H]
\centering
\caption{Metric Results}\label{Table21}
\begin{tabular}{llcc}
\hline
$n$ & \textbf{Metric} & \textbf{BCF (Mean $\pm$ SD)} & \textbf{ps-BART (Mean $\pm$ SD)} \\
\hline
100 & RMSE\(_{ATE}\) & $276 \pm 69$ & $92 \pm 39$ \\
& MAE\(_{ATE}\)  & $150 \pm 32$ & $53 \pm 21$ \\
& MAPE\(_{ATE}\) & $14 \pm 56$ & $3 \pm 12$ \\
& Len\(_{ATE}\) & $406 \pm 72$ & $424 \pm 118$ \\
& Cover\(_{ATE}\) & $0.801 \pm 0.076$ & $0.986 \pm 0.010$ \\
& RMSE\(_{CATE}\) & $485 \pm 208$ & $171 \pm 99$ \\
& MAE\(_{CATE}\)  & $337 \pm 122$ & $117 \pm 61$ \\
& MAPE\(_{CATE}\) & $56 \pm 92$ & $16 \pm 24$ \\
& Len\(_{CATE}\) & $456 \pm 96$ & $532 \pm 184$ \\
& Cover\(_{CATE}\) & $0.510 \pm 0.060$ & $0.940 \pm 0.045$ \\
& SEC\(_{ATE}\)  & $0.028 \pm 0.033$ & $0.001 \pm 0.001$ \\
& AEC\(_{ATE}\)  & $0.149 \pm 0.076$ & $0.036 \pm 0.009$ \\
& SEC\(_{CATE}\)  & $0.197 \pm 0.051$ & $0.002 \pm 0.002$ \\
& AEC\(_{CATE}\)  & $0.440 \pm 0.060$ & $0.038 \pm 0.025$ \\
\hline
250 & RMSE\(_{ATE}\) & $281 \pm 94$ & $30 \pm 21$ \\
& MAE\(_{ATE}\)  & $147 \pm 32$ & $18 \pm 16$ \\
& MAPE\(_{ATE}\) & $10 \pm 17$ & $0.4 \pm 0.5$ \\
& Len\(_{ATE}\) & $297 \pm 70$ & $106 \pm 40$ \\
& Cover\(_{ATE}\) & $0.719 \pm 0.111$ & $0.952 \pm 0.152$ \\
& RMSE\(_{CATE}\) & $736 \pm 588$ & $78 \pm 81$ \\
& MAE\(_{CATE}\)  & $486 \pm 351$ & $46 \pm 46$ \\
& MAPE\(_{CATE}\) & $78 \pm 178$ & $3 \pm 3$ \\
& Len\(_{CATE}\) & $352 \pm 74$ & $275 \pm 246$ \\
& Cover\(_{CATE}\) & $0.377 \pm 0.043$ & $0.967 \pm 0.033$ \\
& SEC\(_{ATE}\)  & $0.066 \pm 0.084$ & $0.023 \pm 0.125$ \\
& AEC\(_{ATE}\)  & $0.231 \pm 0.111$ & $0.055 \pm 0.142$ \\
& SEC\(_{CATE}\)  & $0.331 \pm 0.050$ & $0.001 \pm 0.004$ \\
& AEC\(_{CATE}\)  & $0.573 \pm 0.043$ & $0.030 \pm 0.022$ \\
\hline
500 & RMSE\(_{ATE}\) & $272 \pm 68$ & $19 \pm 13$ \\
& MAE\(_{ATE}\)  & $143 \pm 27$ & $10 \pm 3$ \\
& MAPE\(_{ATE}\) & $17 \pm 66$ & $0.2 \pm 0.5$ \\
& Len\(_{ATE}\) & $227 \pm 47$ & $44 \pm 21$ \\
& Cover\(_{ATE}\) & $0.656 \pm 0.124$ & $0.863 \pm 0.050$ \\
& RMSE\(_{CATE}\) & $864 \pm 606$ & $73 \pm 108$ \\
& MAE\(_{CATE}\)  & $563 \pm 362$ & $40 \pm 59$ \\
& MAPE\(_{CATE}\) & $79 \pm 111$ & $2 \pm 2$ \\
& Len\(_{CATE}\) & $318 \pm 56$ & $192 \pm 196$ \\
& Cover\(_{CATE}\) & $0.344 \pm 0.040$ & $0.884 \pm 0.038$ \\
& SEC\(_{ATE}\)  & $0.102 \pm 0.094$ & $0.010 \pm 0.010$ \\
& AEC\(_{ATE}\)  & $0.294 \pm 0.124$ & $0.090 \pm 0.046$ \\
& SEC\(_{CATE}\)  & $0.369 \pm 0.050$ & $0.006 \pm 0.008$ \\
& AEC\(_{CATE}\)  & $0.606 \pm 0.040$ & $0.066 \pm 0.038$ \\
\hline
\end{tabular}
\end{table}

\begin{landscape}
\begin{table}[h!]
\centering
\caption{Statistical Test Results: p-values for Different Metrics (n=100)}\label{Table22}
\begin{tabular}{lccccc}
\hline
\textbf{Metric} & \textbf{Fligner-Policello Test} & \textbf{Mann-Whitney U Test} & \textbf{Kruskal-Wallis H Test} & \textbf{Levene's Test} & \textbf{Brown-Forsythe Test} \\
\hline
RMSE\(_{ATE}\)  & $0$                    & \textit{N/A}    & \textit{N/A}   & $2.85 \times 10^{-5}$  & $2.16 \times 10^{-5}$  \\
MAE\(_{ATE}\)   & $0$                    & \textit{N/A}    & \textit{N/A}   & $3.39 \times 10^{-5}$  & $7.76 \times 10^{-5}$  \\
MAPE\(_{ATE}\)  & $9.20 \times 10^{-67}$  & \textit{N/A}    & \textit{N/A}   & $0.0228$               & $0.1352$               \\
Len\(_{ATE}\)   & $0.8339$               & \textit{N/A}    & \textit{N/A}   & $6.13 \times 10^{-4}$  & $0.00453$              \\
RMSE\(_{CATE}\) & $2.37 \times 10^{-258}$ & \textit{N/A}    & \textit{N/A}   & $1.76 \times 10^{-6}$  & $1.16 \times 10^{-4}$  \\
MAE\(_{CATE}\)  & $0$                    & \textit{N/A}    & \textit{N/A}   & $6.15 \times 10^{-6}$  & $2.38 \times 10^{-4}$  \\
MAPE\(_{CATE}\) & $4.20 \times 10^{-105}$ & \textit{N/A}    & \textit{N/A}   & $0.0146$               & $0.0378$               \\
Len\(_{CATE}\)  & $0.00227$              & \textit{N/A}    & \textit{N/A}   & $7.97 \times 10^{-5}$  & $0.00290$              \\
SEC\(_{ATE}\)   & $1.22 \times 10^{-174}$ & \textit{N/A}    & \textit{N/A}   & $4.48 \times 10^{-12}$ & $2.45 \times 10^{-9}$  \\
AEC\(_{ATE}\)   & $1.22 \times 10^{-174}$ & \textit{N/A}    & \textit{N/A}   & $9.66 \times 10^{-18}$ & $3.47 \times 10^{-17}$ \\
SEC\(_{CATE}\)  & $0$                    & \textit{N/A}    & \textit{N/A}   & $5.22 \times 10^{-26}$ & $4.87 \times 10^{-26}$ \\
AEC\(_{CATE}\)  & $0$                    & \textit{N/A}    & \textit{N/A}   & $4.01 \times 10^{-10}$ & $7.95 \times 10^{-10}$ \\
\hline
\end{tabular}
\end{table}
\end{landscape}

\begin{landscape}
\begin{table}[h!]
\centering
\caption{Statistical Test Results: p-values for Different Metrics (n=250)}\label{Table23}
\begin{tabular}{lccccc}
\hline
\textbf{Metric} & \textbf{Fligner-Policello Test} & \textbf{Mann-Whitney U Test} & \textbf{Kruskal-Wallis H Test} & \textbf{Levene's Test} & \textbf{Brown-Forsythe Test} \\
\hline
RMSE\(_{ATE}\)  & $0$                    & \textit{N/A}    & \textit{N/A}   & $3.76 \times 10^{-7}$  & $8.95 \times 10^{-7}$  \\
MAE\(_{ATE}\)   & $0$                    & \textit{N/A}    & \textit{N/A}   & $3.10 \times 10^{-9}$  & $2.40 \times 10^{-8}$  \\
MAPE\(_{ATE}\)  & $0$                    & \textit{N/A}    & \textit{N/A}   & $1.64 \times 10^{-6}$  & $3.62 \times 10^{-4}$  \\
Len\(_{ATE}\)   & $0$                    & \textit{N/A}    & \textit{N/A}   & $9.29 \times 10^{-5}$  & $8.68 \times 10^{-5}$  \\
RMSE\(_{CATE}\) & $0$                    & \textit{N/A}    & \textit{N/A}   & $1.98 \times 10^{-8}$  & $8.03 \times 10^{-6}$  \\
MAE\(_{CATE}\)  & $0$                    & \textit{N/A}    & \textit{N/A}   & $2.24 \times 10^{-8}$  & $7.61 \times 10^{-6}$  \\
MAPE\(_{CATE}\) & $0$                    & \textit{N/A}    & \textit{N/A}   & $0.00147$              & $0.0217$               \\
Len\(_{CATE}\)  & $9.93 \times 10^{-22}$  & \textit{N/A}    & \textit{N/A}   & $9.85 \times 10^{-4}$  & $0.0163$               \\
SEC\(_{ATE}\)   & \textit{N/A}           & $1.46 \times 10^{-30}$ & $1.44 \times 10^{-30}$  & $0.9278$               & $0.2347$               \\
AEC\(_{ATE}\)   & \textit{N/A}           & $1.46 \times 10^{-30}$ & $1.44 \times 10^{-30}$  & $0.0566$               & $0.0154$               \\
SEC\(_{CATE}\)  & $0$                    & \textit{N/A}    & \textit{N/A}   & $2.17 \times 10^{-28}$ & $3.20 \times 10^{-27}$ \\
AEC\(_{CATE}\)  & $0$                    & \textit{N/A}    & \textit{N/A}   & $9.08 \times 10^{-12}$ & $1.28 \times 10^{-11}$ \\
\hline
\end{tabular}
\end{table}
\end{landscape}

\begin{landscape}
\begin{table}[h!]
\centering
\caption{Statistical Test Results: p-values for Different Metrics (n=500)}\label{Table24}
\begin{tabular}{lccccc}
\hline
\textbf{Metric} & \textbf{Fligner-Policello Test} & \textbf{Mann-Whitney U Test} & \textbf{Kruskal-Wallis H Test} & \textbf{Levene's Test} & \textbf{Brown-Forsythe Test} \\
\hline
RMSE\(_{ATE}\)  & $0$                    & \textit{N/A}    & \textit{N/A}   & $2.01 \times 10^{-14}$  & $1.76 \times 10^{-11}$  \\
MAE\(_{ATE}\)   & $0$                    & \textit{N/A}    & \textit{N/A}   & $2.13 \times 10^{-15}$  & $6.70 \times 10^{-12}$  \\
MAPE\(_{ATE}\)  & $0$                    & \textit{N/A}    & \textit{N/A}   & $0.00815$               & $0.0625$                \\
Len\(_{ATE}\)   & $0$                    & \textit{N/A}    & \textit{N/A}   & $2.96 \times 10^{-14}$  & $9.60 \times 10^{-14}$  \\
RMSE\(_{CATE}\) & $0$                    & \textit{N/A}    & \textit{N/A}   & $8.65 \times 10^{-11}$  & $1.16 \times 10^{-6}$   \\
MAE\(_{CATE}\)  & $0$                    & \textit{N/A}    & \textit{N/A}   & $7.10 \times 10^{-11}$  & $7.80 \times 10^{-7}$   \\
MAPE\(_{CATE}\) & $0$                    & \textit{N/A}    & \textit{N/A}   & $6.76 \times 10^{-5}$   & $0.00175$               \\
Len\(_{CATE}\)  & $1.35 \times 10^{-31}$  & \textit{N/A}    & \textit{N/A}   & $1.67 \times 10^{-6}$   & $0.00268$               \\
SEC\(_{ATE}\)   & $0$                    & \textit{N/A}    & \textit{N/A}   & $3.38 \times 10^{-15}$  & $4.39 \times 10^{-10}$  \\
AEC\(_{ATE}\)   & $0$                    & \textit{N/A}    & \textit{N/A}   & $7.40 \times 10^{-11}$  & $8.15 \times 10^{-9}$   \\
SEC\(_{CATE}\)  & $0$                    & \textit{N/A}    & \textit{N/A}   & $1.90 \times 10^{-23}$  & $2.47 \times 10^{-22}$  \\
AEC\(_{CATE}\)  & \textit{N/A}           & $2.56 \times 10^{-34}$ & $2.52 \times 10^{-34}$  & $0.1751$               & $0.1873$               \\
\hline
\end{tabular}
\end{table}
\end{landscape}

\subsubsection{Specification 3}

Table \ref{Table25} shows the results for Specification 3. Based on these results, it can be easily seen that the ps-BART model is superior in both point-wise estimation and uncertainty estimation, while also being more robust than the benchmark model. The superiority of the proposed model is also statistically significant based on the results of the statistical tests found in Tables \ref{Table22}, \ref{Table23}, and \ref{Table24}. Interestingly, the uncertainty estimation performance for both the ATE and CATE functions estimation of the proposed model considerably decreased when compared to Specification 1 and 2. It could be theorized that this is due to the addition of the $3\sqrt{2}C_3^2$ in the determination of the treatment value, which increases the complexity and nonlinearity of the problem.

\begin{figure}[H]
\centering
\begin{subfigure}{0.45\textwidth}
    \centering
     \includegraphics[scale=0.2]{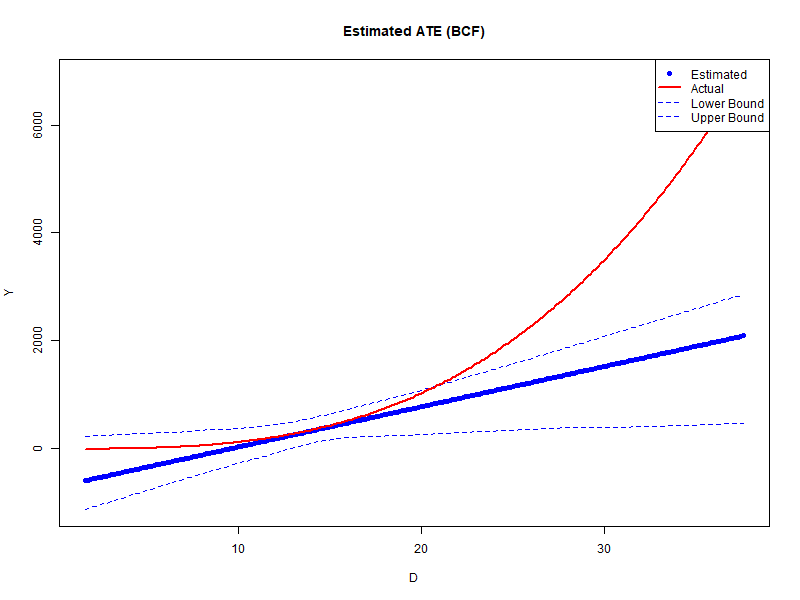}
\end{subfigure}
\hfill
\begin{subfigure}{0.45\textwidth}
    \centering
     \includegraphics[scale=0.2]{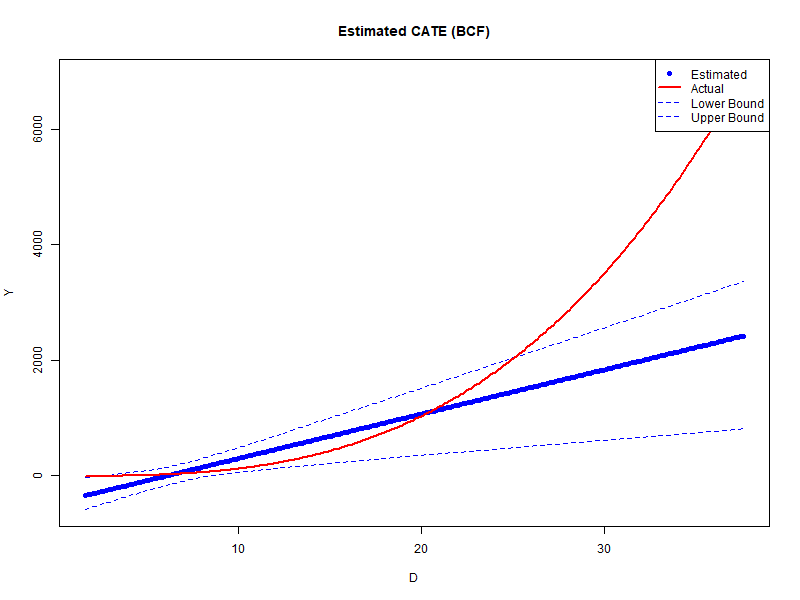}
\end{subfigure}
\caption{BCF ATE and CATE Functions Estimation for N=100 (for CATE, an Example of a Random $\mathbf{x_i}$ of a Random Simulation is used)}\label{Figure37}
\end{figure}

\begin{figure}[H]
\centering
\begin{subfigure}{0.45\textwidth}
    \centering
    \includegraphics[scale=0.2]{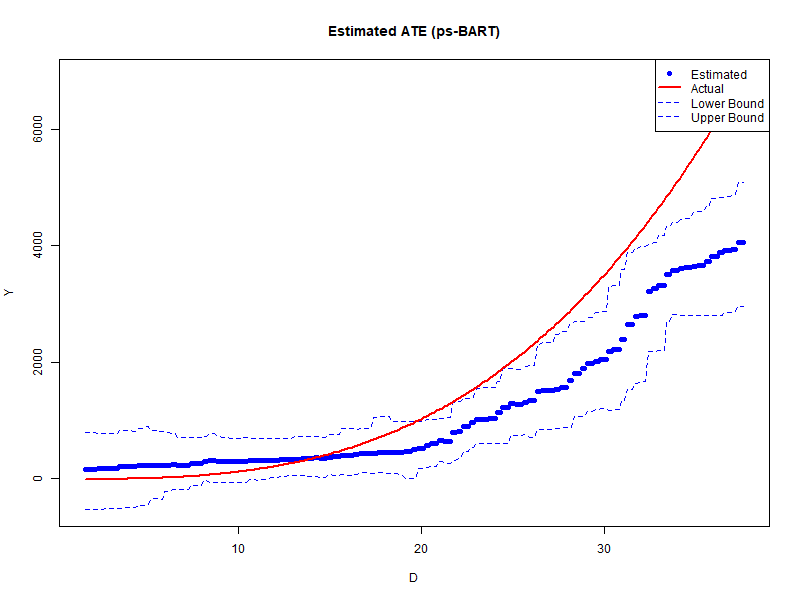}
\end{subfigure}
\hfill
\begin{subfigure}{0.45\textwidth}
    \centering
     \includegraphics[scale=0.2]{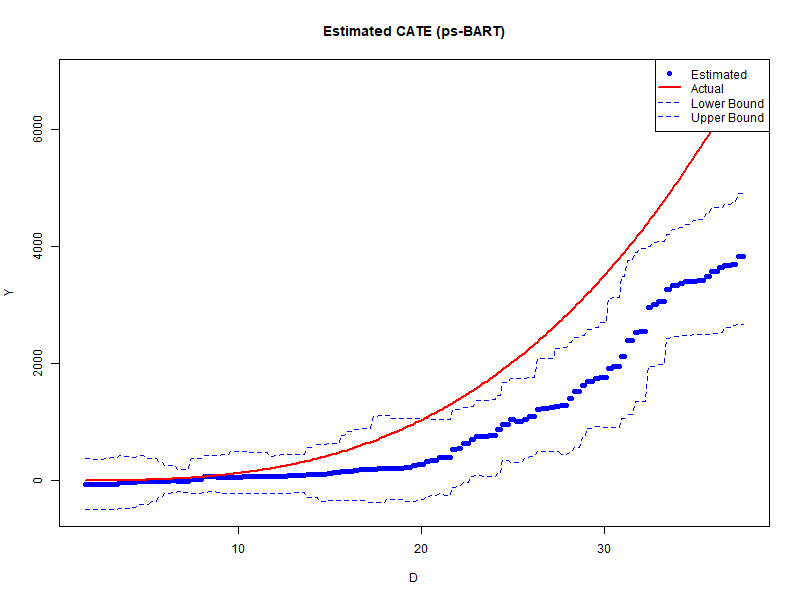}
\end{subfigure}
\caption{ps-BART ATE and CATE Functions Estimation for N=100 (for CATE, an Example of a Random $\mathbf{x_i}$ of a Random Simulation is used)}\label{Figure38}
\end{figure}

\begin{figure}[H]
\centering
\begin{subfigure}{0.45\textwidth}
    \centering
    \includegraphics[scale=0.2]{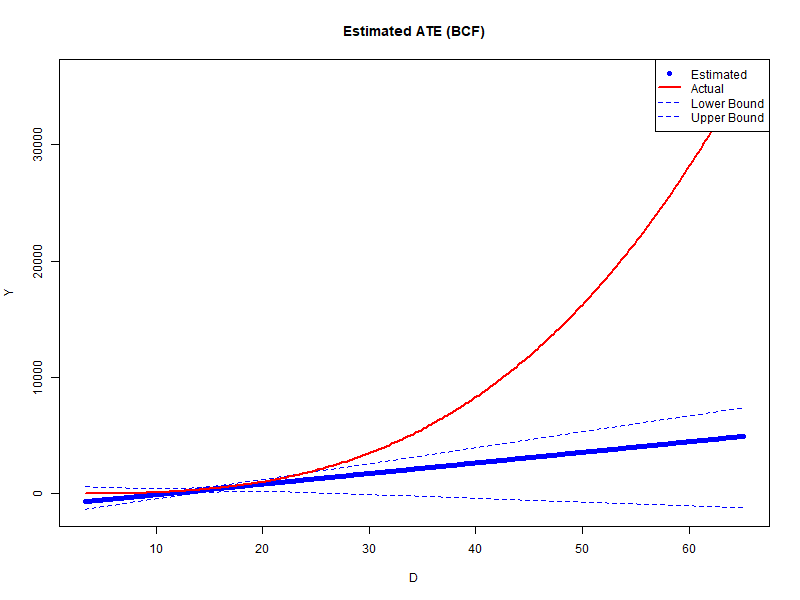}
\end{subfigure}
\hfill
\begin{subfigure}{0.45\textwidth}
    \centering
    \includegraphics[scale=0.2]{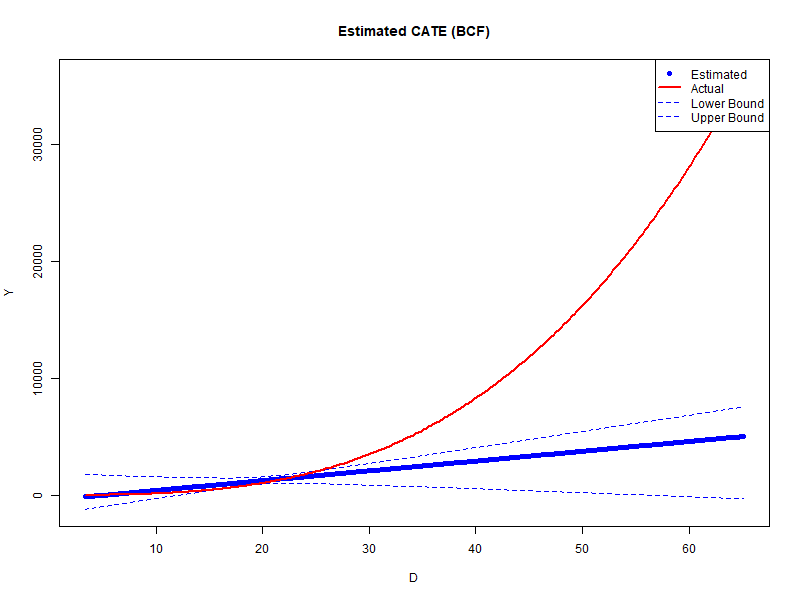}
\end{subfigure}
\caption{BCF ATE and CATE Functions Estimation for N=250 (for CATE, an Example of a Random $\mathbf{x_i}$ of a Random Simulation is used)}\label{Figure39}
\end{figure}

\begin{figure}[H]
\centering
\begin{subfigure}{0.45\textwidth}
    \centering
    \includegraphics[scale=0.2]{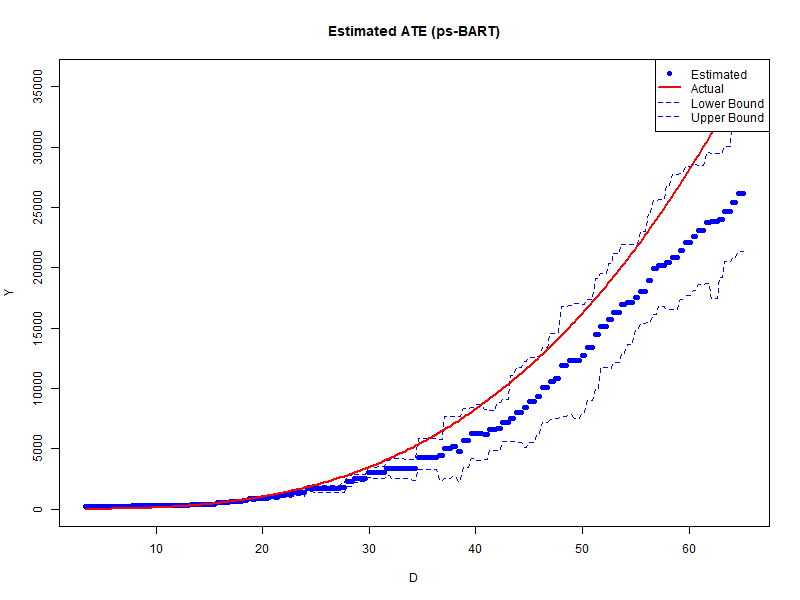}
\end{subfigure}
\hfill
\begin{subfigure}{0.45\textwidth}
    \centering
    \includegraphics[scale=0.2]{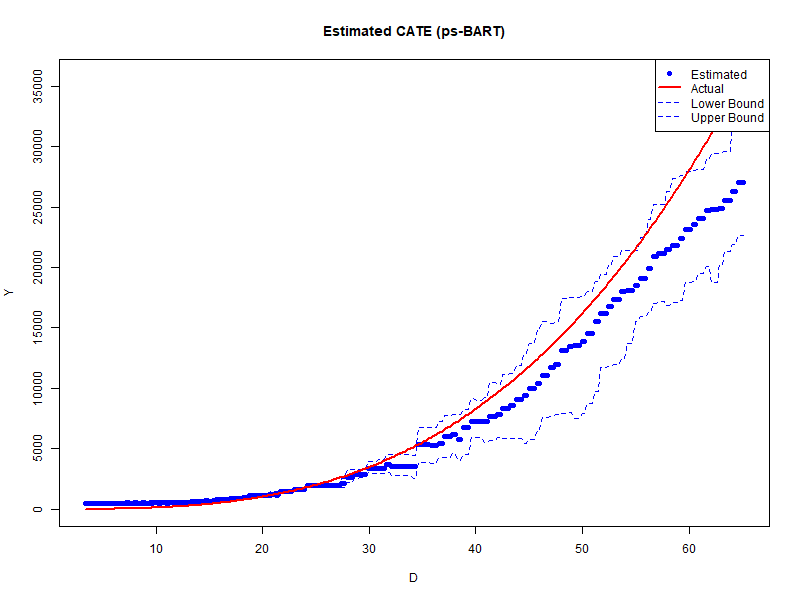}
\end{subfigure}
\caption{ps-BART ATE and CATE Functions Estimation for N=250 (for CATE, an Example of a Random $\mathbf{x_i}$ of a Random Simulation is used)}\label{Figure40}
\end{figure}

\begin{figure}[H]
\centering
\begin{subfigure}{0.45\textwidth}
    \centering
    \includegraphics[scale=0.2]{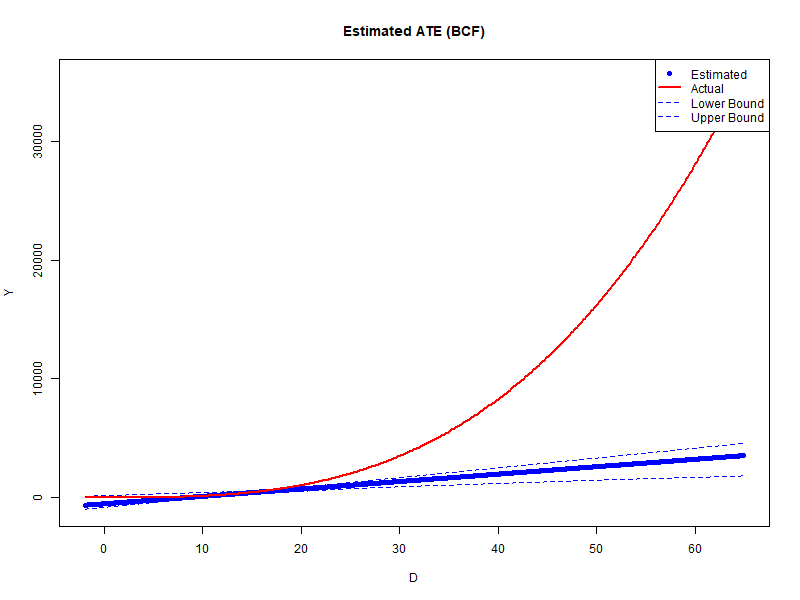}
\end{subfigure}
\hfill
\begin{subfigure}{0.45\textwidth}
    \centering
    \includegraphics[scale=0.2]{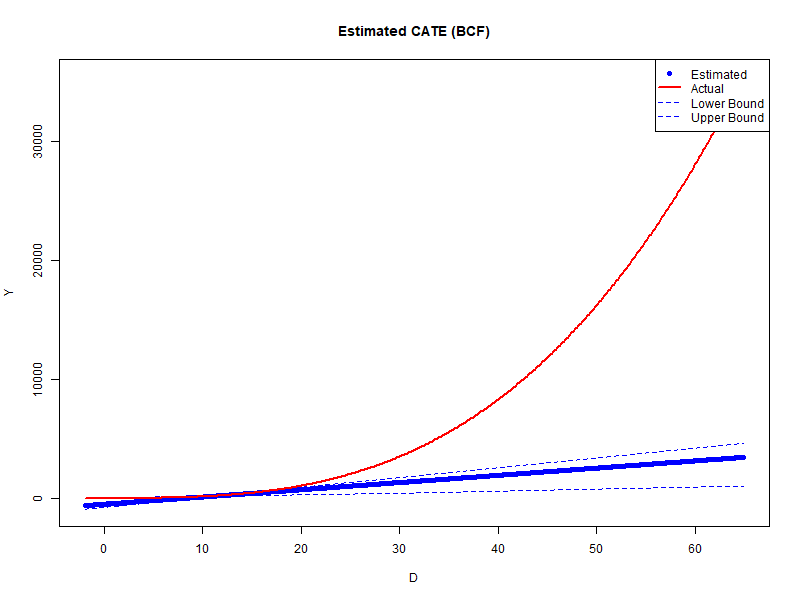}
\end{subfigure}
\caption{BCF ATE and CATE Functions Estimation for N=500 (for CATE, an Example of a Random $\mathbf{x_i}$ of a Random Simulation is used)}\label{Figure41}
\end{figure}

\begin{figure}[H]
\centering
\begin{subfigure}{0.45\textwidth}
    \centering
    \includegraphics[scale=0.2]{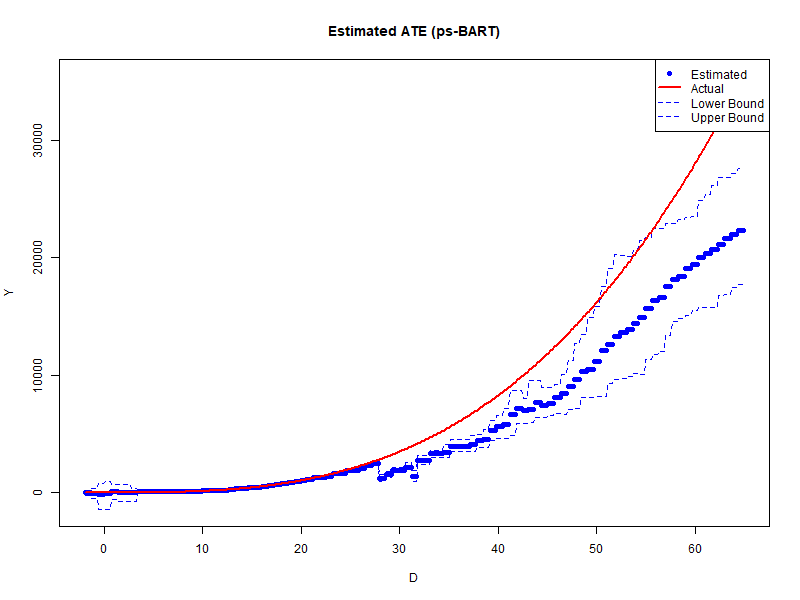}
\end{subfigure}
\hfill
\begin{subfigure}{0.45\textwidth}
    \centering
    \includegraphics[scale=0.2]{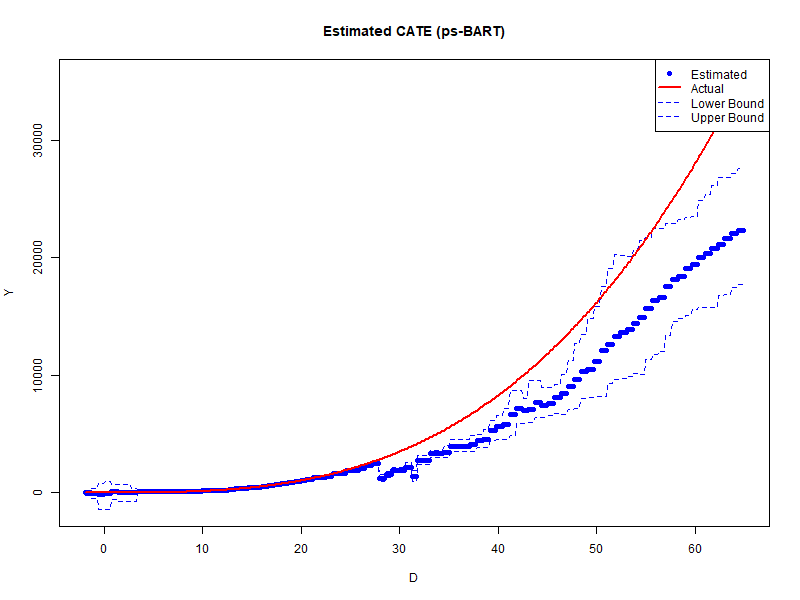}
\end{subfigure}
\caption{ps-BART ATE and CATE Functions Estimation for N=500 (for CATE, an Example of a Random $\mathbf{x_i}$ of a Random Simulation is used)}\label{Figure42}
\end{figure}

Figures \ref{Figure37}, \ref{Figure38}, \ref{Figure39}, \ref{Figure40}, \ref{Figure41}, and \ref{Figure42} graphically illustrate the results of Table \ref{Table25}. The misspecification of the BCF model and the poor uncertainty estimation performance of the proposed model can be clearly seen in the figures.

\begin{table}[H]
\centering
\caption{Metric Results}\label{Table25}
\begin{tabular}{llcc}
\hline
$n$ & \textbf{Metric} & \textbf{BCF (Mean $\pm$ SD)} & \textbf{ps-BART (Mean $\pm$ SD)} \\
\hline
100 & RMSE\(_{ATE}\) & $1410 \pm 793$ & $805 \pm 411$ \\
& MAE\(_{ATE}\)  & $473 \pm 156$ & $407 \pm 159$ \\
& MAPE\(_{ATE}\) & $17 \pm 63$ & $7 \pm 24$ \\
& Len\(_{ATE}\) & $1021 \pm 363$ & $1845 \pm 1041$ \\
& Cover\(_{ATE}\) & $0.844 \pm 0.066$ & $0.921 \pm 0.067$ \\
& RMSE\(_{CATE}\) & $3471 \pm 2689$ & $1849 \pm 1315$ \\
& MAE\(_{CATE}\)  & $2167 \pm 1653$ & $1307 \pm 903$ \\
& MAPE\(_{CATE}\) & $64 \pm 138$ & $47 \pm 36$ \\
& Len\(_{CATE}\) & $1765 \pm 914$ & $2961 \pm 2105$ \\
& Cover\(_{CATE}\) & $0.474 \pm 0.065$ & $0.707 \pm 0.109$ \\
& SEC\(_{ATE}\)  & $0.015 \pm 0.018$ & $0.005 \pm 0.012$ \\
& AEC\(_{ATE}\)  & $0.106 \pm 0.065$ & $0.049 \pm 0.054$ \\
& SEC\(_{CATE}\)  & $0.231 \pm 0.062$ & $0.071 \pm 0.056$ \\
& AEC\(_{CATE}\)  & $0.476 \pm 0.065$ & $0.243 \pm 0.109$ \\
\hline
250 & RMSE\(_{ATE}\) & $1727 \pm 798$ & $546 \pm 321$ \\
& MAE\(_{ATE}\)  & $495 \pm 109$ & $198 \pm 101$ \\
& MAPE\(_{ATE}\) & $9 \pm 15$ & $3 \pm 5$ \\
& Len\(_{ATE}\) & $717 \pm 270$ & $390 \pm 142$ \\
& Cover\(_{ATE}\) & $0.760 \pm 0.081$ & $0.703 \pm 0.150$ \\
& RMSE\(_{CATE}\) & $6508 \pm 4667$ & $1867 \pm 1398$ \\
& MAE\(_{CATE}\)  & $4080 \pm 2923$ & $1187 \pm 836$ \\
& MAPE\(_{CATE}\) & $57 \pm 104$ & $25 \pm 41$ \\
& Len\(_{CATE}\) & $1642 \pm 968$ & $2344 \pm 2665$ \\
& Cover\(_{CATE}\) & $0.335 \pm 0.043$ & $0.536 \pm 0.143$ \\
& SEC\(_{ATE}\)  & $0.043 \pm 0.031$ & $0.083 \pm 0.108$ \\
& AEC\(_{ATE}\)  & $0.190 \pm 0.080$ & $0.247 \pm 0.149$ \\
& SEC\(_{CATE}\)  & $0.380 \pm 0.052$ & $0.192 \pm 0.122$ \\
& AEC\(_{CATE}\)  & $0.615 \pm 0.043$ & $0.414 \pm 0.143$ \\
\hline
500 & RMSE\(_{ATE}\) & $1933 \pm 605$ & $425 \pm 271$ \\
& MAE\(_{ATE}\)  & $510 \pm 87$ & $97 \pm 56$ \\
& MAPE\(_{ATE}\) & $4 \pm 4$ & $1 \pm 2$ \\
& Len\(_{ATE}\) & $476 \pm 125$ & $192 \pm 80$ \\
& Cover\(_{ATE}\) & $0.661 \pm 0.053$ & $0.801 \pm 0.057$ \\
& RMSE\(_{CATE}\) & $9549 \pm 4961$ & $1974 \pm 1475$ \\
& MAE\(_{CATE}\)  & $6059 \pm 3162$ & $1184 \pm 934$ \\
& MAPE\(_{CATE}\) & $44 \pm 30$ & $11 \pm 10$ \\
& Len\(_{CATE}\) & $1406 \pm 486$ & $2325 \pm 1908$ \\
& Cover\(_{CATE}\) & $0.247 \pm 0.023$ & $0.594 \pm 0.137$ \\
& SEC\(_{ATE}\)  & $0.086 \pm 0.031$ & $0.025 \pm 0.017$ \\
& AEC\(_{ATE}\)  & $0.289 \pm 0.053$ & $0.149 \pm 0.057$ \\
& SEC\(_{CATE}\)  & $0.494 \pm 0.033$ & $0.145 \pm 0.104$ \\
& AEC\(_{CATE}\)  & $0.703 \pm 0.023$ & $0.356 \pm 0.137$ \\
\hline
\end{tabular}
\end{table}

\begin{landscape}
\begin{table}[h!]
\centering
\caption{Statistical Test Results: p-values for Different Metrics (n=100)}\label{Table26}
\begin{tabular}{lccccc}
\hline
\textbf{Metric} & \textbf{Fligner-Policello Test} & \textbf{Mann-Whitney U Test} & \textbf{Kruskal-Wallis H Test} & \textbf{Levene's Test} & \textbf{Brown-Forsythe Test} \\
\hline
RMSE\(_{ATE}\)  & $6.77 \times 10^{-31}$  & \textit{N/A}    & \textit{N/A}   & $4.65 \times 10^{-4}$  & $0.00342$               \\
MAE\(_{ATE}\)   & \textit{N/A}            & $4.82 \times 10^{-4}$  & $4.80 \times 10^{-4}$  & $0.9608$               & $0.9225$                \\
MAPE\(_{ATE}\)  & \textit{N/A}            & $4.23 \times 10^{-6}$  & $4.21 \times 10^{-6}$  & $0.0638$               & $0.2209$                \\
Len\(_{ATE}\)   & $7.81 \times 10^{-35}$  & \textit{N/A}    & \textit{N/A}   & $4.89 \times 10^{-7}$  & $6.22 \times 10^{-6}$   \\
RMSE\(_{CATE}\) & $1.46 \times 10^{-20}$  & \textit{N/A}    & \textit{N/A}   & $4.29 \times 10^{-4}$  & $0.00344$               \\
MAE\(_{CATE}\)  & $3.04 \times 10^{-13}$  & \textit{N/A}    & \textit{N/A}   & $0.00201$              & $0.0112$                \\
MAPE\(_{CATE}\) & \textit{N/A}            & $0.2742$        & $0.2737$       & $0.0856$               & $0.2845$                \\
Len\(_{CATE}\)  & $1.63 \times 10^{-14}$  & \textit{N/A}    & \textit{N/A}   & $7.51 \times 10^{-5}$  & $0.00148$               \\
SEC\(_{ATE}\)   & $3.84 \times 10^{-18}$  & \textit{N/A}    & \textit{N/A}   & $1.10 \times 10^{-4}$  & $3.16 \times 10^{-4}$   \\
AEC\(_{ATE}\)   & $3.84 \times 10^{-18}$  & \textit{N/A}    & \textit{N/A}   & $0.00193$              & $0.00135$               \\
SEC\(_{CATE}\)  & \textit{N/A}            & $4.76 \times 10^{-30}$ & $4.70 \times 10^{-30}$  & $0.2349$               & $0.2413$                \\
AEC\(_{CATE}\)  & $0$                    & \textit{N/A}    & \textit{N/A}   & $2.31 \times 10^{-6}$  & $2.83 \times 10^{-6}$   \\
\hline
\end{tabular}
\end{table}
\end{landscape}

\begin{landscape}
\begin{table}[h!]
\centering
\caption{Statistical Test Results: p-values for Different Metrics (n=250)}\label{Table27}
\begin{tabular}{lccccc}
\hline
\textbf{Metric} & \textbf{Fligner-Policello Test} & \textbf{Mann-Whitney U Test} & \textbf{Kruskal-Wallis H Test} & \textbf{Levene's Test} & \textbf{Brown-Forsythe Test} \\
\hline
RMSE\(_{ATE}\)  & $0$                    & \textit{N/A}    & \textit{N/A}   & $1.86 \times 10^{-5}$  & $1.16 \times 10^{-4}$   \\
MAE\(_{ATE}\)   & \textit{N/A}            & $3.57 \times 10^{-31}$ & $3.52 \times 10^{-31}$ & $0.1515$               & $0.1177$                \\
MAPE\(_{ATE}\)  & $6.24 \times 10^{-16}$  & \textit{N/A}    & \textit{N/A}   & $4.44 \times 10^{-4}$  & $0.00717$               \\
Len\(_{ATE}\)   & $7.35 \times 10^{-82}$  & \textit{N/A}    & \textit{N/A}   & $4.23 \times 10^{-4}$  & $0.00118$               \\
RMSE\(_{CATE}\) & $1.12 \times 10^{-185}$ & \textit{N/A}    & \textit{N/A}   & $8.62 \times 10^{-7}$  & $1.21 \times 10^{-5}$   \\
MAE\(_{CATE}\)  & $4.05 \times 10^{-191}$ & \textit{N/A}    & \textit{N/A}   & $8.01 \times 10^{-7}$  & $1.29 \times 10^{-5}$   \\
MAPE\(_{CATE}\) & $2.73 \times 10^{-43}$  & \textit{N/A}    & \textit{N/A}   & $0.0138$               & $0.1046$                \\
Len\(_{CATE}\)  & $0.0154$               & \textit{N/A}    & \textit{N/A}   & $0.00295$              & $0.0236$                \\
SEC\(_{ATE}\)   & $0.00839$              & \textit{N/A}    & \textit{N/A}   & $3.87 \times 10^{-6}$  & $1.06 \times 10^{-4}$   \\
AEC\(_{ATE}\)   & $0.00839$              & \textit{N/A}    & \textit{N/A}   & $9.78 \times 10^{-7}$  & $2.49 \times 10^{-6}$   \\
SEC\(_{CATE}\)  & $1.08 \times 10^{-65}$  & \textit{N/A}    & \textit{N/A}   & $1.04 \times 10^{-15}$ & $1.98 \times 10^{-12}$  \\
AEC\(_{CATE}\)  & $1.08 \times 10^{-65}$  & \textit{N/A}    & \textit{N/A}   & $1.47 \times 10^{-22}$ & $3.30 \times 10^{-22}$  \\
\hline
\end{tabular}
\end{table}
\end{landscape}

\begin{landscape}
\begin{table}[h!]
\centering
\caption{Statistical Test Results: p-values for Different Metrics (n=500)}\label{Table28}
\begin{tabular}{lccccc}
\hline
\textbf{Metric} & \textbf{Fligner-Policello Test} & \textbf{Mann-Whitney U Test} & \textbf{Kruskal-Wallis H Test} & \textbf{Levene's Test} & \textbf{Brown-Forsythe Test} \\
\hline
RMSE\(_{ATE}\)  & $0$                    & \textit{N/A}    & \textit{N/A}   & $1.01 \times 10^{-7}$  & $8.87 \times 10^{-7}$   \\
MAE\(_{ATE}\)   & $0$                    & \textit{N/A}    & \textit{N/A}   & $3.38 \times 10^{-5}$  & $2.52 \times 10^{-5}$   \\
MAPE\(_{ATE}\)  & $5.76 \times 10^{-137}$ & \textit{N/A}    & \textit{N/A}   & $3.15 \times 10^{-7}$  & $2.38 \times 10^{-5}$   \\
Len\(_{ATE}\)   & $0$                    & \textit{N/A}    & \textit{N/A}   & $1.35 \times 10^{-5}$  & $9.55 \times 10^{-5}$   \\
RMSE\(_{CATE}\) & $0$                    & \textit{N/A}    & \textit{N/A}   & $3.34 \times 10^{-14}$ & $1.99 \times 10^{-10}$  \\
MAE\(_{CATE}\)  & $0$                    & \textit{N/A}    & \textit{N/A}   & $2.49 \times 10^{-14}$ & $1.48 \times 10^{-10}$  \\
MAPE\(_{CATE}\) & $5.89 \times 10^{-172}$ & \textit{N/A}    & \textit{N/A}   & $2.12 \times 10^{-6}$  & $1.41 \times 10^{-4}$   \\
Len\(_{CATE}\)  & $0.00255$              & \textit{N/A}    & \textit{N/A}   & $1.30 \times 10^{-12}$ & $9.37 \times 10^{-8}$   \\
SEC\(_{ATE}\)   & $0$                    & \textit{N/A}    & \textit{N/A}   & $1.53 \times 10^{-7}$  & $2.00 \times 10^{-7}$   \\
AEC\(_{ATE}\)\textbf{*}   & \textit{N/A}           & \textit{N/A}    & \textit{N/A}   & \textit{N/A}               & \textit{N/A}            \\
SEC\(_{CATE}\)  & $0$                    & \textit{N/A}    & \textit{N/A}   & $2.37 \times 10^{-18}$ & $2.48 \times 10^{-12}$  \\
AEC\(_{CATE}\)  & $0$                    & \textit{N/A}    & \textit{N/A}   & $5.77 \times 10^{-27}$ & $8.28 \times 10^{-24}$  \\
\hline
\end{tabular}
\end{table}
\textbf{*:} AEC\(_{ATE}\) had p-values of $1.45 \times 10^{-43}$ for the T-test and $0.55$ for the F-test
\end{landscape}

\subsubsection{Conclusion}

In conclusion, it can be affirmed that the ps-BART model is clearly superior to the BCF model for considerably nonlinear DGPs in both point-wise and uncertainty estimation of the ATE and CATE functions. Additionally, for this set of DGPs, the proposed model is more robust than the benchmark model, contradicting the results (and respective conclusion) of the first set of DGPs. Such contradiction could be possibly explained by proposing that for significantly nonlinear DGPs the ps-BART model becomes more robust than the BCF model due to the considerably misspecification of the latter while for slightly nonlinear/considerably linear DGPs, the benchmark model is more robust. Yet, to confirm or disprove this possible explanation, the results of the 3rd set of DGPs are needed.

\subsection{3rd Set of DGPs}

\subsubsection{Linear Relationship}

Table \ref{Table29} presents the results of all metrics for the 3rd Set of DGPs with Linear Relationship. Considering the point-wise estimation performance metrics, both models have a similar performance and robustness for ATE function estimation while the benchmark model being superior and more robust for CATE function estimation. Moving to the uncertainty estimation performance measures, it can be seen that the ps-BART model is considerably better in ATE function estimation and slightly better in CATE function estimation, albeit also having a considerably poor performance for CATE function estimation. 

\begin{figure}[H]
\centering
\begin{subfigure}{0.45\textwidth}
    \centering
     \includegraphics[scale=0.2]{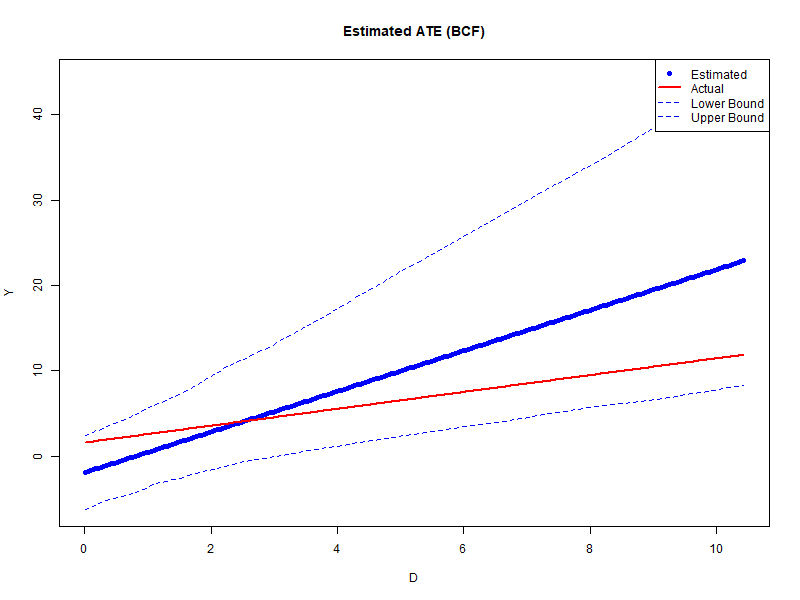}
\end{subfigure}
\hfill
\begin{subfigure}{0.45\textwidth}
    \centering
     \includegraphics[scale=0.2]{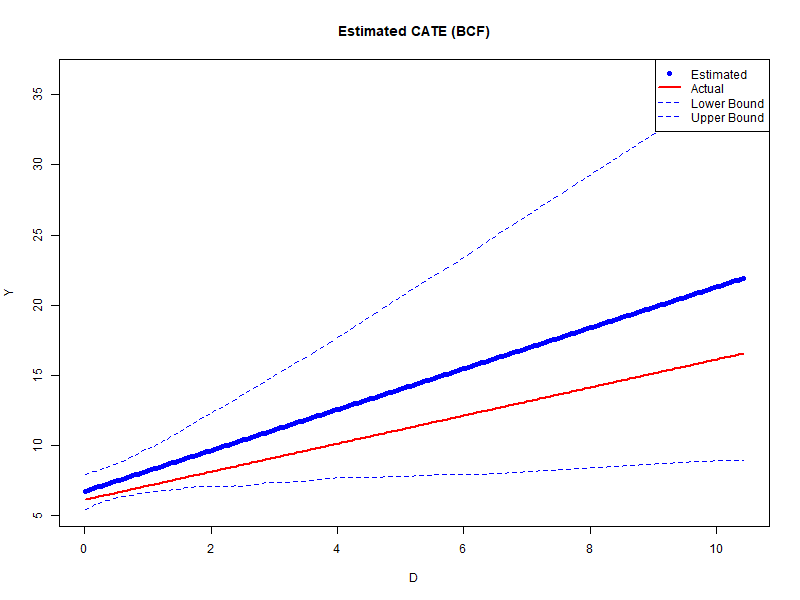}
\end{subfigure}
\caption{BCF ATE and CATE Functions Estimation for N=100 (for CATE, an Example of a Random $\mathbf{x_i}$ of a Random Simulation is used)}\label{Figure43}
\end{figure}

\begin{figure}[H]
\centering
\begin{subfigure}{0.45\textwidth}
    \centering
    \includegraphics[scale=0.2]{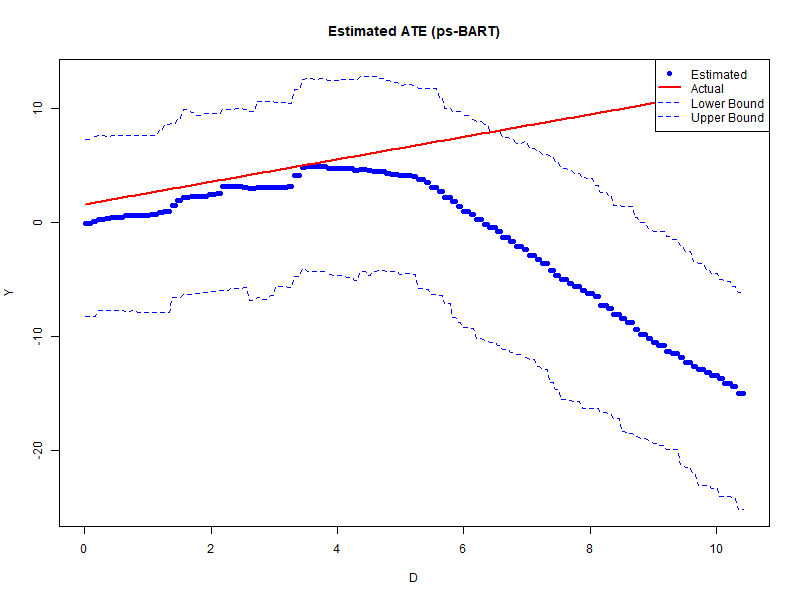}
\end{subfigure}
\hfill
\begin{subfigure}{0.45\textwidth}
    \centering
     \includegraphics[scale=0.2]{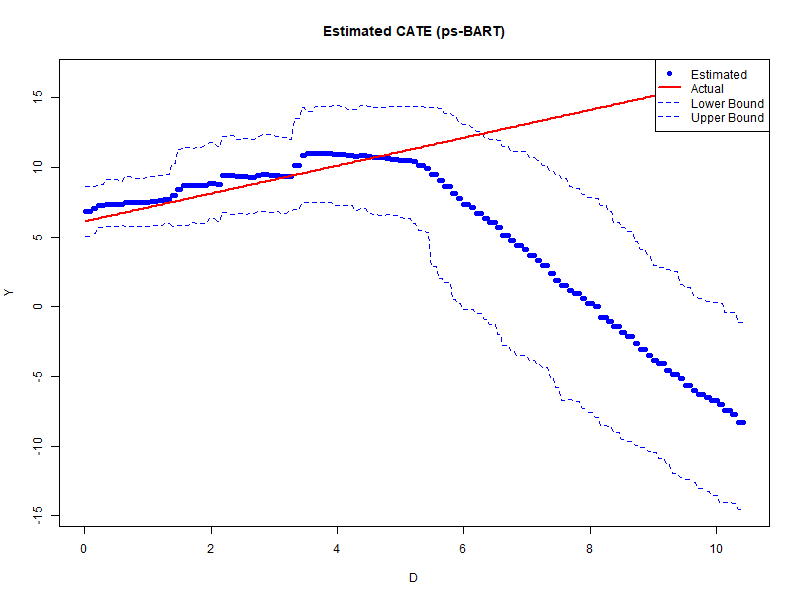}
\end{subfigure}
\caption{ps-BART ATE and CATE Functions Estimation for N=100 (for CATE, an Example of a Random $\mathbf{x_i}$ of a Random Simulation is used)}\label{Figure44}
\end{figure}

\begin{figure}[H]
\centering
\begin{subfigure}{0.45\textwidth}
    \centering
    \includegraphics[scale=0.2]{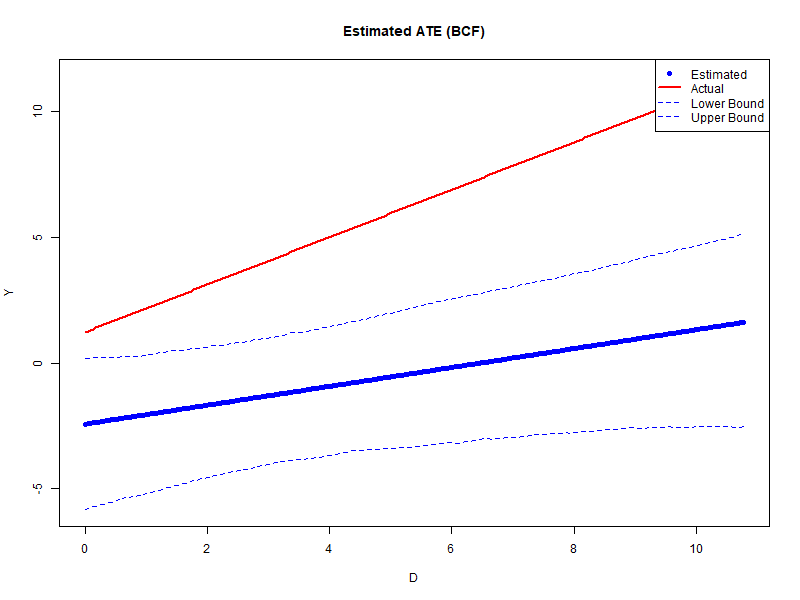}
\end{subfigure}
\hfill
\begin{subfigure}{0.45\textwidth}
    \centering
    \includegraphics[scale=0.2]{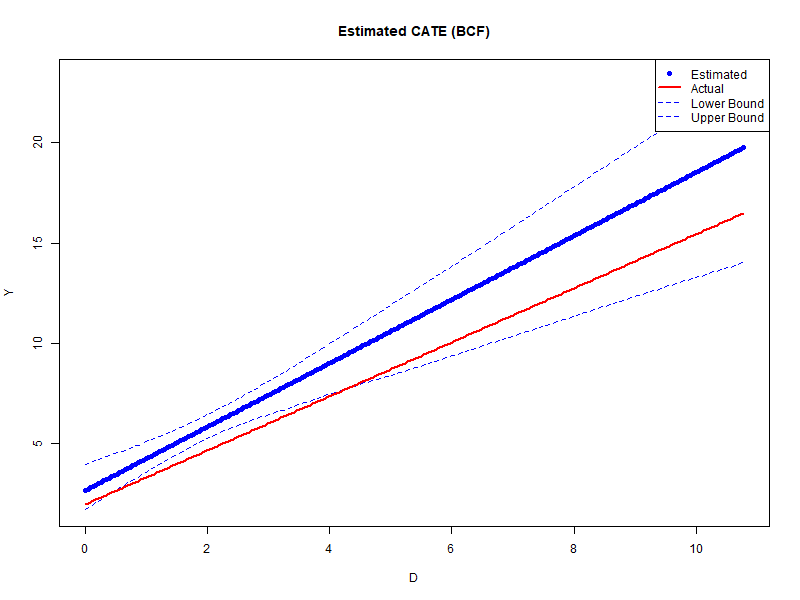}
\end{subfigure}
\caption{BCF ATE and CATE Functions Estimation for N=250 (for CATE, an Example of a Random $\mathbf{x_i}$ of a Random Simulation is used)}\label{Figure45}
\end{figure}

\begin{figure}[H]
\centering
\begin{subfigure}{0.45\textwidth}
    \centering
    \includegraphics[scale=0.2]{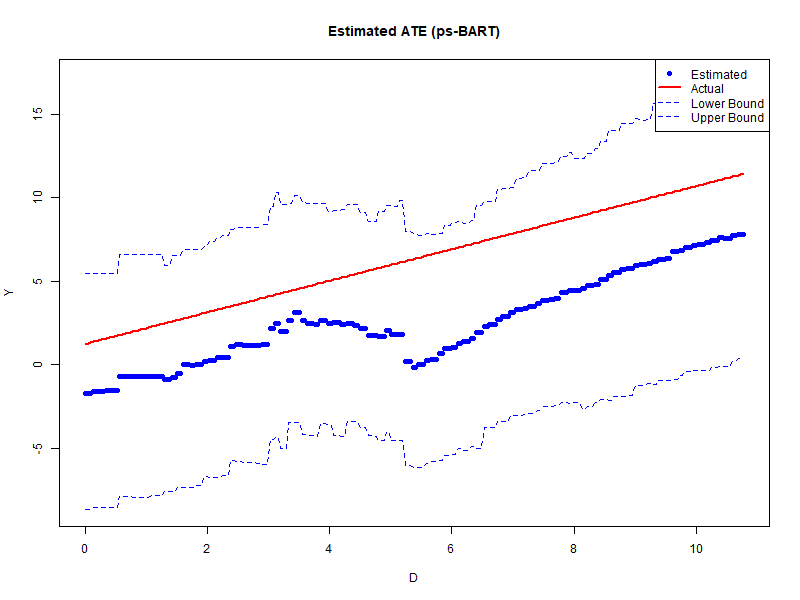}
\end{subfigure}
\hfill
\begin{subfigure}{0.45\textwidth}
    \centering
    \includegraphics[scale=0.2]{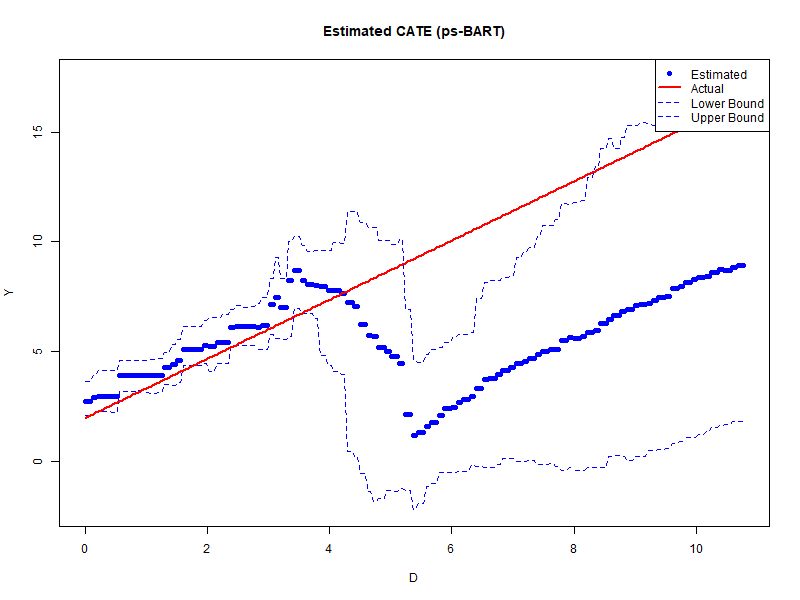}
\end{subfigure}
\caption{ps-BART ATE and CATE Functions Estimation for N=250 (for CATE, an Example of a Random $\mathbf{x_i}$ of a Random Simulation is used)}\label{Figure46}
\end{figure}

\begin{figure}[H]
\centering
\begin{subfigure}{0.45\textwidth}
    \centering
    \includegraphics[scale=0.2]{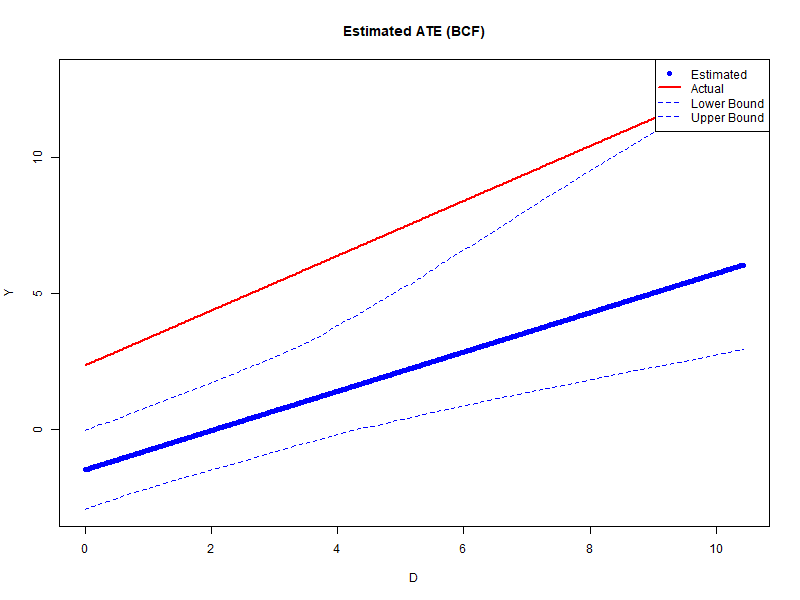}
\end{subfigure}
\hfill
\begin{subfigure}{0.45\textwidth}
    \centering
    \includegraphics[scale=0.2]{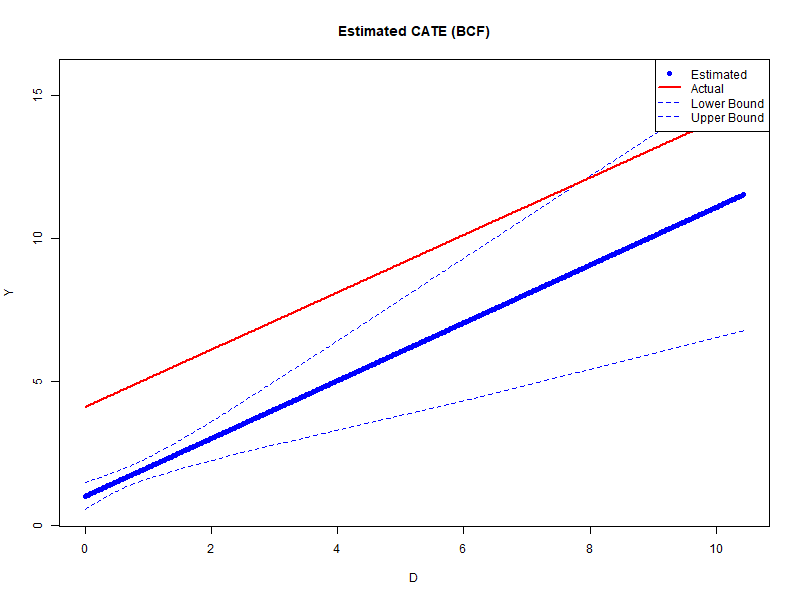}
\end{subfigure}
\caption{BCF ATE and CATE Functions Estimation for N=500 (for CATE, an Example of a Random $\mathbf{x_i}$ of a Random Simulation is used)}\label{Figure47}
\end{figure}

\begin{figure}[H]
\centering
\begin{subfigure}{0.45\textwidth}
    \centering
    \includegraphics[scale=0.2]{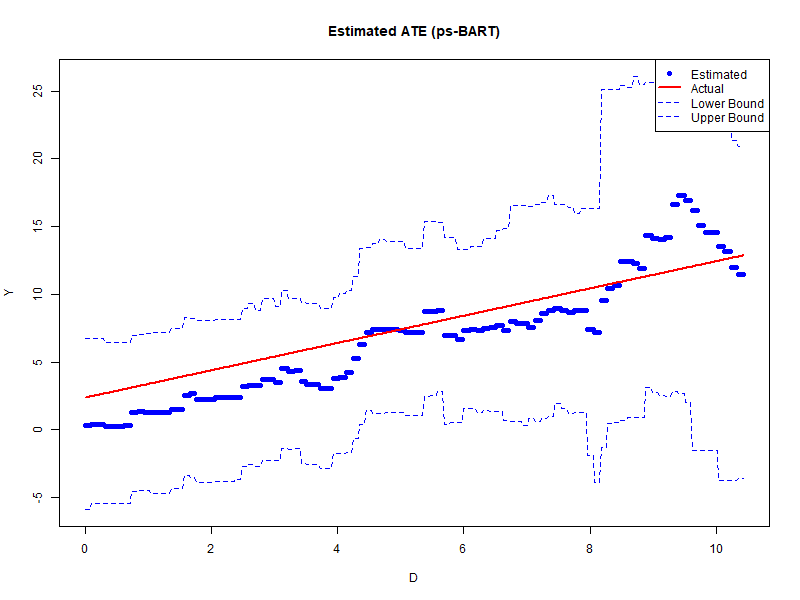}
\end{subfigure}
\hfill
\begin{subfigure}{0.45\textwidth}
    \centering
    \includegraphics[scale=0.2]{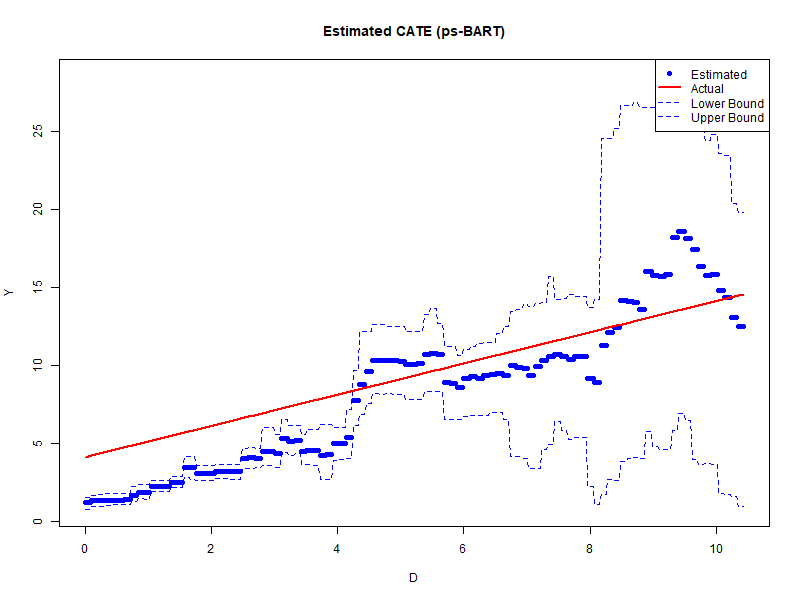}
\end{subfigure}
\caption{ps-BART ATE and CATE Functions Estimation for N=500 (for CATE, an Example of a Random $\mathbf{x_i}$ of a Random Simulation is used)}\label{Figure48}
\end{figure}

Figures \ref{Figure43}, \ref{Figure44}, \ref{Figure45}, \ref{Figure46}, \ref{Figure47}, and \ref{Figure48} graphically illustrate the results of Table \ref{Table29}. The correct specification of the BCF model for the underlying DGP, which leads to its advantage over the proposed nonparametric model, can be visually seen in these figures.

The statistical tests results for $n=100$, $n=250$, and $n=500$ can be seen in Tables \ref{Table30}, \ref{Table31}, and \ref{Table32} respectively. For $n=100$, it can be concluded (based on the results) that the similarity of both models in point-wise estimation performance for ATE function estimation is statistically significant while the superiority and greater robustness of the benchmark model is statistically significant for CATE function estimation. Moving to uncertainty estimation, it can be seen that the ps-BART model is statistically significantly better than the BCF model for ATE function estimation, while for CATE function estimation, it cannot be said that the two models possess a different performance. 

Moving to $n=250$ and $n=500$, the aforementioned conclusions remain true, apart from the affirmation that it cannot be said that the two models possess a different uncertainty estimation performance for CATE function estimation as now the ps-BART model is statistically significantly superior (though its performance is considerably poor).

\begin{table}[H]
\centering
\caption{Metric Results}\label{Table29}
\begin{tabular}{llcc}
\hline
$n$ & \textbf{Metric} & \textbf{BCF (Mean $\pm$ SD)} & \textbf{ps-BART (Mean $\pm$ SD)} \\
\hline
100 & RMSE\(_{ATE}\) & $2.838 \pm 0.878$ & $3.061 \pm 0.868$ \\
& MAE\(_{ATE}\)  & $2.751 \pm 0.873$ & $2.785 \pm 0.800$ \\
& MAPE\(_{ATE}\) & $1.050 \pm 0.386$ & $1.045 \pm 0.409$ \\
& Len\(_{ATE}\) & $8.550 \pm 1.529$ & $13.272 \pm 4.629$ \\
& Cover\(_{ATE}\) & $0.847 \pm 0.307$ & $0.989 \pm 0.025$ \\
& RMSE\(_{CATE}\) & $4.582 \pm 0.827$ & $6.785 \pm 3.619$ \\
& MAE\(_{CATE}\)  & $4.339 \pm 0.720$ & $5.843 \pm 2.629$ \\
& MAPE\(_{CATE}\) & $4.397 \pm 3.362$ & $4.707 \pm 3.394$ \\
& Len\(_{CATE}\) & $8.022 \pm 3.551$ & $8.496 \pm 3.783$ \\
& Cover\(_{CATE}\) & $0.468 \pm 0.102$ & $0.442 \pm 0.095$ \\
& SEC\(_{ATE}\)  & $0.104 \pm 0.238$ & $0.002 \pm 0.001$ \\
& AEC\(_{ATE}\)  & $0.179 \pm 0.269$ & $0.044 \pm 0.014$ \\
& SEC\(_{CATE}\)  & $0.243 \pm 0.097$ & $0.267 \pm 0.097$ \\
& AEC\(_{CATE}\)  & $0.482 \pm 0.102$ & $0.508 \pm 0.095$ \\
\hline
250 & RMSE\(_{ATE}\) & $3.049 \pm 0.787$ & $3.136 \pm 0.908$ \\
& MAE\(_{ATE}\)  & $2.953 \pm 0.765$ & $2.973 \pm 0.891$ \\
& MAPE\(_{ATE}\) & $1.080 \pm 0.273$ & $1.088 \pm 0.390$ \\
& Len\(_{ATE}\) & $6.004 \pm 1.319$ & $13.730 \pm 3.432$ \\
& Cover\(_{ATE}\) & $0.444 \pm 0.397$ & $0.994 \pm 0.015$ \\
& RMSE\(_{CATE}\) & $4.263 \pm 0.509$ & $7.146 \pm 3.565$ \\
& MAE\(_{CATE}\)  & $4.104 \pm 0.433$ & $5.995 \pm 2.400$ \\
& MAPE\(_{CATE}\) & $3.576 \pm 2.335$ & $4.130 \pm 2.591$ \\
& Len\(_{CATE}\) & $4.618 \pm 1.783$ & $8.033 \pm 3.589$ \\
& Cover\(_{CATE}\) & $0.285 \pm 0.076$ & $0.375 \pm 0.066$ \\
& SEC\(_{ATE}\)  & $0.412 \pm 0.364$ & $0.002 \pm 0.001$ \\
& AEC\(_{ATE}\)  & $0.530 \pm 0.365$ & $0.046 \pm 0.007$ \\
& SEC\(_{CATE}\)  & $0.448 \pm 0.098$ & $0.335 \pm 0.077$ \\
& AEC\(_{CATE}\)  & $0.665 \pm 0.076$ & $0.575 \pm 0.066$ \\
\hline
500 & RMSE\(_{ATE}\) & $3.169 \pm 1.142$ & $3.126 \pm 0.923$ \\
& MAE\(_{ATE}\)  & $3.051 \pm 1.093$ & $3.017 \pm 0.941$ \\
& MAPE\(_{ATE}\) & $1.082 \pm 0.360$ & $1.070 \pm 0.333$ \\
& Len\(_{ATE}\) & $3.995 \pm 0.823$ & $13.270 \pm 3.759$ \\
& Cover\(_{ATE}\) & $0.173 \pm 0.281$ & $0.983 \pm 0.105$ \\
& RMSE\(_{CATE}\) & $4.015 \pm 0.228$ & $7.304 \pm 2.547$ \\
& MAE\(_{CATE}\)  & $3.914 \pm 0.195$ & $6.067 \pm 1.709$ \\
& MAPE\(_{CATE}\) & $3.335 \pm 2.015$ & $3.697 \pm 2.062$ \\
& Len\(_{CATE}\) & $3.022 \pm 0.869$ & $7.507 \pm 2.988$ \\
& Cover\(_{CATE}\) & $0.200 \pm 0.054$ & $0.319 \pm 0.074$ \\
& SEC\(_{ATE}\)  & $0.683 \pm 0.302$ & $0.012 \pm 0.085$ \\
& AEC\(_{ATE}\)  & $0.784 \pm 0.264$ & $0.058 \pm 0.093$ \\
& SEC\(_{CATE}\)  & $0.566 \pm 0.081$ & $0.403 \pm 0.091$ \\
& AEC\(_{CATE}\)  & $0.750 \pm 0.054$ & $0.631 \pm 0.074$ \\
\hline
\end{tabular}
\end{table}

\begin{landscape}
\begin{table}[h!]
\centering
\caption{Statistical Test Results: p-values for Different Metrics (n=100)}\label{Table30}
\begin{tabular}{lccccc}
\hline
\textbf{Metric} & \textbf{Fligner-Policello Test} & \textbf{Mann-Whitney U Test} & \textbf{Kruskal-Wallis H Test} & \textbf{Levene's Test} & \textbf{Brown-Forsythe Test} \\
\hline
RMSE\(_{ATE}\)\textbf{*}  & \textit{N/A}           & \textit{N/A}    & \textit{N/A}   & \textit{N/A}            & \textit{N/A}            \\
MAE\(_{ATE}\)\textbf{*}   & \textit{N/A}           & \textit{N/A}    & \textit{N/A}   & \textit{N/A}              & \textit{N/A}            \\
MAPE\(_{ATE}\)  & \textit{N/A}           & $0.9231$        & $0.9221$       & $0.5095$              & $0.6075$                \\
Len\(_{ATE}\)   & $1.40 \times 10^{-72}$ & \textit{N/A}    & \textit{N/A}   & $9.25 \times 10^{-9}$ & $1.54 \times 10^{-6}$   \\
RMSE\(_{CATE}\) & $4.98 \times 10^{-28}$ & \textit{N/A}    & \textit{N/A}   & $1.06 \times 10^{-7}$ & $3.46 \times 10^{-5}$   \\
MAE\(_{CATE}\)  & $1.30 \times 10^{-17}$ & \textit{N/A}    & \textit{N/A}   & $3.51 \times 10^{-7}$ & $5.34 \times 10^{-5}$   \\
MAPE\(_{CATE}\) & \textit{N/A}           & $0.0959$        & $0.0956$       & $0.9890$              & $0.9558$                \\
Len\(_{CATE}\)  & \textit{N/A}           & $0.3375$        & $0.3369$       & $0.7350$              & $0.8401$                \\
SEC\(_{ATE}\)   & $1.86 \times 10^{-20}$ & \textit{N/A}    & \textit{N/A}   & $6.53 \times 10^{-17}$ & $3.54 \times 10^{-5}$   \\
AEC\(_{ATE}\)   & $1.86 \times 10^{-20}$ & \textit{N/A}    & \textit{N/A}   & $2.72 \times 10^{-21}$ & $1.06 \times 10^{-5}$   \\
SEC\(_{CATE}\)  & \textit{N/A}           & $0.1019$        & $0.1016$       & $0.5731$              & $0.5961$                \\
AEC\(_{CATE}\)\textbf{*}  & \textit{N/A}           & \textit{N/A}    & \textit{N/A}   & $0.4287$              & \textit{N/A}            \\
\hline
\end{tabular}
\end{table}
\textbf{*:} RMSE\(_{ATE}\), MAE\(_{ATE}\), and AEC\(_{CATE}\) had p-values of $0.07$, $0.77$, and $0.06$ for the T-test and $0.91$, $0.39$, and $0.43$ for the F-test respectively.
\end{landscape}

\begin{landscape}
\begin{table}[h!]
\centering
\caption{Statistical Test Results: p-values for Different Metrics (n=250)}\label{Table31}
\begin{tabular}{lccccc}
\hline
\textbf{Metric} & \textbf{Fligner-Policello Test} & \textbf{Mann-Whitney U Test} & \textbf{Kruskal-Wallis H Test} & \textbf{Levene's Test} & \textbf{Brown-Forsythe Test} \\
\hline
RMSE\(_{ATE}\)  & \textit{N/A}            & $0.4165$        & $0.4158$        & $0.2550$               & $0.2549$                \\
MAE\(_{ATE}\)\textbf{*}   & \textit{N/A}            & \textit{N/A}    & \textit{N/A}    & \textit{N/A}              & \textit{N/A}            \\
MAPE\(_{ATE}\)  & $0.9001$                & \textit{N/A}    & \textit{N/A}    & $0.0101$               & $0.0110$                \\
Len\(_{ATE}\)   & $0$                     & \textit{N/A}    & \textit{N/A}    & $1.75 \times 10^{-12}$ & $6.51 \times 10^{-12}$  \\
RMSE\(_{CATE}\) & $1.83 \times 10^{-185}$ & \textit{N/A}    & \textit{N/A}    & $1.40 \times 10^{-10}$ & $3.77 \times 10^{-7}$   \\
MAE\(_{CATE}\)  & $6.99 \times 10^{-65}$  & \textit{N/A}    & \textit{N/A}    & $3.00 \times 10^{-11}$ & $1.54 \times 10^{-7}$   \\
MAPE\(_{CATE}\) & \textit{N/A}            & $0.0035$        & $0.0035$        & $0.3926$               & $0.4815$                \\
Len\(_{CATE}\)  & $5.15 \times 10^{-47}$  & \textit{N/A}    & \textit{N/A}    & $5.01 \times 10^{-5}$  & $5.82 \times 10^{-4}$   \\
SEC\(_{ATE}\)   & $5.21 \times 10^{-96}$  & \textit{N/A}    & \textit{N/A}    & $5.26 \times 10^{-57}$ & $2.07 \times 10^{-56}$  \\
AEC\(_{ATE}\)   & $5.21 \times 10^{-96}$  & \textit{N/A}    & \textit{N/A}    & $1.32 \times 10^{-56}$ & $3.23 \times 10^{-41}$  \\
SEC\(_{CATE}\)  & \textit{N/A}            & $6.60 \times 10^{-15}$ & $6.53 \times 10^{-15}$  & $0.0703$               & $0.1193$               \\
AEC\(_{CATE}\)  & \textit{N/A}            & $6.60 \times 10^{-15}$ & $6.53 \times 10^{-15}$  & $0.4335$               & $0.6741$               \\
\hline
\end{tabular}
\end{table}
\textbf{*:} MAE\(_{ATE}\) had p-values of $0.86$ for the T-test and $0.13$ for the F-test.
\end{landscape}

\begin{landscape}
\begin{table}[h!]
\centering
\caption{Statistical Test Results: p-values for Different Metrics (n=500)}\label{Table32}
\begin{tabular}{lccccc}
\hline
\textbf{Metric} & \textbf{Fligner-Policello Test} & \textbf{Mann-Whitney U Test} & \textbf{Kruskal-Wallis H Test} & \textbf{Levene's Test} & \textbf{Brown-Forsythe Test} \\
\hline
RMSE\(_{ATE}\)  & $0.8134$                & \textit{N/A}    & \textit{N/A}    & $0.0194$               & $0.0493$                \\
MAE\(_{ATE}\)   & $0.9750$                & \textit{N/A}    & \textit{N/A}    & $0.0364$               & $0.0660$                \\
MAPE\(_{ATE}\)  & \textit{N/A}            & $0.9212$        & $0.9202$        & $0.3540$               & $0.3766$                \\
Len\(_{ATE}\)   & $0$                     & \textit{N/A}    & \textit{N/A}    & $4.48 \times 10^{-13}$ & $9.27 \times 10^{-11}$  \\
RMSE\(_{CATE}\) & $0$                     & \textit{N/A}    & \textit{N/A}    & $1.89 \times 10^{-18}$ & $5.58 \times 10^{-13}$  \\
MAE\(_{CATE}\)  & $0$                     & \textit{N/A}    & \textit{N/A}    & $7.70 \times 10^{-19}$ & $2.07 \times 10^{-14}$  \\
MAPE\(_{CATE}\) & \textit{N/A}            & $0.0023$        & $0.0023$        & $0.8456$               & $0.8436$                \\
Len\(_{CATE}\)  & $0$                     & \textit{N/A}    & \textit{N/A}    & $2.65 \times 10^{-14}$ & $5.07 \times 10^{-11}$  \\
SEC\(_{ATE}\)   & $0$                     & \textit{N/A}    & \textit{N/A}    & $3.03 \times 10^{-29}$ & $7.23 \times 10^{-11}$  \\
AEC\(_{ATE}\)   & $0$                     & \textit{N/A}    & \textit{N/A}    & $2.34 \times 10^{-17}$ & $1.26 \times 10^{-7}$   \\
SEC\(_{CATE}\)\textbf{*}  & \textit{N/A}            & \textit{N/A}    & \textit{N/A}    & \textit{N/A}              & \textit{N/A}            \\
AEC\(_{CATE}\)  & $8.99 \times 10^{-94}$  & \textit{N/A}    & \textit{N/A}    & $0.0036$               & $0.0114$                \\
\hline
\end{tabular}
\end{table}
\textbf{*:} SEC\(_{CATE}\) had p-values of $2.68 \times 10^{-29}$ for the T-test and $0.24$ for the F-test.
\end{landscape}

\subsubsection{Nonlinear Relationship}

Table \ref{Table33} shows the performance measures results for the 3rd Set of DGPs with Nonlinear Relationship. For $n=100$, it can be seen that both models have a similar performance and robustness for ATE function estimation regarding point-wise estimation, while the ps-BART model being slightly better and more robust for CATE function estimation. Table \ref{Table34} shows that the similarity between the models is statistically significant for ATE function estimation while the slight superiority of the ps-BART model is not for CATE function estimation. Concerning uncertainty estimation, both models have a similar performance and robustness considering ATE function estimation (which is also confirmed by the results found in Table \ref{Table34}), while the benchmark model being statistically significantly better for CATE function estimation. For $n=250$ and $n=500$, both models still have a similar performance for ATE function estimation regarding point-wise estimation (which is confirmed by the results of Tables \ref{Table35} and \ref{Table36}), but now the proposed model is statistically significantly more robust. Regarding point-wise CATE function estimation, the ps-BART model's advantage over the BCF model is now more pronounced, making it statistically significant. Regarding uncertainty estimation, the proposed model has a better performance and robustness considering ATE function estimation (though only statistically significant for $n=250$), while the benchmark model being statistically significantly better for CATE function estimation.

\begin{figure}[H]
\centering
\begin{subfigure}{0.45\textwidth}
    \centering
     \includegraphics[scale=0.2]{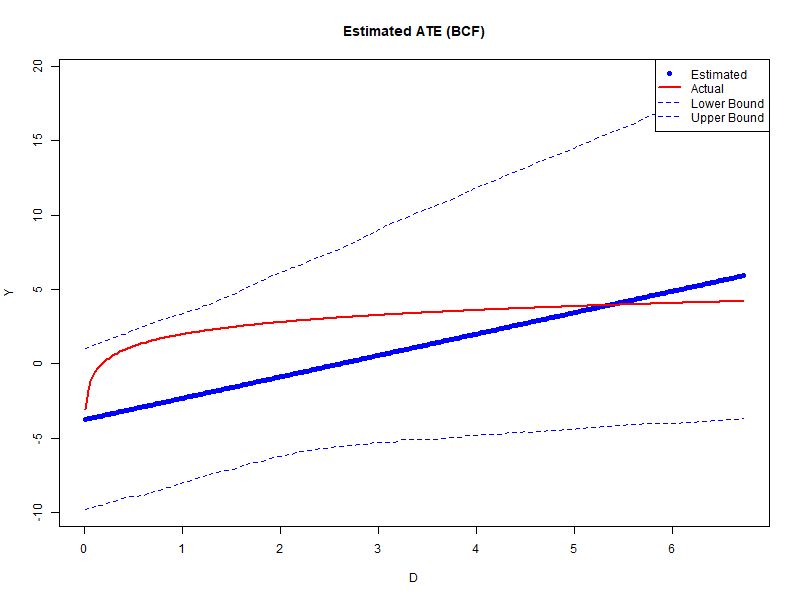}
\end{subfigure}
\hfill
\begin{subfigure}{0.45\textwidth}
    \centering
     \includegraphics[scale=0.2]{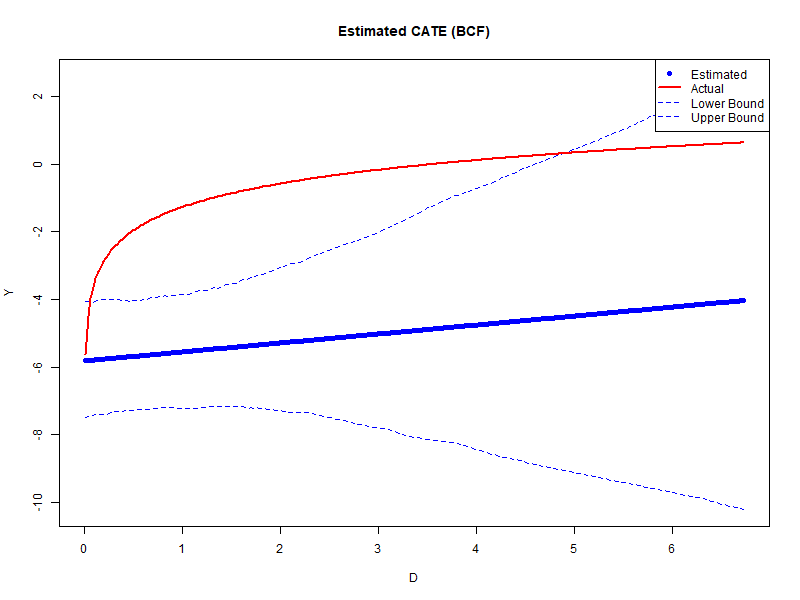}
\end{subfigure}
\caption{BCF ATE and CATE Functions Estimation for N=100 (for CATE, an Example of a Random $\mathbf{x_i}$ of a Random Simulation is used)}\label{Figure49}
\end{figure}

\begin{figure}[H]
\centering
\begin{subfigure}{0.45\textwidth}
    \centering
    \includegraphics[scale=0.2]{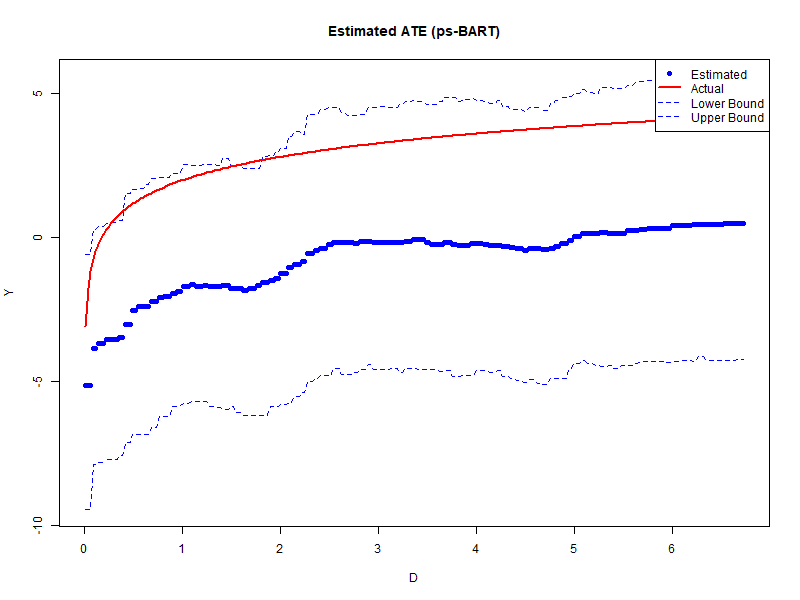}
\end{subfigure}
\hfill
\begin{subfigure}{0.45\textwidth}
    \centering
     \includegraphics[scale=0.2]{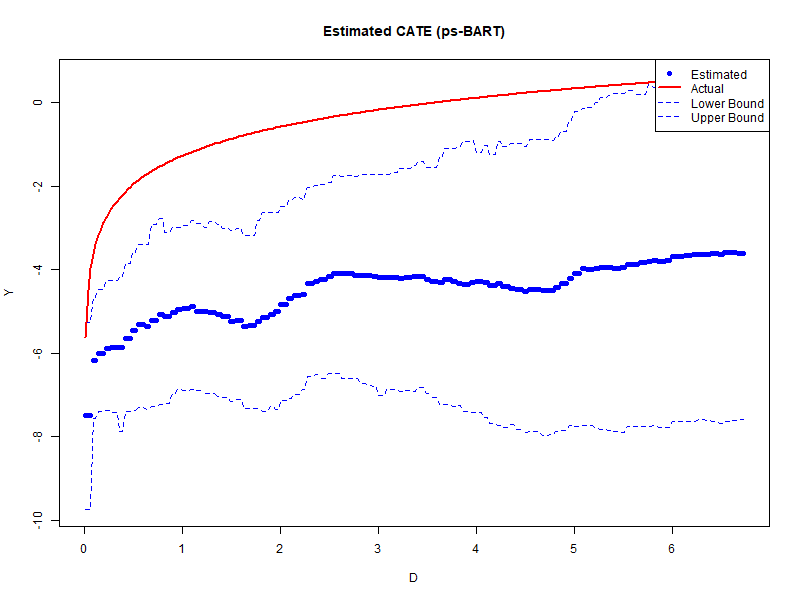}
\end{subfigure}
\caption{ps-BART ATE and CATE Functions Estimation for N=100 (for CATE, an Example of a Random $\mathbf{x_i}$ of a Random Simulation is used)}\label{Figure50}
\end{figure}

\begin{figure}[H]
\centering
\begin{subfigure}{0.45\textwidth}
    \centering
    \includegraphics[scale=0.2]{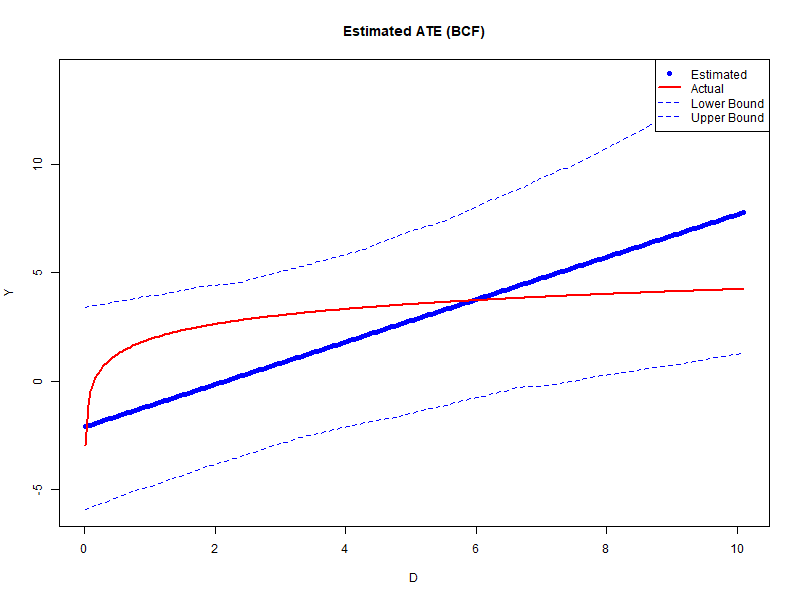}
\end{subfigure}
\hfill
\begin{subfigure}{0.45\textwidth}
    \centering
    \includegraphics[scale=0.2]{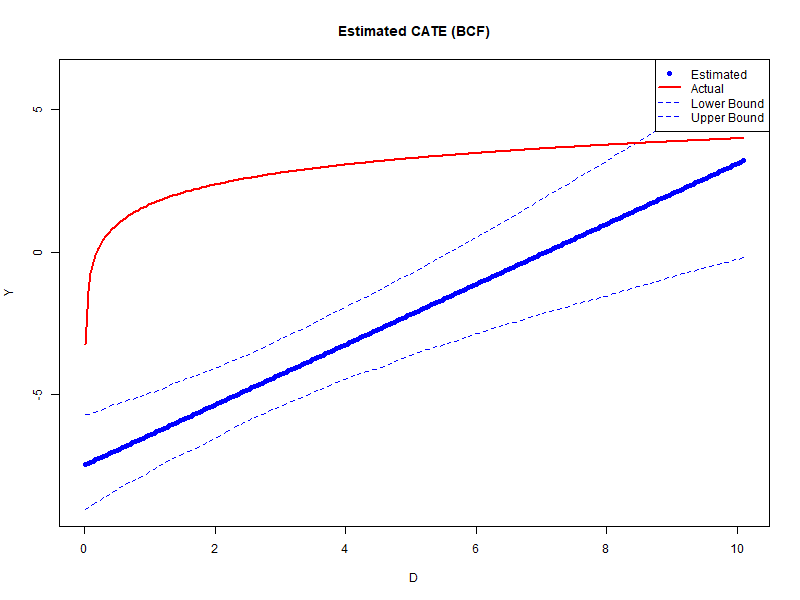}
\end{subfigure}
\caption{BCF ATE and CATE Functions Estimation for N=250 (for CATE, an Example of a Random $\mathbf{x_i}$ of a Random Simulation is used)}\label{Figure51}
\end{figure}

\begin{figure}[H]
\centering
\begin{subfigure}{0.45\textwidth}
    \centering
    \includegraphics[scale=0.2]{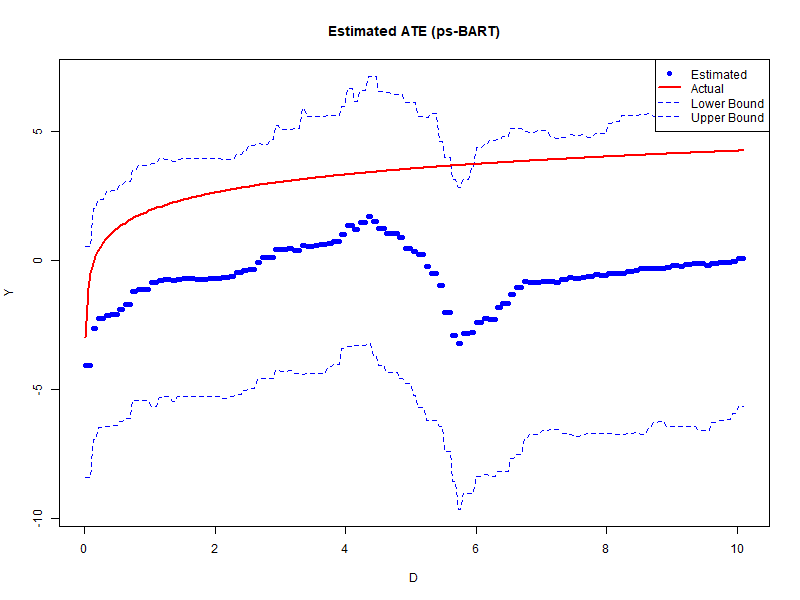}
\end{subfigure}
\hfill
\begin{subfigure}{0.45\textwidth}
    \centering
    \includegraphics[scale=0.2]{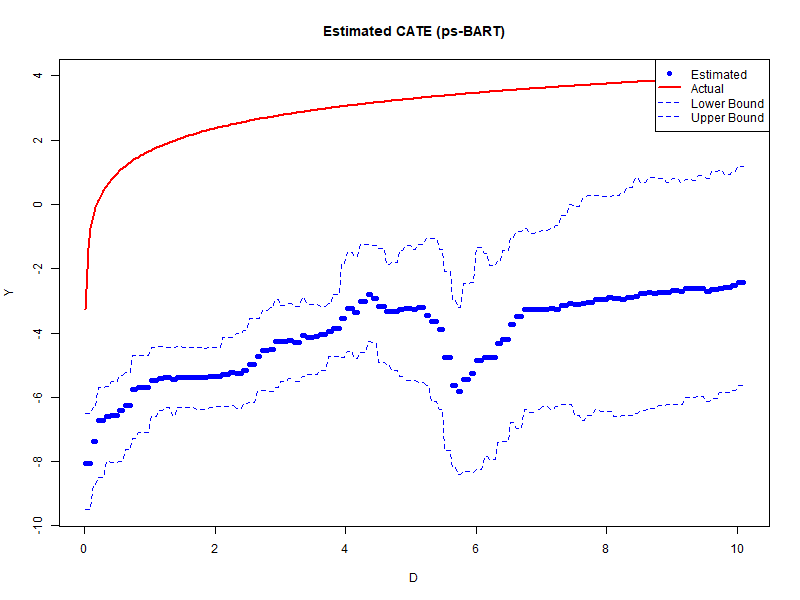}
\end{subfigure}
\caption{ps-BART ATE and CATE Functions Estimation for N=250 (for CATE, an Example of a Random $\mathbf{x_i}$ of a Random Simulation is used)}\label{Figure52}
\end{figure}

\begin{figure}[H]
\centering
\begin{subfigure}{0.45\textwidth}
    \centering
    \includegraphics[scale=0.2]{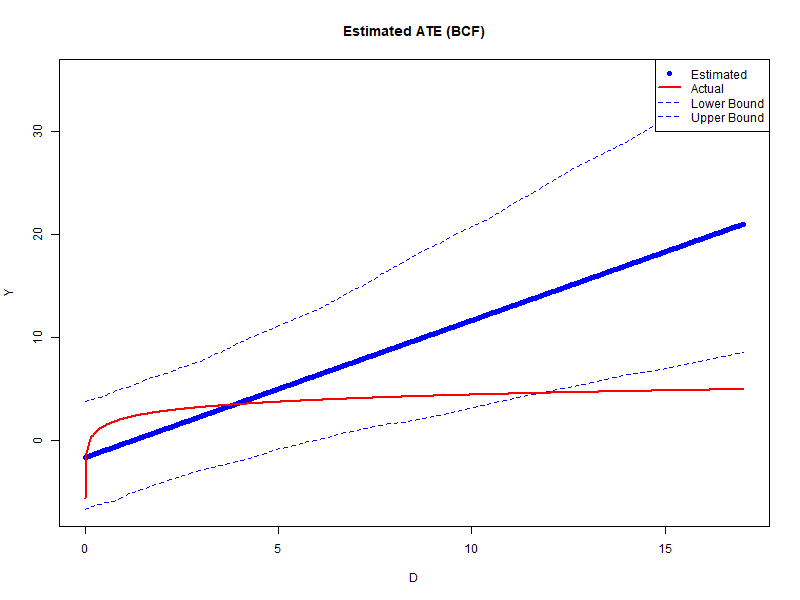}
\end{subfigure}
\hfill
\begin{subfigure}{0.45\textwidth}
    \centering
    \includegraphics[scale=0.2]{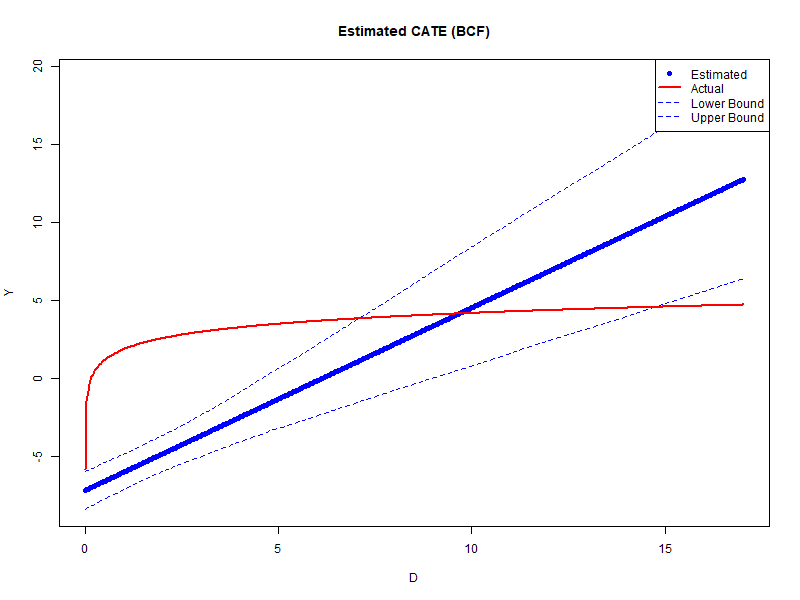}
\end{subfigure}
\caption{BCF ATE and CATE Functions Estimation for N=500 (for CATE, an Example of a Random $\mathbf{x_i}$ of a Random Simulation is used)}\label{Figure53}
\end{figure}

\begin{figure}[H]
\centering
\begin{subfigure}{0.45\textwidth}
    \centering
    \includegraphics[scale=0.2]{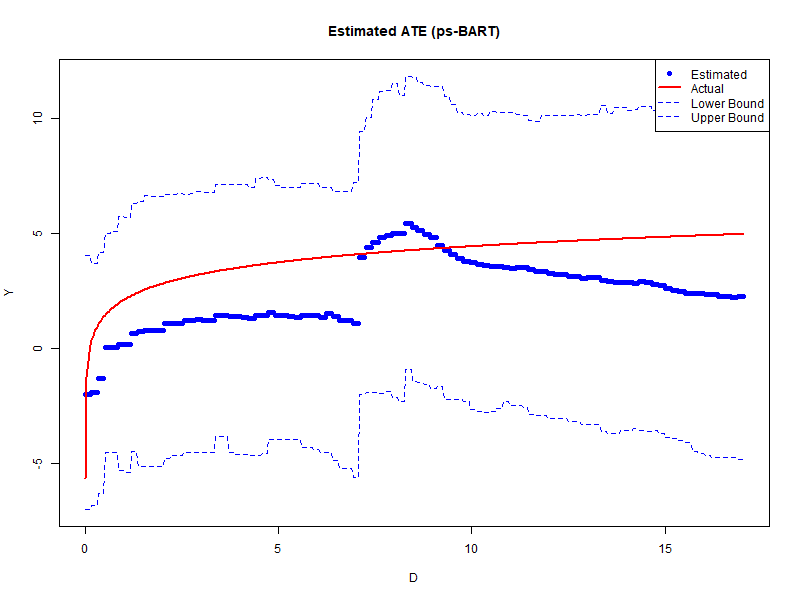}
\end{subfigure}
\hfill
\begin{subfigure}{0.45\textwidth}
    \centering
    \includegraphics[scale=0.2]{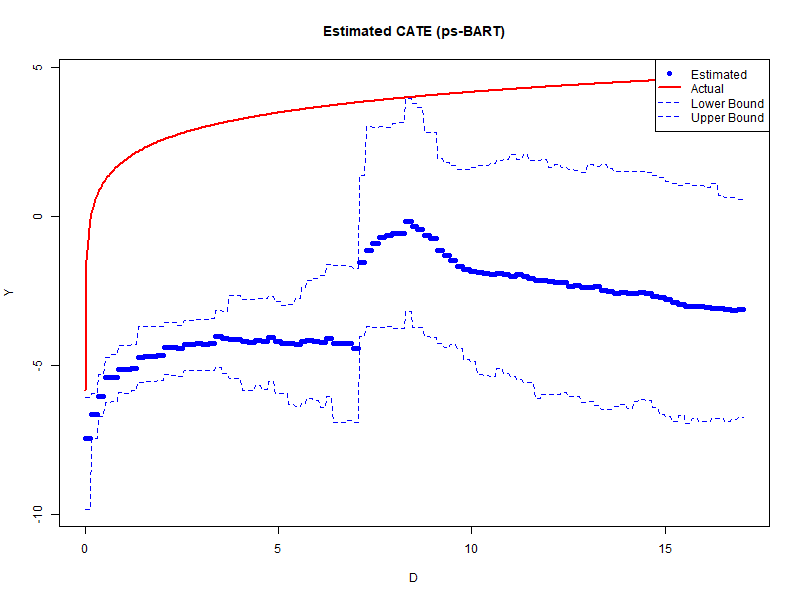}
\end{subfigure}
\caption{ps-BART ATE and CATE Functions Estimation for N=500 (for CATE, an Example of a Random $\mathbf{x_i}$ of a Random Simulation is used)}\label{Figure54}
\end{figure}

Figures \ref{Figure49}, \ref{Figure50}, \ref{Figure51}, \ref{Figure52}, \ref{Figure53}, and \ref{Figure54} graphically illustrate the results of Table \ref{Table33}. The misspecification of the BCF model for the underlying DGP can be visually seen in these figures.

\begin{table}[H]
\centering
\caption{Metric Results}\label{Table33}
\begin{tabular}{llcc}
\hline
$n$ & \textbf{Metric} & \textbf{BCF (Mean $\pm$ SD)} & \textbf{ps-BART (Mean $\pm$ SD)} \\
\hline
100 & RMSE\(_{ATE}\) & $2.914 \pm 0.859$ & $2.861 \pm 0.802$ \\
& MAE\(_{ATE}\)  & $2.753 \pm 0.883$ & $2.793 \pm 0.826$ \\
& MAPE\(_{ATE}\) & $6.165 \pm 6.970$ & $6.430 \pm 7.020$ \\
& Len\(_{ATE}\) & $10.572 \pm 1.742$ & $10.153 \pm 1.554$ \\
& Cover\(_{ATE}\) & $0.953 \pm 0.155$ & $0.958 \pm 0.162$ \\
& RMSE\(_{CATE}\) & $4.600 \pm 0.779$ & $4.337 \pm 0.467$ \\
& MAE\(_{CATE}\)  & $4.126 \pm 0.613$ & $4.127 \pm 0.447$ \\
& MAPE\(_{CATE}\) & $8.567 \pm 7.974$ & $7.892 \pm 7.129$ \\
& Len\(_{CATE}\) & $9.011 \pm 2.629$ & $6.367 \pm 1.013$ \\
& Cover\(_{CATE}\) & $0.561 \pm 0.092$ & $0.451 \pm 0.071$ \\
& SEC\(_{ATE}\)  & $0.024 \pm 0.087$ & $0.026 \pm 0.125$ \\
& AEC\(_{ATE}\)  & $0.086 \pm 0.128$ & $0.078 \pm 0.143$ \\
& SEC\(_{CATE}\)  & $0.159 \pm 0.076$ & $0.254 \pm 0.071$ \\
& AEC\(_{CATE}\)  & $0.389 \pm 0.092$ & $0.499 \pm 0.071$ \\
\hline
250 & RMSE\(_{ATE}\) & $3.044 \pm 0.778$ & $2.986 \pm 0.590$ \\
& MAE\(_{ATE}\)  & $2.859 \pm 0.806$ & $2.936 \pm 0.596$ \\
& MAPE\(_{ATE}\) & $7.492 \pm 16.323$ & $7.932 \pm 12.287$ \\
& Len\(_{ATE}\) & $9.743 \pm 1.498$ & $10.254 \pm 1.733$ \\
& Cover\(_{ATE}\) & $0.902 \pm 0.198$ & $0.972 \pm 0.109$ \\
& RMSE\(_{CATE}\) & $4.909 \pm 0.985$ & $4.135 \pm 0.328$ \\
& MAE\(_{CATE}\)  & $4.326 \pm 0.782$ & $3.959 \pm 0.310$ \\
& MAPE\(_{CATE}\) & $8.324 \pm 3.958$ & $7.346 \pm 3.948$ \\
& Len\(_{CATE}\) & $7.376 \pm 1.853$ & $4.924 \pm 0.767$ \\
& Cover\(_{CATE}\) & $0.462 \pm 0.063$ & $0.363 \pm 0.056$ \\
& SEC\(_{ATE}\)  & $0.041 \pm 0.098$ & $0.012 \pm 0.076$ \\
& AEC\(_{ATE}\)  & $0.123 \pm 0.162$ & $0.060 \pm 0.094$ \\
& SEC\(_{CATE}\)  & $0.242 \pm 0.062$ & $0.347 \pm 0.064$ \\
& AEC\(_{CATE}\)  & $0.488 \pm 0.063$ & $0.587 \pm 0.056$ \\
\hline
500 & RMSE\(_{ATE}\) & $3.132 \pm 0.935$ & $3.037 \pm 0.664$ \\
& MAE\(_{ATE}\)  & $2.884 \pm 0.995$ & $2.995 \pm 0.667$ \\
& MAPE\(_{ATE}\) & $8.715 \pm 9.356$ & $11.079 \pm 14.312$ \\
& Len\(_{ATE}\) & $9.220 \pm 1.675$ & $10.365 \pm 1.716$ \\
& Cover\(_{ATE}\) & $0.859 \pm 0.260$ & $0.982 \pm 0.060$ \\
& RMSE\(_{CATE}\) & $5.284 \pm 1.186$ & $4.060 \pm 0.454$ \\
& MAE\(_{CATE}\)  & $4.586 \pm 0.933$ & $3.904 \pm 0.380$ \\
& MAPE\(_{CATE}\) & $9.458 \pm 7.254$ & $8.290 \pm 5.972$ \\
& Len\(_{CATE}\) & $6.122 \pm 1.485$ & $4.185 \pm 0.853$ \\
& Cover\(_{CATE}\) & $0.375 \pm 0.049$ & $0.303 \pm 0.063$ \\
& SEC\(_{ATE}\)  & $0.075 \pm 0.172$ & $0.005 \pm 0.018$ \\
& AEC\(_{ATE}\)  & $0.154 \pm 0.228$ & $0.053 \pm 0.042$ \\
& SEC\(_{CATE}\)  & $0.333 \pm 0.057$ & $0.423 \pm 0.080$ \\
& AEC\(_{CATE}\)  & $0.575 \pm 0.049$ & $0.647 \pm 0.063$ \\
\hline
\end{tabular}
\end{table}

\begin{landscape}
\begin{table}[h!]
\centering
\caption{Statistical Test Results: p-values for Different Metrics (n=100)}\label{Table34}
\begin{tabular}{lccccc}
\hline
\textbf{Metric} & \textbf{Fligner-Policello Test} & \textbf{Mann-Whitney U Test} & \textbf{Kruskal-Wallis H Test} & \textbf{Levene's Test} & \textbf{Brown-Forsythe Test} \\
\hline
RMSE\(_{ATE}\)  & \textit{N/A}            & $0.9659$        & $0.9649$        & $0.8676$               & $0.9111$                \\
MAE\(_{ATE}\)   & \textit{N/A}            & $0.3958$        & $0.3952$        & $0.8342$               & $0.8786$                \\
MAPE\(_{ATE}\)  & \textit{N/A}            & $0.2721$        & $0.2715$        & $0.8261$               & $0.8958$                \\
Len\(_{ATE}\)   & \textit{N/A}            & $0.0954$        & $0.0952$        & $0.3249$               & $0.3676$                \\
RMSE\(_{CATE}\) & $0.0203$                & \textit{N/A}    & \textit{N/A}    & $0.0010$               & $0.0112$                \\
MAE\(_{CATE}\)  & $0.3959$                & \textit{N/A}    & \textit{N/A}    & $0.0458$               & $0.1266$                \\
MAPE\(_{CATE}\) & \textit{N/A}            & $0.1518$        & $0.1515$        & $0.7203$               & $0.7876$                \\
Len\(_{CATE}\)  & $1.61 \times 10^{-36}$  & \textit{N/A}    & \textit{N/A}    & $1.08 \times 10^{-9}$  & $9.84 \times 10^{-9}$   \\
SEC\(_{ATE}\)   & \textit{N/A}            & $0.1260$        & $0.1256$        & $0.6469$               & $0.8582$                \\
AEC\(_{ATE}\)   & \textit{N/A}            & $0.1260$        & $0.1256$        & $0.5490$               & $0.8137$                \\
SEC\(_{CATE}\)  & \textit{N/A}            & $6.08 \times 10^{-16}$ & $6.02 \times 10^{-16}$  & $0.8055$               & $0.9724$                \\
AEC\(_{CATE}\)  & $8.29 \times 10^{-29}$  & \textit{N/A}    & \textit{N/A}    & $0.0364$               & $0.0445$                \\
\hline
\end{tabular}
\end{table}
\end{landscape}

\begin{landscape}
\begin{table}[h!]
\centering
\caption{Statistical Test Results: p-values for Different Metrics (n=250)}\label{Table35}
\begin{tabular}{lccccc}
\hline
\textbf{Metric} & \textbf{Fligner-Policello Test} & \textbf{Mann-Whitney U Test} & \textbf{Kruskal-Wallis H Test} & \textbf{Levene's Test} & \textbf{Brown-Forsythe Test} \\
\hline
RMSE\(_{ATE}\)  & $0.7607$                & \textit{N/A}    & \textit{N/A}    & $0.0256$               & $0.0342$                \\
MAE\(_{ATE}\)   & $0.3566$                & \textit{N/A}    & \textit{N/A}    & $0.0128$               & $0.0165$                \\
MAPE\(_{ATE}\)  & \textit{N/A}            & $0.0182$        & $0.0181$        & $0.6651$               & $0.7818$                \\
Len\(_{ATE}\)   & \textit{N/A}            & $0.0330$        & $0.0329$        & $0.1170$               & $0.1303$                \\
RMSE\(_{CATE}\) & $3.43 \times 10^{-24}$  & \textit{N/A}    & \textit{N/A}    & $6.36 \times 10^{-10}$ & $1.80 \times 10^{-7}$   \\
MAE\(_{CATE}\)  & $6.63 \times 10^{-4}$   & \textit{N/A}    & \textit{N/A}    & $9.24 \times 10^{-8}$  & $6.09 \times 10^{-6}$   \\
MAPE\(_{CATE}\) & \textit{N/A}            & $0.0094$        & $0.0093$        & $0.3417$               & $0.4743$                \\
Len\(_{CATE}\)  & $1.42 \times 10^{-79}$  & \textit{N/A}    & \textit{N/A}    & $6.56 \times 10^{-11}$ & $4.04 \times 10^{-10}$  \\
SEC\(_{ATE}\)   & $4.49 \times 10^{-5}$   & \textit{N/A}    & \textit{N/A}    & $9.20 \times 10^{-5}$  & $0.0237$                \\
AEC\(_{ATE}\)   & $4.49 \times 10^{-5}$   & \textit{N/A}    & \textit{N/A}    & $9.14 \times 10^{-9}$  & $0.0034$                \\
SEC\(_{CATE}\)\textbf{*}  & \textit{N/A}            & \textit{N/A}    & \textit{N/A}    & \textit{N/A}               & \textit{N/A}            \\
AEC\(_{CATE}\)\textbf{*}  & \textit{N/A}            & \textit{N/A}    & \textit{N/A}    & \textit{N/A}               & \textit{N/A}            \\
\hline
\end{tabular}
\end{table}
\textbf{*:} SEC\(_{CATE}\) and AEC\(_{CATE}\) had p-values of $1.67 \times 10^{-24}$ and $3.07 \times 10^{-24}$ for the T-test and $0.65$ and $0.25$ for the F-test respectively.
\end{landscape}

\begin{landscape}
\begin{table}[h!]
\centering
\caption{Statistical Test Results: p-values for Different Metrics (n=500)}\label{Table36}
\begin{tabular}{lccccc}
\hline
\textbf{Metric} & \textbf{Fligner-Policello Test} & \textbf{Mann-Whitney U Test} & \textbf{Kruskal-Wallis H Test} & \textbf{Levene's Test} & \textbf{Brown-Forsythe Test} \\
\hline
RMSE\(_{ATE}\)  & $0.8850$                & \textit{N/A}    & \textit{N/A}    & $0.0128$               & $0.0149$                \\
MAE\(_{ATE}\)   & $0.0456$                & \textit{N/A}    & \textit{N/A}    & $0.0008$               & $0.0011$                \\
MAPE\(_{ATE}\)  & \textit{N/A}            & $0.0370$        & $0.0369$        & $0.2122$               & $0.4973$                \\
Len\(_{ATE}\)   & \textit{N/A}            & $1.88 \times 10^{-6}$ & $1.87 \times 10^{-6}$  & $0.8948$               & $0.8562$                \\
RMSE\(_{CATE}\) & $3.60 \times 10^{-84}$  & \textit{N/A}    & \textit{N/A}    & $6.35 \times 10^{-10}$ & $8.42 \times 10^{-8}$   \\
MAE\(_{CATE}\)  & $9.01 \times 10^{-17}$  & \textit{N/A}    & \textit{N/A}    & $1.42 \times 10^{-8}$  & $1.07 \times 10^{-6}$   \\
MAPE\(_{CATE}\) & \textit{N/A}            & $0.0014$        & $0.0014$        & $0.5627$               & $0.6286$                \\
Len\(_{CATE}\)  & $1.99 \times 10^{-66}$  & \textit{N/A}    & \textit{N/A}    & $4.16 \times 10^{-5}$  & $1.28 \times 10^{-4}$   \\
SEC\(_{ATE}\)   & $0.6648$                & \textit{N/A}    & \textit{N/A}    & $2.10 \times 10^{-15}$ & $5.91 \times 10^{-5}$   \\
AEC\(_{ATE}\)   & $0.6648$                & \textit{N/A}    & \textit{N/A}    & $5.06 \times 10^{-19}$ & $7.21 \times 10^{-6}$   \\
SEC\(_{CATE}\)  & $9.40 \times 10^{-27}$  & \textit{N/A}    & \textit{N/A}    & $0.0026$               & $0.0055$                \\
AEC\(_{CATE}\)\textbf{*}  & \textit{N/A}            & \textit{N/A}    & \textit{N/A}    & \textit{N/A}              & \textit{N/A}            \\
\hline
\end{tabular}
\end{table}
\textbf{*:} AEC\(_{CATE}\) had p-values of $2.18 \times 10^{-16}$ for the T-test and $0.02$ for the F-test.
\end{landscape}

\subsubsection{Conclusion}

The results of this set of DGPs demonstrate the fact that (semi-)parametric models outperform nonparametric models when correctly specified for the underlying DGP \parencite{Jabot2015,Liu2011,Robinson2010}. However, the BCF model is interestingly superior to the ps-BART model in uncertainty estimation for the CATE function estimation for the Nonlinear Relationship DGP while being worse in point-wise estimation. However, this can be easily explained by the fact that the BCF model is using considerably larger confidence interval lengths than the ps-BART model, creating the false impression that the BCF model has a better uncertainty estimation for the CATE function estimation. Lastly, the results of this set of DGPs confirm the hypothesis that for significantly nonlinear DGPs the ps-BART model becomes more robust than the BCF model due to the considerably misspecification of the latter while for slightly nonlinear/considerably linear DGPs, the benchmark model is more robust.

\section{Conclusion}  \label{Sec5}

This paper presents a significant advancement in the literature on causal inference, particularly in the estimation of ATE and CATE for continuous treatments. The proposed generalized ps-BART model effectively addresses the limitations of the BCF model by offering a more flexible and accurate approach to capturing the potentially nonlinear and complex relationships between the treatment, covariates, and outcome variables. The nonparametric nature of the ps-BART model provides several advantages over its benchmark counterpart, including its ability to adapt to a variety of functional forms, its reduced risk of misspecification, and its suitability across a wide range of DGPs, increasing its potential use in the Health and Social Sciences research.

The experimental results of this paper, based on three distinct sets of DGPs, further highlight the strengths of the ps-BART model relative to the BCF model. It is worth remembering that when we say below that a DGP is (non)linear, we mean that the DGP has a (non)linear relationship between the treatment and outcome.

For moderately nonlinear DGPs (i.e., the first set of DGPs), the ps-BART model consistently outperforms the BCF model in both point-wise and uncertainty estimation of ATE and CATE functions, though only for when the sample size is \( n \geq 250 \) for slightly nonlinear DGPs (i.e., when the semi-parametric assumption of the BCF model is not far off from the true DGP). Interestingly, while the BCF model demonstrated more robustness in ATE point-wise estimation in slightly nonlinear scenarios, this performance edge is attributable to the benchmark model’s slightly alignment with linearity assumptions in these DGPs (where the relationship between the treatment and outcome becomes linear for high treatment doses). Nonetheless, the proposed model’s flexibility gives it a distinct advantage in capturing even subtle nonlinear effects, which positions ps-BART as the more reliable model for most real-world data where the true functional form is not strictly linear.

In highly nonlinear DGPs (i.e., the second set of DGPs), the ps-BART model demonstrated clear superiority in both point-wise and uncertainty estimation. The flexibility of the ps-BART model allowed it to handle the complex relationships between the treatment and outcome more effectively than the BCF model, which exhibited significant performance degradation due to misspecification. Interestingly, in this set of DGPs, the ps-BART model also proved to be more robust than the BCF model in both ATE and CATE estimation, further demonstrating that for significantly nonlinear DGPs the ps-BART model becomes more robust than the BCF model due to the considerably misspecification of the latter while for slightly nonlinear/considerably linear DGPs, the benchmark model is more robust

Regarding the third set of DGPs considered in this study, it can be seen that in cases where the DGP aligns closely with the parametric assumptions underlying the BCF model, as expected, the BCF model performed better in point-wise estimation of the ATE and CATE functions, as already expected from the theory \parencite{Jabot2015,Liu2011,Robinson2010}. However, the results also showed that the BCF model achieved better uncertainty estimation in CATE for certain the nonlinear DGP considered in  the third set of DGPs, which was largely due to its reliance on overly wide confidence intervals. This gave the false impression of more accurate uncertainty estimation, when in reality, the ps-BART model provided tighter intervals that more accurately reflected the underlying distribution of the treatment effect. These results reaffirm the hypothesis that ps-BART is better suited for nonlinear DGPs, whereas the BCF model maintains some advantages in linear DGPs.

Hence, this research addresses a critical gap in the current literature on causal inference for continuous treatments by providing a fully nonparametric model that avoids the restrictive assumptions imposed by parametric or semi-parametric models such as the BCF model. The ps-BART model’s ability to provide accurate and flexible estimates of ATE and CATE functions, while simultaneously offering probabilistic estimation through uncertainty quantification, makes it a valuable addition to the toolkit for causal inference. The development of this model represents a key step toward closing the gap in probabilistic estimation of causal effects for continuous treatments, where existing the existing benchmark model fails to provide reliable estimates for when the relationship between the treatment and outcome is nonlinear. By introducing a more flexible alternative, this research opens up new avenues for further exploration in the field and the application of the proposed model in the Health and Social Sciences literature. 

Finally, regarding causal inference research (i.e., the research to create the tools to properly perform causal inference), several future research directions emerge from the results of this study. Firstly, the ps-BART model could be extended to handle time-varying treatments, where the treatment and outcome are observed over multiple time points. The flexibility of the model positions it well for this task, which could further expand its applicability in real-world scenarios such as clinical trials or longitudinal studies. Additionally, future research could explore the performance of the ps-BART model in high-dimensional covariate spaces, where feature selection and model regularization play crucial roles. Understanding how the proposed model performs under different regularization strategies in high-dimensional settings would be valuable for extending its use to large-scale observational studies. Finally, future research could investigate optimal hyperparameter and prior tuning for ps-BART using cross-validation techniques to further enhance the model’s performance across different DGPs.

\section{Supplementary Research Material and Code}
The supplementary research material and code used in this paper can be found in the following GitHub repository: 

\printbibliography

@article{Yao2021,
  title = {A Survey on Causal Inference},
  volume = {15},
  ISSN = {1556-472X},
  url = {http://dx.doi.org/10.1145/3444944},
  DOI = {10.1145/3444944},
  number = {5},
  journal = {ACM Transactions on Knowledge Discovery from Data},
  publisher = {Association for Computing Machinery (ACM)},
  author = {Yao,  Liuyi and Chu,  Zhixuan and Li,  Sheng and Li,  Yaliang and Gao,  Jing and Zhang,  Aidong},
  year = {2021},
  month = may,
  pages = {1–46}
}

@inproceedings{10.1145/3394486.3406460,
author = {Cui, Peng and Shen, Zheyan and Li, Sheng and Yao, Liuyi and Li, Yaliang and Chu, Zhixuan and Gao, Jing},
title = {Causal Inference Meets Machine Learning},
year = {2020},
isbn = {9781450379984},
publisher = {Association for Computing Machinery},
address = {New York, NY, USA},
url = {https://doi.org/10.1145/3394486.3406460},
doi = {10.1145/3394486.3406460},
pages = {3527–3528},
numpages = {2},
location = {Virtual Event, CA, USA},
series = {KDD '20}
}

@article{Kuang2020,
  title = {Causal Inference},
  volume = {6},
  ISSN = {2095-8099},
  url = {http://dx.doi.org/10.1016/j.eng.2019.08.016},
  DOI = {10.1016/j.eng.2019.08.016},
  number = {3},
  journal = {Engineering},
  publisher = {Elsevier BV},
  author = {Kuang,  Kun and Li,  Lian and Geng,  Zhi and Xu,  Lei and Zhang,  Kun and Liao,  Beishui and Huang,  Huaxin and Ding,  Peng and Miao,  Wang and Jiang,  Zhichao},
  year = {2020},
  month = mar,
  pages = {253–263}
}

@article{Crown2019,
  title = {Real-World Evidence,  Causal Inference,  and Machine Learning},
  volume = {22},
  ISSN = {1098-3015},
  url = {http://dx.doi.org/10.1016/j.jval.2019.03.001},
  DOI = {10.1016/j.jval.2019.03.001},
  number = {5},
  journal = {Value in Health},
  publisher = {Elsevier BV},
  author = {Crown,  William H.},
  year = {2019},
  month = may,
  pages = {587–592}
}

@article{Grimmer2014,
  title = {We Are All Social Scientists Now: How Big Data,  Machine Learning,  and Causal Inference Work Together},
  volume = {48},
  ISSN = {1537-5935},
  url = {http://dx.doi.org/10.1017/S1049096514001784},
  DOI = {10.1017/s1049096514001784},
  number = {01},
  journal = {PS: Political Science \& Politics},
  publisher = {Cambridge University Press (CUP)},
  author = {Grimmer,  Justin},
  year = {2014},
  month = dec,
  pages = {80–83}
}

@article{Ohlsson2020,
  title = {Applying Causal Inference Methods in Psychiatric Epidemiology: A Review},
  volume = {77},
  ISSN = {2168-622X},
  url = {http://dx.doi.org/10.1001/jamapsychiatry.2019.3758},
  DOI = {10.1001/jamapsychiatry.2019.3758},
  number = {6},
  journal = {JAMA Psychiatry},
  publisher = {American Medical Association (AMA)},
  author = {Ohlsson,  Henrik and Kendler,  Kenneth S.},
  year = {2020},
  month = jun,
  pages = {637}
}

@article{Bembom2007,
  title = {A practical illustration of the importance of realistic individualized treatment rules in causal inference},
  volume = {1},
  ISSN = {1935-7524},
  url = {http://dx.doi.org/10.1214/07-EJS105},
  DOI = {10.1214/07-ejs105},
  number = {none},
  journal = {Electronic Journal of Statistics},
  publisher = {Institute of Mathematical Statistics},
  author = {Bembom,  Oliver and van der Laan,  Mark J.},
  year = {2007},
  month = jan 
}

@article{Sobel2000,
  title = {Causal Inference in the Social Sciences},
  volume = {95},
  ISSN = {1537-274X},
  url = {http://dx.doi.org/10.1080/01621459.2000.10474243},
  DOI = {10.1080/01621459.2000.10474243},
  number = {450},
  journal = {Journal of the American Statistical Association},
  publisher = {Informa UK Limited},
  author = {Sobel,  Michael E.},
  year = {2000},
  month = jun,
  pages = {647–651}
}

@book{murnane2010methods,
  title={Methods matter: Improving causal inference in educational and social science research},
  author={Murnane, Richard J and Willett, John B},
  year={2010},
  publisher={Oxford University Press}
}

@misc{https://doi.org/10.48550/arxiv.1811.00157,
  doi = {10.48550/ARXIV.1811.00157},
  url = {https://arxiv.org/abs/1811.00157},
  author = {Lee,  Ying-Ying},
  title = {Partial Mean Processes with Generated Regressors: Continuous Treatment Effects and Nonseparable Models},
  publisher = {arXiv},
  year = {2018},
  copyright = {arXiv.org perpetual,  non-exclusive license}
}

@article{Wu2022,
  title = {Matching on Generalized Propensity Scores with Continuous Exposures},
  volume = {119},
  ISSN = {1537-274X},
  url = {http://dx.doi.org/10.1080/01621459.2022.2144737},
  DOI = {10.1080/01621459.2022.2144737},
  number = {545},
  journal = {Journal of the American Statistical Association},
  publisher = {Informa UK Limited},
  author = {Wu,  Xiao and Mealli,  Fabrizia and Kioumourtzoglou,  Marianthi-Anna and Dominici,  Francesca and Braun,  Danielle},
  year = {2022},
  month = dec,
  pages = {757–772}
}

@misc{https://doi.org/10.48550/arxiv.1907.13258,
  doi = {10.48550/ARXIV.1907.13258},
  url = {https://arxiv.org/abs/1907.13258},
  author = {Rothenh\"{a}usler,  Dominik and Yu,  Bin},
  title = {Incremental causal effects},
  publisher = {arXiv},
  year = {2019},
  copyright = {arXiv.org perpetual,  non-exclusive license}
}

@misc{Hirano2004,
  title = {The Propensity Score with Continuous Treatments},
  ISBN = {9780470090459},
  ISSN = {1940-6347},
  url = {http://dx.doi.org/10.1002/0470090456.ch7},
  DOI = {10.1002/0470090456.ch7},
  journal = {Applied Bayesian Modeling and Causal Inference from Incomplete‐Data Perspectives},
  publisher = {Wiley},
  author = {Hirano,  Keisuke and Imbens,  Guido W.},
  year = {2004},
  month = jul,
  pages = {73–84}
}

@article{Kennedy2016,
  title = {Non-parametric Methods for Doubly Robust Estimation of Continuous Treatment Effects},
  volume = {79},
  ISSN = {1467-9868},
  url = {http://dx.doi.org/10.1111/rssb.12212},
  DOI = {10.1111/rssb.12212},
  number = {4},
  journal = {Journal of the Royal Statistical Society Series B: Statistical Methodology},
  publisher = {Oxford University Press (OUP)},
  author = {Kennedy,  Edward H. and Ma,  Zongming and McHugh,  Matthew D. and Small,  Dylan S.},
  year = {2016},
  month = sep,
  pages = {1229–1245}
}

@article{Moodie2010,
  title = {Estimation of dose–response functions for longitudinal data using the generalised propensity score},
  volume = {21},
  ISSN = {1477-0334},
  url = {http://dx.doi.org/10.1177/0962280209340213},
  DOI = {10.1177/0962280209340213},
  number = {2},
  journal = {Statistical Methods in Medical Research},
  publisher = {SAGE Publications},
  author = {Moodie,  Erica EM and Stephens,  David A},
  year = {2010},
  month = may,
  pages = {149–166}
}

@misc{https://doi.org/10.48550/arxiv.2007.09845,
  doi = {10.48550/ARXIV.2007.09845},
  url = {https://arxiv.org/abs/2007.09845},
  author = {Woody,  Spencer and Carvalho,  Carlos M. and Hahn,  P. Richard and Murray,  Jared S.},
  title = {Estimating heterogeneous effects of continuous exposures using Bayesian tree ensembles: revisiting the impact of abortion rates on crime},
  publisher = {arXiv},
  year = {2020},
  copyright = {arXiv.org perpetual,  non-exclusive license}
}

@article{Hahn2020,
  title = {Bayesian Regression Tree Models for Causal Inference: Regularization,  Confounding,  and Heterogeneous Effects (with Discussion)},
  volume = {15},
  ISSN = {1936-0975},
  url = {http://dx.doi.org/10.1214/19-BA1195},
  DOI = {10.1214/19-ba1195},
  number = {3},
  journal = {Bayesian Analysis},
  publisher = {Institute of Mathematical Statistics},
  author = {Hahn,  P. Richard and Murray,  Jared S. and Carvalho,  Carlos M.},
  year = {2020},
  month = sep 
}

@InProceedings{pmlr-v130-curth21a,
  title = 	 { Nonparametric Estimation of Heterogeneous Treatment Effects: From Theory to Learning Algorithms },
  author =       {Curth, Alicia and van der Schaar, Mihaela},
  booktitle = 	 {Proceedings of The 24th International Conference on Artificial Intelligence and Statistics},
  pages = 	 {1810--1818},
  year = 	 {2021},
  editor = 	 {Banerjee, Arindam and Fukumizu, Kenji},
  volume = 	 {130},
  series = 	 {Proceedings of Machine Learning Research},
  month = 	 {13--15 Apr},
  publisher =    {PMLR},
  url = 	 {https://proceedings.mlr.press/v130/curth21a.html},
}

@book{CausalBook,
  title     = "Applied Causal Inference Powered by ML and AI",
  author    = "Chernozhukov, Victor and Hansen, Christian and Kallus, Nathan and Spindler, Martin and Syrgkanis, Vasilis",
  year      = 2024,
  publisher = "Online"
}

@article{Knzel2019,
  title = {Metalearners for estimating heterogeneous treatment effects using machine learning},
  volume = {116},
  ISSN = {1091-6490},
  url = {http://dx.doi.org/10.1073/pnas.1804597116},
  DOI = {10.1073/pnas.1804597116},
  number = {10},
  journal = {Proceedings of the National Academy of Sciences},
  publisher = {Proceedings of the National Academy of Sciences},
  author = {K\"{u}nzel,  S\"{o}ren R. and Sekhon,  Jasjeet S. and Bickel,  Peter J. and Yu,  Bin},
  year = {2019},
  month = feb,
  pages = {4156–4165}
}

@misc{https://doi.org/10.48550/arxiv.2004.14497,
  doi = {10.48550/ARXIV.2004.14497},
  url = {https://arxiv.org/abs/2004.14497},
  author = {Kennedy,  Edward H.},
  title = {Towards optimal doubly robust estimation of heterogeneous causal effects},
  publisher = {arXiv},
  year = {2020},
  copyright = {arXiv.org perpetual,  non-exclusive license}
}

@misc{https://doi.org/10.48550/arxiv.1712.04912,
  doi = {10.48550/ARXIV.1712.04912},
  url = {https://arxiv.org/abs/1712.04912},
  author = {Nie,  Xinkun and Wager,  Stefan},
  title = {Quasi-Oracle Estimation of Heterogeneous Treatment Effects},
  publisher = {arXiv},
  year = {2017},
  copyright = {arXiv.org perpetual,  non-exclusive license}
}

@inbook{Liu2016,
  title = {Special topics on linear mixed models},
  ISBN = {9780128013427},
  url = {http://dx.doi.org/10.1016/B978-0-12-801342-7.00007-1},
  DOI = {10.1016/b978-0-12-801342-7.00007-1},
  booktitle = {Methods and Applications of Longitudinal Data Analysis},
  publisher = {Elsevier},
  author = {Liu,  Xian},
  year = {2016},
  pages = {205–242}
}

@inbook{Martin2022,
  title = {Direct Gibbs posterior inference on risk minimizers: Construction,  concentration,  and calibration},
  ISBN = {9780323952682},
  ISSN = {0169-7161},
  url = {http://dx.doi.org/10.1016/bs.host.2022.06.004},
  DOI = {10.1016/bs.host.2022.06.004},
  booktitle = {Advancements in Bayesian Methods and Implementation},
  publisher = {Elsevier},
  author = {Martin,  Ryan and Syring,  Nicholas},
  year = {2022},
  pages = {1–41}
}

@inbook{Cai2003,
  title = {Nonparametric Methods in Continuous-Time Finance: A Selective Review},
  ISBN = {9780444513786},
  url = {http://dx.doi.org/10.1016/B978-044451378-6/50019-3},
  DOI = {10.1016/b978-044451378-6/50019-3},
  booktitle = {Recent Advances and Trends in Nonparametric Statistics},
  publisher = {Elsevier},
  author = {Cai,  Zongwu and Hong,  Yongmiao},
  year = {2003},
  pages = {283–302}
}

@inbook{Hayashi2011,
  title = {Structural Equation Modeling},
  ISBN = {9780444537379},
  url = {http://dx.doi.org/10.1016/B978-0-444-53737-9.50010-4},
  DOI = {10.1016/b978-0-444-53737-9.50010-4},
  booktitle = {Essential Statistical Methods for Medical Statistics},
  publisher = {Elsevier},
  author = {Hayashi,  Kentaro and Bentler,  Peter M. and Yuan,  Ke-Hai},
  year = {2011},
  pages = {202–234}
}

@article{Chipman2010,
  title = {BART: Bayesian additive regression trees},
  volume = {4},
  ISSN = {1932-6157},
  url = {http://dx.doi.org/10.1214/09-AOAS285},
  DOI = {10.1214/09-aoas285},
  number = {1},
  journal = {The Annals of Applied Statistics},
  publisher = {Institute of Mathematical Statistics},
  author = {Chipman,  Hugh A. and George,  Edward I. and McCulloch,  Robert E.},
  year = {2010},
  month = mar 
}

@article{Hahn2018,
  title = {Regularization and Confounding in Linear Regression for Treatment Effect Estimation},
  volume = {13},
  ISSN = {1936-0975},
  url = {http://dx.doi.org/10.1214/16-BA1044},
  DOI = {10.1214/16-ba1044},
  number = {1},
  journal = {Bayesian Analysis},
  publisher = {Institute of Mathematical Statistics},
  author = {Hahn,  P. Richard and Carvalho,  Carlos M. and Puelz,  David and He,  Jingyu},
  year = {2018},
  month = mar 
}

@article{Jabot2015,
  title = {Why preferring parametric forecasting to nonparametric methods?},
  volume = {372},
  ISSN = {0022-5193},
  url = {http://dx.doi.org/10.1016/j.jtbi.2014.07.038},
  DOI = {10.1016/j.jtbi.2014.07.038},
  journal = {Journal of Theoretical Biology},
  publisher = {Elsevier BV},
  author = {Jabot,  Franck},
  year = {2015},
  month = may,
  pages = {205–210}
}

@article{Liu2011,
  title = {Parametric or nonparametric? A parametricness index for model selection},
  volume = {39},
  ISSN = {0090-5364},
  url = {http://dx.doi.org/10.1214/11-AOS899},
  DOI = {10.1214/11-aos899},
  number = {4},
  journal = {The Annals of Statistics},
  publisher = {Institute of Mathematical Statistics},
  author = {Liu,  Wei and Yang,  Yuhong},
  year = {2011},
  month = aug 
}

@article{Robinson2010,
  title = {Efficient estimation of the semiparametric spatial autoregressive model},
  volume = {157},
  ISSN = {0304-4076},
  url = {http://dx.doi.org/10.1016/j.jeconom.2009.10.031},
  DOI = {10.1016/j.jeconom.2009.10.031},
  number = {1},
  journal = {Journal of Econometrics},
  publisher = {Elsevier BV},
  author = {Robinson,  P.M.},
  year = {2010},
  month = jul,
  pages = {6–17}
}

@misc{Souto2024,
  doi = {10.48550/ARXIV.2409.05161},
  url = {https://arxiv.org/abs/2409.05161},
  author = {Souto,  Hugo Gobato and Neto,  Francisco Louzada},
  title = {Really Doing Great at Model Evaluation for CATE Estimation? A Critical Consideration of Current Model Evaluation Practices in Treatment Effect Estimation},
  publisher = {arXiv},
  year = {2024},
  copyright = {Creative Commons Attribution Non Commercial Share Alike 4.0 International}
}

\end{document}